\documentclass[14pt a4paper]{article}
\usepackage{algorithm}
\usepackage{algpseudocode}
\usepackage{amsmath}
\usepackage{amssymb}
\usepackage{bm}
\usepackage{booktabs}
\usepackage{color}
\usepackage{cite}
\usepackage{epsfig}
\usepackage{epstopdf}
\usepackage{flafter}
\usepackage{cite}
\usepackage{float}
\usepackage{ifthen}
\usepackage{multirow}
\usepackage{makecell} 
\usepackage{pifont}
\usepackage{subfigure}
\usepackage{times}
\usepackage{threeparttable}
\usepackage{url}  
\usepackage{subfigure,graphicx}
\usepackage{float}
\usepackage{color,multirow}
\usepackage{makecell}
\usepackage{url}  
\usepackage{ifthen}  
\usepackage[T1]{fontenc}
\usepackage{graphicx}
\usepackage[colorlinks,linkcolor=red,anchorcolor=gray,citecolor=blue,urlcolor=green]{hyperref}
\usepackage{graphics}
\usepackage{makecell}
\usepackage{booktabs}
\renewcommand{\figurename}{Fig.}
\newcommand{\R}{\mathbb{R}}
\newcommand{\C}{\mathbb{C}}

\newcommand{\<}{\langle}
\renewcommand{\>}{\rangle}

\newcommand{\rank}{\operatorname{rank}}

\renewcommand{\theequation}{\thesection.\arabic{equation}}
\renewcommand{\thetable}{\thesection.\arabic{table}}
\renewcommand{\thefigure}{\thesection.\arabic{figure}}

\newtheorem{theorem}{\textbf{Theorem}}
\newtheorem{lemma}{\textbf{Lemma}}
\newtheorem{definition}{\textbf{Definition}}
\newcommand{\argmin}{\mathop{\mathrm{arg\,min}}}

\usepackage[left=1.25in,right=1.25in,top=1.25in,bottom=1.25in,letterpaper]{geometry}
\date{ }

\author{Yun-Yang Liu
\thanks{Y.-Y. Liu, X.-L. Zhao, Y.-B. Zheng, and T.-Z Huang are with the School of Mathematical Sciences, University of Electronic Science and Technology of China, Chengdu, Sichuan 611731, P. R. China. E-mails: lyymath@126.com, xlzhao122003@163.com, zhengyubang@163.com, and tingzhuhuang@126.com.},
Xi-Le Zhao,
Guang-Jing Song
\thanks{G.-J Song is with the School of  Mathematics and Information Sciences, Weifang University, Weifang 261061, P. R. China. E-mail:sgjshu@163.com.},
Yu-Bang Zheng,
Ting-Zhu Huang}
\begin{document}
\title{Fully-Connected Tensor Network Decomposition for Robust Tensor Completion Problem}
\maketitle
\begin{abstract}
 The robust tensor completion (RTC) problem, which aims to reconstruct a low-rank tensor from partially observed tensor contaminated by a sparse tensor, has received increasing attention. In this paper, by leveraging the superior expression of the fully-connected tensor network (FCTN) decomposition, we propose a $\textbf{FCTN}$-based $\textbf{r}$obust $\textbf{c}$onvex optimization model (RC-FCTN) for the RTC problem. Then, we rigorously establish the exact recovery guarantee for the RC-FCTN. For solving the constrained optimization model RC-FCTN, we develop an alternating direction method of multipliers (ADMM)-based algorithm,  which enjoys the global convergence guarantee. Moreover, we suggest a $\textbf{FCTN}$-based $\textbf{r}$obust $\textbf{n}$on$\textbf{c}$onvex optimization model (RNC-FCTN) for the RTC problem. A proximal alternating minimization (PAM)-based algorithm is developed to solve the proposed RNC-FCTN. Meanwhile, we theoretically derive the convergence of the PAM-based algorithm. Comprehensive numerical experiments in several applications, such as video completion and video background subtraction, demonstrate that proposed methods are superior to several state-of-the-art methods.
\end{abstract}


\section{Introduction}
Owing to various unpredictable or unavoidable reasons, the observed data is often contaminated by noise and suffers from missing information \cite{multi,8606166,YANG2020124783}, which significantly limits the accuracy of subsequent applications. The data recovery problems, such as image/video completion \cite{article1,CHEN2021100} and image/video noise removal \cite{xiong,zhuang,jia}, are very significant. As the generalization of matrices \cite{li}, tensors \cite{che,che2} can represent higher-dimensional data, most of which is assumed to be low-rank in the recovery problems. Therefore, many data recovery problems are uniformly modeled as a robust tensor completion (RTC) problem, aiming to restructure a low-rank component and a sparse component from the observed data.

 \begin{figure}[!htp]
\centering
\includegraphics[width=0.99\linewidth]{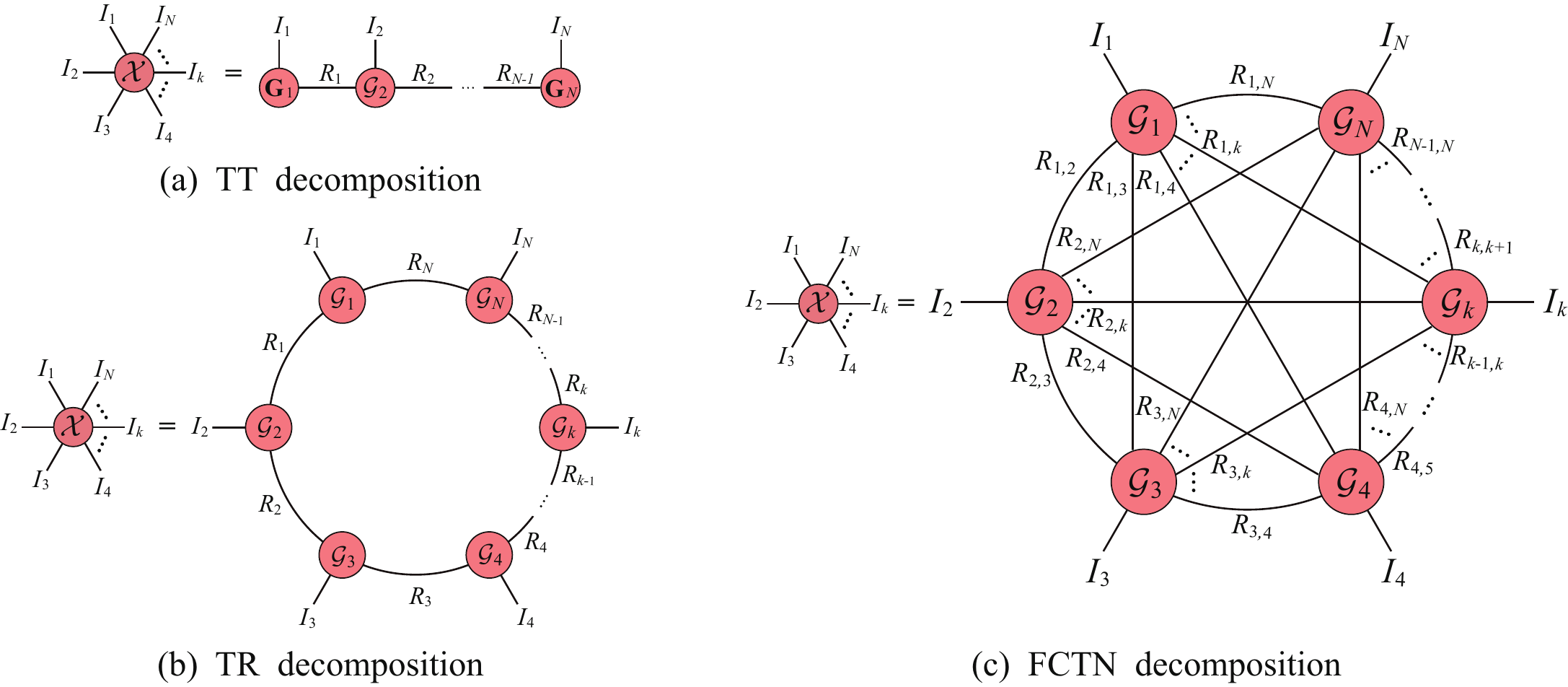}
  \caption{The illustration of tensor network decomposition.}\vspace{-0.4cm}
  \label{decomposition}
\end{figure}

Mathematically, the RTC problem can be expressed as:
\begin{equation}\label{add2}
\min_{\mathcal{X},\mathcal{E}} ~~ \rank(\mathcal{X}) +\lambda \|\mathcal{E}\|_1,    ~~\text{s.t. }~  \mathcal{P}_\Omega (\mathcal{X}+\mathcal{E})=\mathcal{P}_\Omega (\mathcal{O}),
\end{equation}
where $\mathcal{O}$ is the observed data, $\mathcal{X}$ and $\mathcal{E}$ are the low-rank component and the sparse component, $\lambda$ is regularization parameter, and $\mathcal{P}_\Omega$ is a projection that the entries in the set $\Omega$ are themselves while the other entries are set to zeros. When $\Omega$ is the whole set, (\ref{add2}) transforms into sparse noise removal problem. When $\lambda=0$, (\ref{add2}) transforms into tensor completion problem. Unlike the matrix case, there exist different kinds of tensor rank, such as Tucker rank \cite{Tucker}, multi-rank and tubal rank \cite{tubal}, tensor train (TT) rank \cite{TT}, and tensor ring (TR) rank \cite{TR}, which are derived from the corresponding tensor decompositions. In general, the minimization of tensor rank is NP-hard \cite{2512329}, the convex/nonconvex relaxation of tensor rank or the low-rank tensor decomposition is usually used instead of the minimization of tensor rank.

Tucker decomposition decomposes an $N$th-order tensor $\mathcal{X} \in \R^{I_1 \times I_2 \times \cdots \times I_N}$ into a small-sized $N$th-order core tensor $\mathcal{G}$ multiplied by a matrix along each mode, i.e., $\mathcal{X}=\mathcal{G} \times_1 \textbf{U}_1 \times_2 \textbf{U}_2 \times_3 \textbf{U}_3$. Tucker rank is a vector whose the $k$th entry is the rank of the mode-$k$ matricization of  $\mathcal{X}$, i.e.,
\begin{equation}
\text{rank}_{\rm{Tu}}(\mathcal{X}):=(\text{rank}(\textbf{X}_{(1)}),\text{rank}(\textbf{X}_{(2)}),\cdots ,\text{rank}(\textbf{X}_{(N)})),
\end{equation}
where $\textbf{X}_{(k)} \in \R^{I_k \times (I_1 \cdots I_{k-1} I_{k+1} \cdots I_N)}$ is the mode-$k$ matricization of $\mathcal{X}$.
Based on recent studies that the nuclear norm ($\|\cdot\|_\ast$) was the convex relaxation of the matrix rank, Liu $\mathit{et}$ $\mathit{al.}$ \cite{6138863} proposed the sum of nuclear norms (SNN) of all unfolding matrices $\sum_{k=1}^N \alpha_k \|\textbf{X}_{(k)}\|_\ast$ as the convex surrogate of tensor Tucker rank, where $\alpha_k\geq 0$ and $\sum_{k=1}^N \alpha_k=1$. Based on the SNN, Huang $\mathit{et}$ $\mathit{al.}$ \cite{2512321} studied the robust low-Tucker-rank tensor completion problem and gave the theoretical guarantee of exact recovery under tensor certain incoherence conditions. The framework of the SNN-based model is easier to calculate. However, the SNN is not the tightest convex envelope of the sum of ranks of unfolding matrices of a tensor. Moreover, unfolding a tensor into matrices along one mode is an unbalanced matricization scheme. Therefore, Tucker rank cannot suitably capture the global information of the tensor.

Tensor singular value decomposition (t-SVD) \cite{KILMER2011641} decomposes a third-order tensor $\mathcal{X} \in \R^{I_1 \times I_2 \times I_3}$ into the tensor-product ($\ast$) of two orthogonal tensors $\mathcal{U}$, $\mathcal{V}$ and an f-diagonal tensor $\mathcal{S}$, i.e., $\mathcal{X}=\mathcal{U} \ast \mathcal{S}\ast \mathcal{V}^{\rm{T}}$. Tensor multi-rank \cite{tubal} is a vector whose entries are the rank of the frontal slice of $\hat{\mathcal{X}}$,
\begin{equation}
\text{rank}_{\rm{mul}}(\mathcal{X}):=(\text{rank}(\hat{\mathcal{X}}^{(1)}),\text{rank}(
\hat{\mathcal{X}}^{(2)}),\cdots,\text{rank}(\hat{\mathcal{X}}^{(I_3)})),
\end{equation}
where $\hat{\mathcal{X}}={\rm{fft}}(\mathcal{X},[],3)$ denotes the tensor obtained by performing one-dimensional Fourier transform on each tube of $\mathcal{X}$, and $\hat{\mathcal{X}}^{(k)}$ denotes the $k$th frontal slice of $\hat{\mathcal{X}}$. Tensor tubal rank \cite{tubal} is the largest element of the tensor multi-rank. Semerci $\mathit{et}$ $\mathit{al.}$ \cite{6737273} developed the tensor nuclear norm (TNN) as the convex envelope of $\mathit{l}_1$ norm of the multi-rank. Utilizing the TNN, Jiang $\mathit{et}$ $\mathit{al.}$ \cite{Jiang} studied the RTC problem and provided the theoretical guarantee for the exact recovery. Recently, Song  $\mathit{et}$ $\mathit{al.}$  \cite{8606165} generalized the t-SVD theory via multiplying by a unity matrix on all tubes instead of the fixed discrete Fourier transform matrix. A data-driven RTC model is also proposed, which has been shown to have obvious advantages in processing third-order tensors. However, the t-SVD focuses on third-order tensors. For high-order data, such as color videos and multi-temporal remote sensing images, the t-SVD and the corresponding TNN-based models may not effectively capture the low-dimensional structure of data.

TT decomposition (as shown in Fig. \ref{decomposition} (a)) \cite{TT} decomposes an $N$th-order tensor $\mathcal{X} \in \R^{I_1 \times I_2 \times \cdots \times I_N}$ into two matrices and $N-2$ third-order tensors, and the element-wise form is expressed as
\begin{equation}
\mathcal{X}(i_1,i_2,\cdots,i_N)=\sum_{r_1 =1}^{R_1} \sum_{r_2 =1}^{R_2}\cdots \sum_{r_{N-1} =1}^{R_{N-1}} \{\textbf{G}_1(i_1,r_1)\mathcal{G}_2(r_1,i_2,r_2)\cdots \textbf{G}_N(r_{N-1},i_N)\}.
\end{equation}
TT rank is defined as an ($N-1$)-dimensional vector $(R_1,R_2,\cdots,R_{N-1})$. I. V. Oseledets \cite{TT} proved that there existed a TT decomposition such that
\begin{equation}
R_k\leq\text{rank}(\textbf{X}_{[k]}),~~~k=1,2,\cdots,N-1,
\end{equation}
where $\textbf{X}_{[k]} \in \R^{\Pi_{n=1}^k I_n \times \Pi_{n=k+1}^N I_n}$ is the $k$-mode matricization of $\mathcal{X}$. In \cite{7859390}, TT nuclear norm (TTNN) $\sum_{k=1}^{N-1} \alpha_k \|\textbf{X}_{[k]}\|_\ast$ was proposed to use  as a convex surrogate of TT rank for more convenient calculation and applied to low-rank tensor completion problem. Equiped with the TTNN, Chen $\mathit{et}$ $\mathit{al.}$ \cite{CHEN2021100} studied the RTC problem. As an improvement of Tucker rank, TT rank help us to study the correlation between the first $k$ modes (rather than one mode) and the rest modes. However, in real applications, the performance of TT decomposition is highly dependent on the permutations of the tensor dimensions.

TR decomposition \cite{TR} (as shown in Fig. \ref{decomposition} (b)) decomposes an $N$th-order tensor $\mathcal{X} \in \R^{I_1 \times I_2 \times \cdots \times I_N}$ into a circular multilinear product of a list of third-order core tensors, and the element-wise form is expressed as
\begin{equation}
\mathcal{X}(i_1,i_2,\cdots,i_N)=\sum_{r_1 =1}^{R_1} \sum_{r_2 =1}^{R_2}\cdots \sum_{r_N =1}^{R_N} \{\mathcal{G}_1(r_1,i_1,r_2)\mathcal{G}_2(r_2,i_2,r_3)\cdots\mathcal{G}_N(r_N,i_N,r_1)\}.
\end{equation}
TR rank is defined as a $N$-dimensional vector $(R_1,R_2,\cdots,R_N)$. Inspired by the connection between TR rank and the rank of circularly unfolding matrices, TR nuclear norm minimization (TRNNM)  \cite{yu2019tensor} $\sum_{k=1}^N \alpha_k \|\textbf{X}_{<k,L>} \|_\ast$ was suggested as a convex surrogate of TR rank for low-rank tensor completion problem, where $\textbf{X}_{<k,L>}\in \R^{\Pi_{i=k}^{k+L-1} I_i \times \Pi_{i=k+L}^{k-1} I_i}$ is the tensor circular unfolding matrix. Then, Huang $\mathit{et}$ $\mathit{al.}$ \cite{9136899} studied the TRNNM-based RTC problem. TR decomposition has generalized representation abilities since it can be viewed as the linear combination of TT decomposition. However, TR decomposition only connects the adjacent two factors, which is highly sensitive to the order of tensor modes.

In order to capture the relationships between any two factor tensors, Zheng $\mathit{et}$ $\mathit{al.}$ \cite{8606065} proposed the fully-connected tensor network (FCTN) decomposition (as shown in Fig. \ref{decomposition} (c)), which decomposes an $N$th-order tensor $\mathcal{X} \in \R^{I_1 \times I_2 \times \cdots \times I_N}$ into $N$ small-sized $N$th-order tensors, the element-wise form is expressed as
 \begin{equation}\label{dec}
 \begin{aligned}
&\mathcal{X}(i_1,i_2,\cdots,i_N)=\sum_{r_{1,2}=1}^{R_{1,2}}\sum_{r_{1,3}=1}^{R_{1,3}}\cdots \sum_{r_{1,N}=1}^{R_{1,N}}\sum_{r_{2,3}=1}^{R_{2,3}}\cdots\sum_{r_{2,N }=1}^{R_{2,N}} \cdots \sum_{r_{N-1,N}=1}^{R_{N-1,N}}
 \{\mathcal{G}_1 (i_1,r_{1,2},r_{1,3},\cdots,r_{1,N})
 \mathcal{G}_2 (r_{1,2},\\
 &i_2,r_{2,3},\cdots,r_{2,N}) \cdots \mathcal{G}_k (r_{1,k},r_{2,k},\cdots,r_{k-1,k},i_k,r_{k,k+1},\cdots,r_{k,N})
 \cdots \mathcal{G}_N (r_{1,N},r_{2,N},\cdots,r_{N-1,N},i_N)\}.
 \end{aligned}
 \end{equation}
FCTN rank is defined as vector $(R_{1,2},R_{1,3}, \cdots, R_{1,N}, R_{2,3}, R_{2,4},\cdots, R_{2,N}, \cdots, R_{N-1,N})$. Compared with other tensor decompositions, the FCTN decomposition obtains superior performance on the tensor completion problem. The reason is that it can flexibly characterize the correlation between arbitrary modes. In this paper, we leverage the strong expression ability of FCTN into the RTC problem. The contribution of this paper is threefold:

 (i) We firstly propose the FCTN nuclear norm as the convex surrogate of the FCTN rank. By applying the FCTN nuclear norm to the RTC problem, we suggest a FCTN-based robust convex optimization model (RC-FCTN). Secondly, we theoretically establish the exact recovery guarantee for the RC-FCTN. Finally, we introduce an alternating direction method of multipliers (ADMM)-based algorithm for solving the proposed RC-FCTN.

(ii) We propose a FCTN-based robust nonconvex optimization model (RNC-FCTN) for the RTC problem and develop a proximal alternating minimization (PAM)-based algorithm to solve the proposed model. Moreover, we theoretically derive the convergence of the PAM-based algorithm.

(iii) Extensive numerical experiments on several tasks, such as video completion and video background subtraction, demonstrate that the proposed methods are superior to several state-of-the-art approaches.

The outline of the paper is as follows. Section \ref{preliminaries} summarizes necessary preliminaries throughout the paper. Section \ref{convex} introduces the convex model RC-FCTN for the RTC problem and establishes the exact recovery guarantee of the RC-FCTN model and the convergence guarantee of the ADMM-based algorithm. Section \ref{nonconvex} presents the nonconvex model RNC-FCTN for the RTC problem and theoretically derives the convergence guarantee of the PAM-based algorithm. Section \ref{experiments} reports extensive numerical experiments to verify the superior performance of the proposed methods. Section \ref{conclusion} concludes this paper.

\section{Preliminaries}\label{preliminaries}
In this section, we summarize the necessary notations and several definitions used in this paper.

We use $x$, $\textbf{x}$, $\textbf{X}$, and $\mathcal{X}$ to denote scalars, vectors, matrices, and tensors, respectively. For tensor $\mathcal{X} \in \R^{I_1 \times I_2 \times \cdots \times I_N}$, we denote $\mathcal{X}(i_1,i_2,\cdots,i_N)$ as its $(i_1,i_2,\cdots,i_N)$th element. The inner product of two tensors $\mathcal{X}$ and $\mathcal{Y}$ with the same size is defined as the sum of the products of their entries, i.e., $\langle \mathcal{X}, \mathcal{Y} \rangle = \sum_{i_1,i_2,\cdots,i_N} \mathcal{X}(i_1,i_2,\cdots,i_N)\mathcal{Y}(i_1,i_2,\cdots,i_N)$. The $\mathit{l}_1$-norm and Frobenius norm of $\mathcal{X}$ are defined as $\|\mathcal{X}\|_1=\sum_{i_1,i_2,\cdots,i_N}|\mathcal{X}(i_1,i_2,\cdots,i_N)|$ and $\|\mathcal{X}\|_F=\sqrt{\sum_{i_1,i_2,\cdots,i_N}|\mathcal{X}(i_1,i_2,\cdots,i_N)|^2}$, respectively.

\begin {definition}(Generalized tensor transposition \cite{8606065})\label{perm}
For an $\mathit{N}$th-order tensor $\mathcal{X} \in \R^{I_1 \times I_2 \times \cdots \times I_N}$ and a specified rearrangement of vector $[1,2,\cdots,N]$, which is denoted as $\textbf{n}$, the $\textbf{n}$-based generalized tensor transposition of $\mathcal{X}$ is rearranging the modes of $\mathcal{X}$ by the specified vector $\textbf{n}$, denoted as $\vec{\mathcal{X}}^{\textbf{n}} \in \R^{I_{n_1} \times I_{n_2} \times \cdots \times I_{n_N}}$. The corresponding operation and its inverse operation are denoted as $\vec{\mathcal{X}}^{\textbf{n}} =\rm{permute}(\mathcal{X},\textbf{n})$ and $\mathcal{X}=\rm{ipermute}$$(\vec{\mathcal{X}}^{\textbf{n}},\textbf{n})$, respectively.
\end {definition}

\begin {definition}(Generalized tensor unfolding \cite{8606065})\label{resh}
For an $\mathit{N}$th-order tensor $\mathcal{X} \in \R^{I_1 \times I_2 \times \cdots \times I_N}$ and a specified rearrangement of vector $[1,2,\cdots,N]$, which is denoted as $\textbf{n}$, the generalized tensor unfolding of $\mathcal{X}$ is a matrix defined as $\textbf{X}_{[\textbf{n}_{1:d};\textbf{n}_{d+1:N}]}=\rm{reshape}$ $(\vec{\mathcal{X}}^{\textbf{n}},
\prod_{i=1}^d I_{n_i}, \prod_{i=d+1}^N I_{n_i})$. The corresponding inverse operation is defined as $\mathcal{X}=\Upsilon (\textbf{X}_{[\textbf{n}_{1:d};\textbf{n}_{d+1:N}]})$.
\end {definition}

\begin {definition}(Tensor contraction \cite{8606065})\label{contraction}
Suppose that $\mathcal{X} \in \R^{I_1 \times I_2 \times \cdots \times I_N}$ and $\mathcal{Y} \in \R^{J_1 \times J_2 \times \cdots \times J_M}$ have d modes of the same size ($I_{n_i}=J_{m_i}$ with $i=1,2,\cdots,d$), the tensor contraction along the $\textbf{n}_{1:d}$th-modes of $\mathcal{X}$ and the $\textbf{m}_{1:d}$th-modes of $\mathcal{Y}$
is an (N+M-2d)th-order tensor that satisfied
\begin{equation}
\mathcal{Z}=\mathcal{X} \times_{\textbf{n}_{1:d}}^{\textbf{m}_{1:d}} \mathcal{Y} \Leftrightarrow \textbf{Z}_{[1:N-d;N-d+1:N+M-2d]}=\textbf{X}_{[\textbf{n}_{d+1:N};\textbf{n}_{1:d}]}\textbf{Y}_
{[\textbf{m}_{1:d};\textbf{m}_{d+1:M}]}.
\end{equation}
\end {definition}

\begin {definition}(FCTN decomposition \cite{8606065})\label{de4}
An Nth-order tensor $\mathcal{X} \in \R^{I_1 \times I_2 \times \cdots \times I_N}$ can be decomposed into a series of Nth-order factor tensors $\mathcal{G}_k \in \R^{R_{1,k}\times R_{2,k}\times \cdots \times R_{k-1,k}\times I_k \times R_{k,k+1} \times \cdots \times R_{k,N}}, (k=1,2,\cdots,N)$, whose elements satisfied
 \begin{equation}\label{dec}
 \begin{aligned}
&\mathcal{X}(i_1,i_2,\cdots,i_N)=\sum_{r_{1,2}=1}^{R_{1,2}}\sum_{r_{1,3}=1}^{R_{1,3}}\cdots \sum_{r_{1,N}=1}^{R_{1,N}}\sum_{r_{2,3}=1}^{R_{2,3}}\cdots\sum_{r_{2,N }=1}^{R_{2,N}} \cdots \sum_{r_{N-1,N}=1}^{R_{N-1,N}} \{\mathcal{G}_1 (i_1,r_{1,2},r_{1,3},\cdots,r_{1,N})
 \mathcal{G}_2 (r_{1,2},\\
 &i_2,r_{2,3},\cdots,r_{2,N}) \cdots \mathcal{G}_k (r_{1,k},r_{2,k},\cdots,r_{k-1,k},i_k,r_{k,k+1},\cdots,r_{k,N})
 \cdots \mathcal{G}_N (r_{1,N},r_{2,N},\cdots,r_{N-1,N},i_N)\}.
 \end{aligned}
 \end{equation}
This decomposition is defined as the FCTN decomposition. The factors $\mathcal{G}_1$, $\mathcal{G}_2$, $\cdots$, and $\mathcal{G}_N$ are the core tensors of $\mathcal{X}$ and can be abbreviated as $\{\mathcal{G}\}_{1:N}$, then $\mathcal{X}=\rm{FCTN}(\{\mathcal{G}\}_{1:\mathit{N}})$. The vector $(R_{1,2}, R_{1,3},\cdots, R_{1,N},$ $R_{2,3},R_{2,4},\cdots, R_{2,N}, \cdots, R_{N-1,N})$ is defined as the FCTN-rank of the original tensor $\mathcal{X}$.
\end {definition}

\begin{lemma}\cite{8606065}
An Nth-order tensor $\mathcal{X} \in \R^{I_1 \times I_2 \times \cdots \times I_N}$ can be represented by (\ref{dec}), then we have
\begin{equation}
{\rm{Rank}} (\textbf{X}_{[\textbf{n}_{1:d};\textbf{n}_{d+1:N}]}) \leq \prod_{i=1}^d \prod_{j=d+1}^N R_{\textbf{n}_i,\textbf{n}_j} (R_{\textbf{n}_i,\textbf{n}_j}=R_{\textbf{n}_j,\textbf{n}_i}, {\text{if}} ~ \textbf{n}_i > \textbf{n}_j).
\end{equation}
\end{lemma}

\section{RC-FCTN model}\label{convex}
By utilizing the relationship between the rank of generalized tensor unfolding matrices and the FCTN rank, we suggest a new FCTN nuclear norm as a convex surrogate of FCTN rank to measure the correlation between arbitrary modes.

\begin {definition}(FCTN nuclear norm)\label{FCTNNN}
For an Nth-order tensor $\mathcal{X} \in \R^{I_1 \times I_2 \times \cdots \times I_N}$, its FCTN nuclear norm is defined as
\begin{equation}
\sum_{k=1}^{\bar{N}} \alpha_k \|\textbf{X}_{[\textbf{n}^k_{\textbf{1}};\textbf{n}^k_{\textbf{2}}]}\|_\ast,
\end{equation}
where $\textbf{n}^k$ is the k-th rearrangement of the vector $[1,2,\cdots,N]$, $\textbf{n}^k_{\textbf{1}}=\textbf{n}^k_{1:\lfloor N/2 \rfloor}$, $\textbf{n}^k_{\textbf{2}}=\textbf{n}^k_{\lfloor N/2 \rfloor+1:N}$, and $\bar{N}=C _N^{\lfloor N/2 \rfloor}$ \footnote{Since the order of the elements in vectors $\textbf{n}^k_{\textbf{1}}$ and $\textbf{n}^k_{\textbf{2}}$ does not effect the singular values of $\textbf{X}_{[\textbf{n}^k_{\textbf{1}};\textbf{n}^k_{\textbf{2}}]}$, $\bar{N}= C _N^{\lfloor N/2 \rfloor}$. If N is even, $\|\textbf{X}_{[\textbf{n}^k_{\textbf{1}};\textbf{n}^k_{\textbf{2}}]}\|_\ast=\|\textbf{X}_{[\textbf{n}^k_{\textbf{2}};\textbf{n}^k_{\textbf{1}}]}\|_\ast$, then $\bar{N}=C _N^{\lfloor N/2 \rfloor}/2$.}.
\end {definition}

Based on the proposed FCTN nuclear norm, we suggest a robust convex optimization model RC-FCTN for the RTC problems as follows:
\begin{equation}\label{eq1}
     \begin{aligned}
\min_{\mathcal{X},\mathcal{E}}&~~\sum_{k=1}^{\bar{N}} \alpha_k \|\textbf{X}_{[\textbf{n}^k_{\textbf{1}};\textbf{n}^k_{\textbf{2}}]}\|_\ast +\lambda \|\mathcal{E}\|_1,\\
\text{s.t. }&~~  \mathcal{P}_\Omega (\mathcal{X}+\mathcal{E})=\mathcal{P}_\Omega (\mathcal{O}).
    \end{aligned}
\end{equation}
where $\mathcal{O}$ is the observed data, $\mathcal{X}$ and $\mathcal{E}$ are the low-rank component and the sparse component, and $\lambda$ is regularization parameter.

We theoretically establish the exact recovery guarantee for the RC-FCTN. Firstly, we design a constructive singular value decomposition (SVD)-based FCTN decomposition that can decompose a large-scale tensor into the small-scale core tensors. Secondly, we propose the FCTN incoherence conditions on the core tensors. Finally, building on the FCTN incoherence conditions, we establish the theoretical guarantee of the exact recovery.

\subsection{SVD-based FCTN decomposition}
In \cite{8606065}, an optimal fixed FCTN-rank approximation of the given tensor can be derived by the alternating least squares method. However, it is not enough to study the exact recovery theory of the robust FCTN completion problem. Therefore, we propose a constructive SVD-based FCTN decomposition to compute the core tensors by sequential SVDs, which is a necessary decomposition to study the exact recovery guarantee.

For an $\mathit{N}$th-order tensor $\mathcal{X} \in \R^{I_1 \times I_2 \times \cdots \times I_N}$, the FCTN rank is $$(R_{1,2}, R_{1,3}, \cdots, R_{1,N}, R_{2,3}, R_{2,4},\cdots, R_{2,N}, \cdots, R_{N-1,N}).$$  Denote $\hat{R}_{i}=\Pi_{j=1,j\neq i}^{N}R_{ij}$, then it follows the results given in \cite{ye2019tensor}, the fully-connected graph based tensor network states can be categorized into three types: subcritical, critical and supercritical. If $\hat{R}_{i}\leq I_{i}(\hat{R}_{i}\geq I_{i}), \forall i=1,...,N$, where at least one inequality is strict, then it is called subcritical (supercritical), if $\hat{R}_{i}= I_{i}, \forall i=1,...,N$, it is critical. In this paper we mainly focus on the subcritical case, since a supercritical FCTN case can be reduced to the subcritical case by a surjective birational map \cite{ye2019tensor}.

\begin{figure}[htp]
\centering
\includegraphics[width=0.99\linewidth]{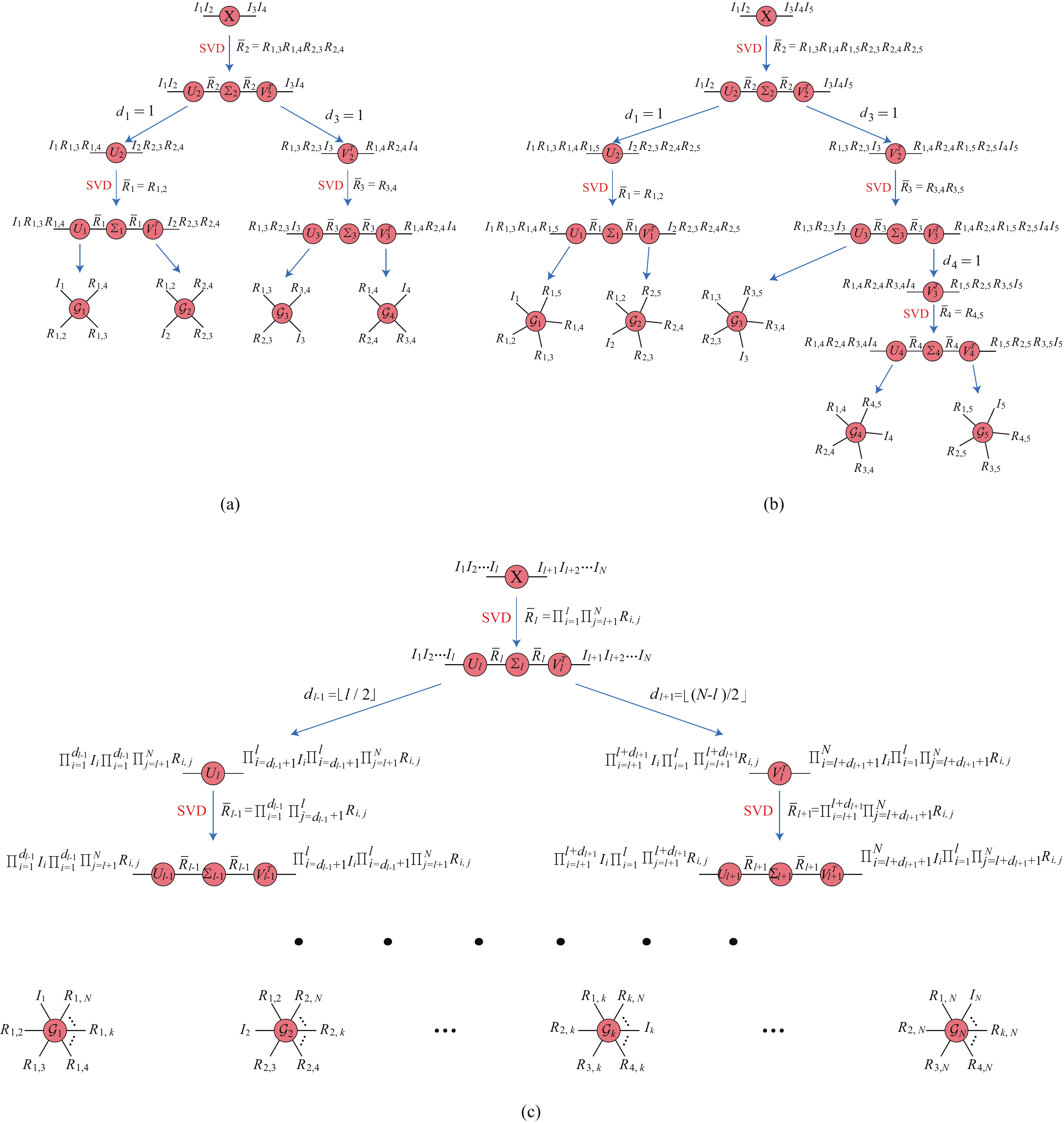}
  \caption{The graphical representation of the SVD-based FCTN decomposition.  (a), (b), and (c) are the graphical representation of the SVD-based FCTN decomposition of a fourth-order tensor, a fifth-order tensor, and an $\mathit{N}$th-order tensor, respectively.}\vspace{-0.4cm}
  \label{SVD-FCTN}
\end{figure}

Denote $d_{l}=l=\lfloor  N/2\rfloor$, $\textbf{n}$ is the rearrangement of vector ($1,2,\cdots,N$), the whole process of SVD-based FCTN decomposition is summarized in Algorithm \ref{FCTNSVD} \footnote{The SVD-based FCTN decomposition can be performed on arbitrary $\textbf{n}$-based generalized tensor transposition. Here, for simplicity, we just take the order $\textbf{n}=[1,2,\cdots,N]$ as an example.}. To facilitate its understanding, we show the  graphical representation of SVD-based FCTN decomposition of a fourth-order tensor, a fifth-order tensor, and an $\mathit{N}$th-order tensor in Fig. \ref{SVD-FCTN}.

\begin{algorithm}
\caption{SVD-based FCTN decomposition for tensor $\mathcal{X}$.}
\textbf{Input:} Tensor $\mathcal{X} \in \R^{I_{1}\times I_{2}\times \cdots \times I_{N}}$, the FCTN rank.\\
\textbf{Output:} Core tensors $\mathcal{G}^{k},k=1,...,N$ relate to the FCTN decomposition.\\
\textbf{1.}~~ Choose one unfolding matrix $\textbf{X}_{[\textbf{n}_{\textbf{1}};\textbf{n}_{\textbf{2}}]}$; \\
\textbf{2.}~~ Compute the rank $\bar{R}_l$ truncated SVD of $\textbf{X}_{[\textbf{n}_{\textbf{1}};\textbf{n}_{\textbf{2}}]}$: $\textbf{X}_{[\textbf{n}_{\textbf{1}};\textbf{n}_{\textbf{2}}]}=U_l \Sigma_l V_l^T+E$, where $d_l=l=\lfloor N/2 \rfloor$,

~~~~~~ $\bar{R}_l=\prod_{i=1}^l \prod_{j=l+1}^N R_{i,j}$, and $\mathit{E}$ is error matrix; \\
\textbf{3.}~~ Fold the matrix $U_l$ and $V_l$: $\mathcal{U}_l=\Upsilon(U_l)\in \C^{I_1 I_2 \cdots I_l \bar{R}_l}$, $\mathcal{V}_l=\Upsilon(V_l)\in \C^{I_{l+1} I_{l+2} \cdots I_N \bar{R}_l}$;\\
\textbf{4.}~~ Denote $d_{l-1}=\lfloor l/2 \rfloor$ and $d_{l+1}=\lfloor (N-l)/2 \rfloor$, unfold the tensor $\mathcal{U}_l$ and $\mathcal{V}_l$ with size $I_1 \prod_{i=l+1}^N R_{1,i}\cdots$

~~~~~~$ I_{d_{l-1}} \prod_{i=l+1}^N R_{d_{l-1},i} \times I_{d_{l-1}+1} \prod_{i=l+1}^N R_{d_{l-1}+1,i}\cdots I_l \prod_{i=l+1}^N, R_{l,i}$ and
$\prod_{i=1}^l R_{i,l+1} I_{l+1}\cdots \prod_{i=1}^l $

~~~~~~$R_{i,l+d_{l+1}} I_{l+d_{l+1}} \times \prod_{i=1}^l R_{i,l+d_{l+1}+1}I_{l+d_{l+1}+1} \cdots \prod_{i=1}^l R_{i,N} I_N$, respectively;\\
\textbf{5.}~~ Compute the rank $\bar{R}_{l-1}$  and rank $\bar{R}_{l+1}$ truncated SVD of new matrix $U_l$ and $V_l$, respectively:

~~~~~~$U_l=U_{l-1}\Sigma_{l-1}V_{l-1}^T+E$ and $V_l=U_{l+1}\Sigma_{l+1}V_{l+1}^T+E$, where $\bar{R}_{l-1}=\prod_{i=1}^{d_{l-1}} \prod_{j=d_{l-1}+1}^l R_{i,j}$ and

~~~~~~$\bar{R}_{l+1}=\prod_{i=l+1}^{l+d_{l+1}} \prod_{j=l+d_{l+1}+1}^N R_{i,j}$;\\
\textbf{6.}~~ Fold the matrix $U_{l-1}$, $V_{l-1}$, $U_{l+1}$, and $V_{l+1}$;

~~~~~~~~~~~~~~~~~~~~~~~~~~~~~~~~~~~~~~~~~~~~~~~~$\cdots$\\
\textbf{7.}~~~$\mathcal{G}_1=\Upsilon(U_1)$ with size $I_1 \times R_{1,2} \times R_{1,3} \times \cdots \times R_{1,N}$,

~~~~~~$\mathcal{G}_2=\Upsilon(U_2)$ with size $R_{1,2}  \times I_2\times R_{2,3} \times \cdots \times R_{2,N}$,

~~~~~~~~~~~~~~~~~~~~~~~~~~~~~~~~~~~~~~~~~~~~~~~~$\cdots$

~~~~~~$\mathcal{G}_k=\Upsilon(U_k)$ with size $R_{1,k}  \times R_{2,k} \times \cdots \times I_k   \times \cdots \times R_{k,N}$,

~~~~~~~~~~~~~~~~~~~~~~~~~~~~~~~~~~~~~~~~~~~~~~~~$\cdots$

~~~~~~$\mathcal{G}_{N}=\Upsilon(U_N)$ with size $R_{1,N}  \times R_{2,N} \times \cdots \times R_{N-1,N}   \times I_N$.
\label{FCTNSVD}
\end{algorithm}

The tensor SVD-based FCTN decomposition can be written as
\begin{equation}
\mathcal{X}=\rm{FCTN} (\{\hat{\mathcal{G}} \}_{1:\mathit{N}},\{\Sigma\}_{1:\mathit{N}-1})= \rm{FCTN} (\hat{\mathcal{G}}_{1},\Sigma_{1},\hat{\mathcal{G}}_{2},\Sigma_{2},\cdots,\Sigma_{\mathit{N}-1},
\hat{\mathcal{G}}_{\mathit{N}}).
\end{equation}
When the $\Sigma_{i}$ ($i=1$, $2$, $\cdots$, $N-1$) are absorbed in the factors $\hat{\mathcal{G}}_{i}$ and $\hat{\mathcal{G}}_{i+1}$, we can get the FCTN decomposition $\mathcal{X}=\rm{FCTN} (\{\mathcal{G}\}_{1:\mathit{N}})$.

\subsection{FCTN incoherence conditions and exact recovery  guarantee}
We firstly propose the FCTN incoherence conditions and then establish the exact recovery  guarantee.

\begin{theorem}\label{invar}
Suppose that $\mathcal{T}\in\R^{I_{1}\times I_{2}\times \cdots \times I_{N}}$ has a SVD-based FCTN decomposition
\begin{equation}\label{eq2}
\mathcal{T}=\rm{FCTN} (\{\hat{\mathcal{G}} \}_{1:\mathit{N}},\{\Sigma\}_{1:\mathit{N}-1})= \rm{FCTN} (\hat{\mathcal{G}}_{1},\Sigma_{1},\hat{\mathcal{G}}_{2},\Sigma_{2},...,\Sigma_{\mathit{N}-1},
\hat{\mathcal{G}}_{\mathit{N}}),
\end{equation}
with the FCTN-rank $(R_{1,2},R_{1,3},\cdots,R_{1,N},R_{2,3},R_{2,4},\cdots,R_{2,N},\cdots,R_{N-1 ,N}).$ Then, its $\textbf{n}$-based generalized tensor transposition $\vec{\mathcal{T}}^\textbf{n}$ can be decomposes as $$\vec{\mathcal{T }}^{\textbf{n}}=(\vec{\mathcal{G}}_{\textbf{n}_1}^{\textbf{n}},\vec{\Sigma }_1,\vec{\mathcal{G}}_{\textbf{n}_2}^{\textbf{n}},\vec{\Sigma }_2,
\cdots,\vec{\Sigma }_{N-1},\vec{\mathcal{G}}_{\textbf{n}_N}^{\textbf{n}}),$$
 FCTN-rank is $(R_{\textbf{n}_1,\textbf{n}_2},R_{\textbf{n}_1,\textbf{n}_3},\cdots,R_{\textbf{n}_1,\textbf{n}_N},
R_{\textbf{n}_2,\textbf{n}_3},R_{\textbf{n}_2,\textbf{n}_4},\cdots,R_{\textbf{n}_2,\textbf{n}_N},
\cdots,R_{\textbf{n}_{N-1},\textbf{n}_N})$, $R_{\textbf{n}_i,\textbf{n}_j}=R_{\textbf{n}_j,\textbf{n}_i},$ ${\text{if}} ~ \textbf{n}_i> \textbf{n}_j$.
\end{theorem}

Based on Theorem \ref{invar}, when we suppose that $\mathcal{T}$ satisfies the following FCTN incoherent conditions for the order $\textbf{n}=[1,2,\cdots,N]$, then for any rearrangement $\textbf{n}$, $\vec{\mathcal{T}}^{\textbf{n}}$ satisfies the FCTN incoherent conditions. Therefore, we just consider the order $\textbf{n}=[1,2,\cdots,N]$.

\begin{definition}[FCTN Incoherence Conditions]\label{newcoherence}
Suppose that $\mathcal{T}\in\R^{I_{1}\times I_{2}\times \cdots \times I_{N}}$ has a SVD-based FCTN decomposition (\ref{eq2}) with FCTN-rank $(R_{1,2},R_{1,3},\cdots,R_{1,N},R_{1,2},R_{2,3},\cdots,R_{2,N},\cdots,R_{1,N},R_{2,N},
\cdots,$ $R_{N-1 ,N}).$ Denote $l=\lfloor N/2 \rfloor$, $e_{il}$ is a base vector with suitable dimension, then $\mathcal{T}$ is said to satisfy the FCTN incoherence conditions if there exists parameters $\mu_i$ such that for any $\xi=\{\xi_{1},\cdots,\xi_{m}\}\subseteq \Delta=\{1,\cdots,i-1,i+1,\cdots,N\}, ~m\leq l$ such that

\begin{align}
\max_{l=1,\cdots,I_{i} \Pi^{N,k\in \Delta-\xi}_{k=1} R_{i,k}}\|(\hat{\textbf{G}}_i)_{[\xi;\Delta-\xi]}\cdot \textbf{e}_{il}\|^2_{F}&\leq \frac{\mu_i \Pi^{\xi_{m}}_{k=1} R_{i,k}}{I_{i} \Pi^{N,k\in \Delta-\xi}_{k=1} R_{i,k}},~~ i=1,2,\cdots,N. \label{eq3}
\end{align}
\end{definition}

Based on the tensor network contraction operation given in Definition \ref{contraction} and the tensor network incoherence conditions on core tensors given in Definition \ref{newcoherence}, we have the following results. For convenience, we denote $n_{(1)}^k=\Pi_{i=1}^l I_{\textbf{n}^k_i}$, $n_{(2)}^k=\Pi_{i=l+1}^N I_{\textbf{n}^k_i}$, $\bar{n}^k=\rm{max}$$(n_{(1)},n_{(2)})$, $\hat{n}^k=\rm{min}$$(n_{(1)},n_{(2)})$, and $r^k=\Pi_{i=1}^l \Pi_{j=l+1}^N R_{\textbf{n}^k_i,\textbf{n}^k_j} (R_{\textbf{n}^k_i,\textbf{n}^k_j}=R_{\textbf{n}^k_j,\textbf{n}^k_i}, if ~ \textbf{n}^k_i> \textbf{n}^k_j)$.

\begin{theorem}\label{TheoremM}
Let $\mathcal{X}\in \R^{I_{1}\times I_{2}\times \cdots \times I_{N}}$ be given with a FCTN decomposition as (\ref{eq2}). Suppose the incoherence conditions given in (\ref{eq3}) are satisfied. Assume that the observation set $\Omega$ is
uniformly distributed among all sets of cardinality $m = \rho \Pi_{k=1}^N I_k.$ Also suppose that each observed entry is independently corrupted with probability $\gamma$. Then, there exist universal constants $c_{k1}, c_{k2} > 0$ such that with probability at least $1 -\sum_{k=1}^{\bar{N}}c_{k1}(\bar{n}^k)^{-c_{k2}}$, the recovery of underlying tensor $\mathcal{X}_0$ with $\hat{\lambda} = \sum_{k=1}^{\bar{N}} \alpha_k /\sqrt{\rho \bar{n}^k}$ is exact, provided that
\begin{equation}\label{eq4}
r^{k} \leq \frac{c_r \hat{n}^k}{\mu^k (\log( \bar{n}^k))^2}\, \,\,\,
\textup{and} \,\,\, \,\gamma \leq c_\gamma  ,k=1,..., \bar{N}
\end{equation}
where $\mu^{k}=\max\{\Pi_{i=1}^l \mu_{\textbf{n}^k_i} \Pi_{i=1}^{l-1} \Pi_{j=i+1}^l R_{\textbf{n}^k_i,\textbf{n}^k_j}^2 ,\Pi_{i=l+1}^{N}\mu_{\textbf{n}^k_i} \Pi_{i=l+1}^{N-1} \Pi_{j=i+1}^N R_{\textbf{n}^k_i,\textbf{n}^k_j}^2\}$, $c_r$ and $c_\gamma$ are two positive constants.

\end{theorem}
{\bf Proof: }
 The proof can be split into two steps. Firstly, we prove that the arbitrary unfolding matrix in \eqref{eq1} satisfies the matrix incoherence conditions in \cite{1970395} when the original tensor satisfies the FCTN incoherence conditions given in \eqref{eq3}. Secondly, we prove that the pair $(\mathcal{X}_{0},\mathcal{E})$ derived from some convex optimization algorithms is the unique optimal solution to problem \eqref{eq1}.

Firstly, we use a fourth-order tensor $\mathcal{T}\in \R^{I_{1}\times I_{2}\times I_{3} \times I_{4}}$ as an example to show the process.
For simplicity, set $\mathbf{n}^1=[1,2,3,4]$ and $l=\lfloor N/2 \rfloor=2.$ Then it follows  the SVD-based FCTN decomposition algorithm we can get
$$\mathcal{T}={\rm{FCTN}}(\hat{\mathcal{G}}_{1},\Sigma_{1},\hat{\mathcal{G}}_{2},\Sigma_{2},
\hat{\mathcal{G}}_{3},\Sigma_{3},\hat{\mathcal{G}}_{4}),$$
and
$$\textbf{T}_{[1,2;3,4]}=\textbf{U}_2 \Sigma_{2}\textbf{V}^H_2,$$
where
$$\textbf{U}_2=\textbf{U}_{[1,2;3,4]}= {\text{perm\&resh}}( (\hat{\textbf{G}}_1)_{[1,3,4;2]}\Sigma_{1}(\hat{\textbf{G}}_2)_{[1;2,3,4]},~I_1 I_2
,~R_{1,3}R_{1,4}R_{2,3}R_{2,4}), $$ $$\textbf{V}_2=\textbf{V}_{[1,2;3,4]}={\text {perm\&resh}}((\hat{\textbf{G}}_3)_{[1,2,3;4]}\Sigma_{3}
(\hat{\textbf{G}}_4)_{[3;1,2,4]},~I_3 I_4,~R_{1,3}R_{1,4}R_{2,3}R_{2,4}),$$
 The $\text{perm\&resh}$ operation means that we first do the permute operation (Definition \ref{perm}) and then do the reshape operation (Definition \ref{resh}) with a specified dimension or size.

By the FCTN incoherence conditions given in \eqref{eq3}, and choosing $i=1$, $\xi=\{2\}$, we obtain
\begin{equation}
\max_{l=1,\cdots,I_{1} R_{1,3} R_{1,4}}\|(\hat{\textbf{G}}_1)_{[2;1,3,4]}\cdot \textbf{e}_{1l}\|^2_{F}\leq \frac{\mu_1  R_{1,2}}{I_{1} R_{1,3} R_{1,4}},
\end{equation}
Choosing $i=2$, $\xi=\{1\}$, we have
\begin{equation}
\max_{l=1,\cdots,I_{2} R_{2,3} R_{2,4}}\|(\hat{\textbf{G}}_2)_{[1;2,3,4]}\cdot \textbf{e}_{2l}\|^2_{F}\leq \frac{\mu_2  R_{1,2}} {I_{2} R_{2,3}           R_{2,4}}.
\end{equation}
Moreover, it follows the SVD-based FCTN decomposition method $\|\Sigma_1\|\leq \sqrt{R_{1,3} R_{1,4} R_{2,3} R_{2,4}}.$ Combine above and note that
\begin{equation}
\begin{aligned}
\|\textbf{U}_{[1,2;3,4]}\|_{\infty}&=\max_{i=1,\cdots,I_{1} R_{1,3} R_{1,4},~j=1,\cdots,I_{2} R_{2,3}R_{2,4}}((\hat{\textbf{G}}_1)_{[2;1,3,4]})_i^T \Sigma_{1}((\hat{\textbf{G}}_2)_{[1;2,3,4]})_{j} \\
&\leq \max_{i=1,\cdots,I_{1} R_{1,3} R_{1,4},~j=1,\cdots,I_{2} R_{2,3}R_{2,4}} \|((\hat{\textbf{G}}_1)_{[2;1,3,4]})_i^T\| \|\Sigma_{1}\| \|((\hat{\textbf{G}}_2)_{[1;2,3,4]})_{j}\|  \\
& \leq \frac{\sqrt{\mu_1 \mu_2}  R_{1,2}}{\sqrt{I_{1}I_{2}}}
\end{aligned}
\end{equation}
where $\textbf{G}_i$ is the $i$th column of the matrix $\textbf{G}$. Then
\begin{equation}\label{newinco1}
\max_{i=1,\cdots,I_{1}I_{2}}\|\textbf{U}_{[1,2;3,4]}^H \cdot \textbf{e}_{i}\|^2_{F}\leq \frac{\mu_1 \mu_2 R^2_{1,2} R_{1,3} R_{1,4} R_{2,3} R_{2,4}}{I_1 I_2}.
\end{equation}
Similarly, we can get
\begin{equation}\label{newinco2}
\max_{j=1,\cdots,I_{3}I_{4}}\|\textbf{V}_{[1,2;3,4]}^H \cdot \textbf{e}_{j}\|^2_{F}\leq \frac{\mu_3 \mu_4 R^2_{3,4} R_{1,3} R_{1,4} R_{2,3} R_{2,4}}{I_3 I_4}.
\end{equation}
Moreover,
\begin{align}
\|\textbf{U}_{[1,2;3,4]} \textbf{V}_{[1,2;3,4]}^H\|_{\infty}&\leq
\sqrt{\frac{\mu_1\mu_2 R^2_{1,2} R_{1,3} R_{1,4} R_{2,3} R_{2,4}}{I_1\cdot I_2}}\cdot \sqrt{ \frac{\mu_3\mu_4 R^2_{3,4} R_{1,3} R_{2,3} R_{1,4}R_{2,4}}{I_3\cdot I_4}}\nonumber\\
&=\frac{\sqrt{\mu_1\mu_2\mu_3\mu_4}R_{1,2}R_{3,4}R_{1,3} R_{1,4} R_{2,3} R_{2,4}}{\sqrt{I_{1}I_{2}I_{3}I_{4}}}.\label{newcor3}
\end{align}
Combine (\ref{newinco1}), \eqref{newinco2}, and \eqref{newcor3}, and set
\begin{align*}
\mu&=\max\{\mu_1\mu_2 R^2_{1,2},\mu_3\mu_4 R^2_{3,4},\mu_1\mu_2\mu_3\mu_4 R^2_{1,2}R^2_{3,4}R_{1,3} R_{1,4} R_{2,3} R_{2,4}\}\\
   &=\mu_1\mu_2 \mu_3 \mu_4 R^2_{1,2}R^2_{3,4}R_{1,3} R_{1,4} R_{2,3} R_{2,4},
\end{align*}
we can verify that the unfolding matrix $\textbf{T}_{[1,2;3,4]}$ satisfies the corresponding incoherence conditions in \cite{1970395}. Similarly, for any specified rearrangement $\textbf{n}^k$, we can also verify that the generalized unfolding matrix $\textbf{T}_{[\textbf{n}_{\textbf{1}}^k;\textbf{n}_{\textbf{2}}^k]}$ satisfies corresponding incoherence conditions in \cite{1970395}.

 For an $N$th-order tensor $\mathcal{T}\in \R^{I_1\times I_2 \times \cdots \times I_N}$, its SVD-based FCTN decomposition can be expressed as
 $$\mathcal{T}={\rm{FCTN}}(\{\hat{\mathcal{G}}\}_{1:N},\{\Sigma\}_{1:N-1})={\rm{FCTN}}(\hat{\mathcal{G}}_1, \Sigma_1,\hat{\mathcal{G}}_2,\cdots,\Sigma_{l-1},\hat{\mathcal{G}}_l,\Sigma_{l},\hat{\mathcal{G}}_{l+1},\cdots ,\hat{\mathcal{G}}_{N}).$$
 For simplicity, setting $\mathbf{n}^1=[1,2,\cdots,N]$, $l=\lfloor N/2 \rfloor$, $\mathbf{n}^1_{1}=[1,\cdots,l]$, and $\mathbf{n}^1_{2}=[l+1,\cdots,N],$ then,
 $$\textbf{T}_{[\textbf{n}_{1}^1;\textbf{n}_{2}^1]}=\textbf{U}_l\Sigma_{l}\textbf{V}_l^H.$$
For $\textbf{U}_l$,  we can get the following iteration expression
\begin{equation}
\begin{aligned}
  \textbf{U}_2&= {\text{perm\&resh}}( (\hat{\textbf{G}}_1)_{[1,3,\cdots,N;2]}\Sigma_{1}(\hat{\textbf{G}}_2)_{[1;2,3,\cdots,N]},~I_1 I_2 ,~\Pi_{i=3}^N R_{1,i} R_{2,i}),\\
  \hat{\textbf{U}}_2&= {\text{perm\&resh}}(\textbf{U}_2,~I_1 I_2 \Pi_{i=4}^N R_{1,i} R_{2,i},~R_{1,3} R_{2,3}),\\
  &~~~~~~~~~~~~~~~~~~~~~~~~~~~~~ \cdots, \\
  \textbf{U}_k&={\text{perm\&resh}} (\hat{\textbf{U}}_{k-1} \Sigma_{k-1} (\hat{\textbf{G}}_k)_{[1,\cdots,k-1;k,\cdots,N]},~\Pi_{i=1}^k I_i ,~ \Pi_{i=1}^k \Pi_{j=k+1}^N R_{i,j}), \\
  \hat{\textbf{U}}_k&={\text{perm\&resh}} (\textbf{U}_k,~\Pi_{i=1}^k I_i \Pi_{i=1}^k \Pi_{j=k+2}^N R_{i,j},~ \Pi_{i=1}^k R_{i,k+1}), \\
  & ~~~~~~~~~~~~~~~~~~~~~~~~~~~~~\cdots, \\
  \textbf{U}_l&={\text{perm\&resh}}(\hat{\textbf{U}}_{l-1} \Sigma_{l-1} (\hat{\textbf{G}}_l)_{[1,\cdots,l-1;l,\cdots,N]},~\Pi_{i=1}^l I_i,~ \Pi_{i=1}^l \Pi_{j=l+1}^N R_{i,j}).
\end{aligned}
\end{equation}
Recall the incoherence conditions in (\ref{eq3}), we have
\begin{equation}
\max_{l=1,\cdots,I_{1} \Pi_{i=3}^N R_{1,i}} \|(\hat{\textbf{G}}_1)_{[2;1,3,\cdots,N]}\cdot \textbf{e}_{1l}\|^2_{F}\leq \frac{\mu_1  R_{1,2}}{I_{1} \Pi_{i=3}^N R_{1,i}}
\end{equation}
and
\begin{equation}
\max_{l=1,\cdots,I_{k} \Pi_{i=k+1}^N R_{k,i}} \|(\hat{\textbf{G}}_k)_{[1,\cdots,k-1;k,\cdots,N]}\cdot \textbf{e}_{kl}\|^2_{F}\leq \frac{\mu_k  \Pi_{i=1}^{k-1} R_{1,i}} {I_{k} \Pi_{i=k+1}^N R_{k,i}}, k=2,\cdots,l.
\end{equation}
Moreover, it follows the SVD-based FCTN decomposition method $\|\Sigma_1\|\leq \sqrt{\Pi_{i=3}^N R_{1,i} R_{2,i}}.$ Combine above and we have
\begin{equation}
\begin{aligned}
\|\hat{\textbf{U}}_2\|_{\infty} \leq &\frac{\sqrt{\mu_1 \mu_2}R_{1,2}}{\sqrt{I_{1}I_{2}}}, \\
\max_{i=1,\cdots,I_{1}I_{2}}\|\textbf{U}_2^H \cdot \textbf{e}_{i}\|^2_{F}\leq & \frac{\mu_1 \mu_2 R^2_{1,2} \Pi_{i=3}^N R_{1,i} R_{2,i}}{I_1 I_2},\\
\|\hat{\textbf{U}}_k\|_{\infty} \leq &\frac{\sqrt{\Pi_{i=1}^k \mu_i} \Pi_{i=1}^{k-1} \Pi_{j=i+1}^k R_{i,j}}{\sqrt{\Pi_{i=1}^k I_i}}, ~~k=3,4,\cdots,l,\\
\max_{i=1,\cdots,\Pi_{i=1}^k I_i}\|\textbf{U}_k^H \cdot \textbf{e}_{i}\|^2_{F}\leq & \frac{\Pi_{i=1}^k \mu_i \Pi_{i=1}^{k-1} \Pi_{j=i+1}^k R_{i,j}^2 \Pi_{i=1}^k \Pi_{j=k+1}^N R_{i,j}}{\Pi_{i=1}^k I_i}, ~~k=3,4,\cdots,l.
\end{aligned}
\end{equation}
Then, we have
\begin{equation}\label{newinco5}
\max_{i=1,\cdots,\Pi_{i=1}^l I_i}\|\textbf{U}_l^H \cdot \textbf{e}_{i}\|^2_{F}\leq \frac{\Pi_{i=1}^l \mu_i \Pi_{i=1}^{l-1} \Pi_{j=i+1}^l R_{i,j}^2 r}{\Pi_{i=1}^l I_i},
\end{equation}
  \begin{equation}\label{newinco6}
\max_{j=1,\cdots,\Pi_{i=l+1}^N I_i}\|\textbf{V}_l^H \cdot \textbf{e}_{j}\|^2_{F}\leq \frac{\Pi_{i=1+1}^N \mu_i \Pi_{i=l+1}^{N-1} \Pi_{j=i+1}^N R_{i,j}^2 r} {\Pi_{i=l+1}^N I_i}.
\end{equation}
Moreover,
 \begin{align}
\|\textbf{U}_l \textbf{V}_l^H\|_{\infty}\leq
\sqrt{\frac{\Pi_{i=1}^N \mu_i \Pi_{i=1}^{l-1} \Pi_{j=i+1}^l R_{i,j}^2 \Pi_{i=l+1}^{N-1} \Pi_{j=i+1}^N R_{i,j}^2 r}{\Pi_{i=1}^N I_i}},\nonumber
\end{align}
where $r=\Pi_{i=1}^l \Pi_{j=l+1}^N R_{i,j}$. To sum up, the unfolding matrix $\textbf{T}_{[\textbf {n}^1_1;\textbf {n}^1_2]}$ satisfies the corresponding incoherence conditions in \cite{1970395}. Similarly, for any specified rearrangement $\textbf{n}^k$, we can also verify that the generalized unfolding matrix $\textbf{T}_{[\textbf{n}_{\textbf{1}}^k;\textbf{n}_{\textbf{2}}^k]}$ satisfies the corresponding incoherence conditions in \cite{1970395}.

Secondly, the convex model (\ref{eq1}) can be regarded as a convex combination of $\bar{N}$ robust matrix completion models. Invoking the Theorem 1.1 in \cite{1970395} and recalling that the incoherence conditions of the unfolding matrices that appeared in (\ref{eq1}) are satisfied, we can prove the Theorem \ref{TheoremM}. \hfill$\Box$

\subsection{ADMM-based algorithm for solving RC-FCTN}
For the convex optimization problem (\ref{eq1}), we develop an ADMM-based algorithm to solve it. By introducing auxiliary variables $\mathcal{L}_k$ $(k=1,2,\cdots,\bar{N})$, the problem (\ref{eq1}) can be rewritten as
\begin{equation}\label{eq5}
     \begin{aligned}
\min_{\mathcal{X},\mathcal{E},\mathcal{L}^k}&~~\sum_{k=1}^{\bar{N}} \alpha_k \|\textbf{L}_{k[\textbf{n}_{\textbf{1}}^k;\textbf{n}_{\textbf{2}}^k]}\|_\ast +\lambda \|\mathcal{S}\|_1+\Phi(\mathcal{Y})\\
\text{s.t. }&~~\mathcal{Y}=\mathcal{X}+\mathcal{E}, \mathcal{S}=\mathcal{E}, \mathcal{L}_k=\mathcal{X}, k=1,2,\cdots,\bar{N}, \\
    \end{aligned}
\end{equation}
where
\begin{equation}
\Phi(\mathcal{Y})=
     \begin{cases}
 0,& ~~\mathcal{P}_\Omega(\mathcal{Y}) =\mathcal{P}_\Omega(\mathcal{O}) \\
\infty,&~~  \text{otherwise}\\
    \end{cases}.
\end{equation}
The augmented Lagrangian function of (\ref{eq5}) is
\begin{equation}
     \begin{aligned}
L(\mathcal{X},\mathcal{E},\mathcal{S},\mathcal{Y},\mathcal{L}_k,\mathcal{Z}_k,&\mathcal{P},\mathcal{Q}) = \sum_{k=1}^{\bar{N}}\Big\{ \alpha_k \|\textbf{L}_{k[\textbf{n}_{\textbf{1}}^k;\textbf{n}_{\textbf{2}}^k]}\|_\ast + \< \mathcal{Z}_k,\mathcal{L}_k-\mathcal{X} \> + \dfrac {\mu_k}2 \|\mathcal{L}_k-\mathcal{X}\|_F^2 \Big\}+\lambda \|\mathcal{S}\|_1 \\
& +\< \mathcal{P}, \mathcal{Y}-\mathcal{X}-\mathcal{E}\> +\dfrac \gamma2 \| \mathcal{Y}-\mathcal{X}-\mathcal{E}\|_F^2 +\< \mathcal{Q}, \mathcal{S}-\mathcal{E}\>+\dfrac \sigma2 \|\mathcal{S}-\mathcal{E}\|_F^2+\Phi(\mathcal{Y}),
     \end{aligned}
\end{equation}
where $\mu_k$, $\gamma$, and $\sigma$ are penalty parameters, and $\mathcal{Z}_k$, $\mathcal{P}$, and $\mathcal{Q}$ are Lagrangian multipliers. According to the ADMM framework \cite{ADMM1}, $\mathcal{L}_k$, $\mathcal{S}$, $\mathcal{Y}$, $\mathcal{X}$, and $\mathcal{E}$ can be divided into two groups, and then the two groups of variables are updated alternately.
\begin{equation}
(\mathcal{L}_k^{t+1},~\mathcal{S}^{t+1},~\mathcal{Y}^{t+1})=\argmin_{\mathcal{L}_k,\mathcal{S},\mathcal{Y}} L(\mathcal{X}^t,\mathcal{E}^t,\mathcal{S},\mathcal{Y},\mathcal{L}_k,\mathcal{Z}_k^t,\mathcal{P}^t,\mathcal{Q}^t),
\end{equation}
and
\begin{equation}
(\mathcal{X}^{t+1},~\mathcal{E}^{t+1})=\argmin_{\mathcal{X},\mathcal{E}} L(\mathcal{X},\mathcal{E},\mathcal{S}^{t+1},\mathcal{Y}^{t+1}, \mathcal{L}_k^{t+1},\mathcal{Z}_k^t,\mathcal{P}^t,\mathcal{Q})^t.
\end{equation}

Now, we present more details of each subproblem.

1) Update $\mathcal{L}_k$: the $\mathcal{L}_k~ (k=1,2,\cdots,\bar{N})$ subproblem can be easily transformed into its equivalent formulation:
\begin{equation}
\begin{aligned}
\textbf{L}_k^{t+1} = &\argmin_{\mathcal{L}_k} \alpha_k \|\textbf{L}_{k[\textbf{n}_{\textbf{1}}^k;\textbf{n}_{\textbf{2}}^k]}\|_\ast +\langle \mathcal{Z}_k,\mathcal{L}_k-\mathcal{X} \rangle + \dfrac {\mu_k}2 \|\mathcal{L}_k-\mathcal{X}\|_F^2 \\
=&\argmin_{\textbf{L}_k} \alpha_k \|\textbf{L}_{k[\textbf{n}_{\textbf{1}}^k;\textbf{n}_{\textbf{2}}^k]}\|_\ast + \dfrac {\mu_k}2 \|\textbf{L}_{k[\textbf{n}_{\textbf{1}}^k;\textbf{n}_{\textbf{2}}^k]}-(\mathcal{X}^t- \dfrac {\mathcal{Z}_k^t}{\mu_k})_{[\textbf{n}_{\textbf{1}}^k;\textbf{n}_{\textbf{2}}^k]}\|_F^2,
\end{aligned}
\end{equation}
which has the closed-form solution
\begin{equation}\label{eq6}
\textbf{L}_{k[\textbf{n}_{\textbf{1}}^k;\textbf{n}_{\textbf{2}}^k]} ^{t+1}= \textbf{U} \Sigma_{\alpha_k/\mu_k} \textbf{V}^T,
\end{equation}
where $(\mathcal{X}^t-\dfrac {\mathcal{Z}_k^t}{\mu_k})_{[\textbf{n}_{\textbf{1}}^k;\textbf{n}_{\textbf{2}}^k]} = \textbf{U} \Sigma \textbf{V}^T$, $\Sigma_{\alpha_k/\mu_k} = \text{diag}(\text{max}(\Sigma_{r,r}-\alpha_k/\mu_k,0))$, and $\Sigma_{r,r}$ is the $\mathit{r}$th singular value of $\Sigma$. $\mathcal{L}_k^{t+1}= \Upsilon(\textbf{L}_k^{t+1})$. The computational complexity of updating $\textbf{L}_k$ is $\mathcal{O}\big(\sum_{i=1}^{\bar{N}} p_i q_i \tt{min}$ $(p_i,q_i)\big)$ ($p_i= \prod_{i=1}^l I_{\textbf{n}_i}$ and $q_i = \prod_{i=l+1}^N I_{\textbf{n}_i}$).

2) Update $\mathcal{S}$: the $\mathcal{S}$-subproblem is
\begin{equation}
\begin{aligned}
\mathcal{S}^{t+1} = &\argmin_{\mathcal{S}} \lambda \|\mathcal{S}\|_1+\langle \mathcal{Q}^t, \mathcal{S}-\mathcal{E}^t\rangle+\dfrac \sigma2 \|\mathcal{S}-\mathcal{E}^t\|_F^2 \\
=&\argmin_{\mathcal{S}} \lambda \|\mathcal{S}\|_1+\dfrac \sigma2 \|\mathcal{S}-\mathcal{E}^t+\dfrac {\mathcal{Q}^t}\sigma\|_F^2.
\end{aligned}
\end{equation}
It has the following closed-form solution:
\begin{equation}\label{eq7}
\mathcal{S}^{t+1}=\text{soft}_{\lambda/ \sigma}(\mathcal{E}^t- \dfrac {\mathcal{Q}^t}\sigma),
\end{equation}
where $\text{soft}_{\lambda/ \sigma}(.)$ denotes the soft shrinkage operator with threshold value $\lambda / \sigma$. The computational complexity of updating $\mathcal{S}$ is $\mathcal{O}(\prod_{i=1}^{N}I_i)$.

3) Update $\mathcal{Y}$: the $\mathcal{Y}$ subproblem is
\begin{equation}
\begin{aligned}
\mathcal{Y}^{t+1} = &\argmin_{\mathcal{Y}} \Phi(\mathcal{Y})+\langle \mathcal{P}^t, \mathcal{Y}-\mathcal{X}^t-\mathcal{E}^t\rangle +\dfrac \gamma2 \| \mathcal{Y}- \mathcal{X}^t-\mathcal{E}^t\|_F^2\\
=&\argmin_{\mathcal{Y}} \Phi(\mathcal{Y})+\dfrac \gamma2 \| \mathcal{Y}-\mathcal{X}^t-\mathcal{E}^t+ \dfrac {\mathcal{P}^t}\gamma\|_F^2.
\end{aligned}
\end{equation}
Therefore, $\mathcal{Y}^{t+1}$ is updated via the following steps:
\begin{equation}\label{eq8}
\mathcal{Y}^{t+1}=\mathcal{Y}^0_{\Omega}+(\mathcal{X}^t+\mathcal{E}^t-\dfrac {\mathcal{P}^t}\gamma)_{\Omega^C},
\end{equation}
where $\Omega^C$ denotes the complementary set of $\Omega$. The computational complexity of updating $\mathcal{Y}$ is $\mathcal{O}(\prod_{i=1}^{N}I_i)$.

4) Update $(\mathcal{X},~\mathcal{E})$: The $(\mathcal{X},~\mathcal{E})$ subproblem is a least squares problem
\begin{equation}\label{eq9}
\begin{aligned}
(\mathcal{X}^{t+1},~\mathcal{E}^{t+1})=\argmin_{\mathcal{X},\mathcal{E}}& \sum_{k=1}^{\bar{N}}\Big\{\langle \mathcal{Z}_k^t,\mathcal{L}_k^{t+1}-\mathcal{X} \rangle + \dfrac {\mu_k}2 \|\mathcal{L}_k^{t+1}-\mathcal{X}\|_F^2 \Big\} +\langle \mathcal{P}^t, \mathcal{Y}^{t+1}-\mathcal{X}-\mathcal{E}\rangle \\
&+\dfrac \gamma2 \| \mathcal{Y}^{t+1} -\mathcal{X}-\mathcal{E}\|_F^2+\langle \mathcal{Q}^t, \mathcal{S}^{t+1}-\mathcal{E}\rangle+\dfrac \sigma2 \|\mathcal{S}^{t+1}-\mathcal{E}\|_F^2 \\
=\argmin_{\mathcal{X},\mathcal{E}}& \sum_{k=1}^{\bar{N}}\Big\{\dfrac {\mu_k}2 \|\mathcal{L}_k^{t+1}-\mathcal{X}+\dfrac {\mathcal{Z}_k^t}{\mu_k}\|_F^2 \Big\}+\dfrac \gamma2 \| \mathcal{Y}^{t+1} -\mathcal{X}-\mathcal{E}+\dfrac {\mathcal{P}^t}\gamma\|_F^2 \\
&+\dfrac \sigma2 \|\mathcal{S}^{t+1}-\mathcal{E}+\dfrac {\mathcal{Q}^t}\sigma\|_F^2.
\end{aligned}
\end{equation}
The objective function of (\ref{eq9}) is represented by $F(\mathcal{X},\mathcal{E})$. Taking $\partial{F}/\partial{\mathcal{X}}=0$ and $\partial{F}/\partial{\mathcal{E}}=0$, we
have
\begin{equation}
(\sum_{k=1}^{\bar{N}} \mu_k+\gamma)\mathcal{X}+\gamma \mathcal{E}= \sum_{k=1}^{\bar{N}}\mu_k (\mathcal{L}_k^{t+1}+ \dfrac {\mathcal{Z}_k^t}{\mu_k})+ \gamma(\mathcal{Y}^{t+1}+ \dfrac {\mathcal{P}^t}\gamma)
\end{equation}
and
\begin{equation}
\gamma\mathcal{X}+ (\gamma+\sigma) \mathcal{E}=\gamma(\mathcal{Y}^{t+1}+\dfrac {\mathcal{P}^t}\gamma)
+\sigma(\mathcal{S}^{t+1}+ \dfrac {\mathcal{Q}^t}\sigma).
\end{equation}
Based on the Cramer's Rule, $\mathcal{X}$ and $\mathcal{E}$ can be exactly obtained as follows:
\begin{equation}\label{eq10}
\mathcal{X}^{t+1}= (\gamma \mathcal{N}^t-(\gamma+\sigma)\mathcal{M}^t)/(\gamma^2- \Big{(}\sum_{k=1}^{\bar{N}}\mu_k+\gamma\Big{)}(\gamma+\sigma))
\end{equation}
and
\begin{equation}\label{eq11}
\mathcal{E}^{t+1}=\Big{(} \gamma \mathcal{M}^t- \Big{(}\sum_{k=1}^{\bar{N}}\mu_k+\gamma \Big{)}\mathcal{N}^t \Big{)}/(\gamma^2- \Big{(}\sum_{k=1}^{\bar{N}}\mu_k+\gamma\Big{)}(\gamma+\sigma)),
\end{equation}
where $\mathcal{M}^t=\sum_{k=1}^{\bar{N}}\mu_k (\mathcal{L}_k^{t+1}+ \mathcal{Z}_k^t/\mu_k)+ \gamma(\mathcal{Y}^{t+1}+ \mathcal{P}^t/\gamma)$ and $\mathcal{N}^t=\gamma(\mathcal{Y}^{t+1}+\mathcal{P}^t/\gamma)
+\sigma(\mathcal{S}^{t+1}+ \mathcal{Q}^t/\sigma)$. The computational complexity of updating $\mathcal{X}$  and $\mathcal{E}$ is $\mathcal{O}(\prod_{i=1}^{N}I_i)$.

5) Update multipliers: the Lagrangian multipliers are updated as follows:
\begin{equation}\label{eq12}
\left \{
\begin{aligned}
&\mathcal{Z}_k^{t+1}=\mathcal{Z}_k^t +\delta \mu_k(\mathcal{X}^{t+1}-\mathcal{L}_k^{t+1}),~~ k=1,2,\cdots,\bar{N},\\
&\mathcal{P}^{t+1}=\mathcal{P}^t +\delta \gamma (\mathcal{Y}-\mathcal{X}-\mathcal{E}),\\
&\mathcal{Q}^{t+1}=\mathcal{Q}^t +\delta \sigma (\mathcal{E}- \mathcal{S}).
\end{aligned}
\right.
\end{equation}
where $\delta$ is the step length. The complexity of updating multipliers is $\mathcal{O}(\prod_{i=1}^{N}I_i)$.

The whole process of the ADMM-based algorithm for solving RC-FCTN is summarized in Algorithm \ref{FCTN1}.

\begin{algorithm}[h]
\caption {ADMM-based algorithm for solving RC-FCTN.}
\textbf{Input:} The observed tensor $\mathcal{Y}^0$, parameter $\lambda$.  \\
\textbf{Initialization:}  $t=0$, $\mathcal{X}^0$, $\mathcal{E}^0$, $\mathcal{S}^0$, $\mathcal{L}_k^0$, Lagrangian multiplies $\mathcal{Z}_k^0~(k=1,2,\cdots,\bar{N})$, $\mathcal{P}$, $\mathcal{Q}$, parameters $\mu_k~(k=1,2,\cdots,\bar{N})$, $\gamma$ and $\sigma$;  \\
\textbf{1:} ~~~~ Update $\mathcal{L}_k^{t+1}$ via (\ref{eq6}); \\
\textbf{2:} ~~~~ Update $\mathcal{S}^{t+1}$ via (\ref{eq7}); \\
\textbf{3:} ~~~~ Update $\mathcal{Y}^{t+1}$ via (\ref{eq8}); \\
\textbf{4:} ~~~~ Update $\mathcal{X}^{t+1}$ and $\mathcal{E}^{t+1}$ via (\ref{eq10}) and (\ref{eq11}); \\
\textbf{5:} ~~~~ Update the multiplies via (\ref{eq12});  \\
\textbf{6:} ~~~~ Check convergence criteria: $ \|\mathcal{X}^{t+1}-\mathcal{X}^t\|_F/ \|\mathcal{X}^t\|_F \leq \epsilon$;  \\
\textbf{7:} ~~~~ If the convergence criteria is not meet, set $t:=t+1$ and go to Step 1. \\
\textbf{Output:} The low-rank component $\mathcal{X}$ and sparse component $\mathcal{E}$.
\label{FCTN1}
\end{algorithm}

\begin {theorem}\label{Thet1}
Within the framework of ADMM \cite{ADMM2}, the sequence $\{\mathcal{L}_k^t,~ \mathcal{S}^t, ~\mathcal{Y}^t, ~\mathcal{X}^t, ~ \mathcal{E}^t\}_{t \in \mathbb{N}}$ obtained by Algorithm \ref{FCTN1}  converges to the global minimum point of the problem (\ref{eq1}).
\end {theorem}

\section{RNC-FCTN model}\label{nonconvex}
In this section, inspired by the superiority of FCTN decomposition, we formulate the following FCTN-based nonconvex optimization model for the RTC problem as follows:
\begin{equation}\label{eq13}
     \begin{aligned}
\min_{\mathcal{X},\mathcal{E},\{\mathcal{F}\}_{1:N}}&~~\dfrac 12\|\mathcal{X}-\rm{FCTN}(\{\mathcal{F}\}_{1:\mathit{N}})\|_F^2 +\lambda \|\mathcal{E}\|_1\\
\text{s.t. }&~~  \mathcal{P}_\Omega (\mathcal{X}+\mathcal{E})=\mathcal{P}_\Omega (\mathcal{O}),
    \end{aligned}
\end{equation}
where $\lambda$ is the tuning parameter. Thus, we can rewrite the problem (\ref{eq13}) as the following unconstraint problem
\begin{equation}\label{eq14}
\min_{\mathcal{X},\mathcal{E},\{\mathcal{F}\}_{1:N},\mathcal{Y}}\dfrac 12\|\mathcal{X}-\rm{FCTN}(\{\mathcal{F}\}_{1:\mathit{N}})\|_F^2 +\lambda \|\mathcal{E}\|_1+\dfrac \beta2 \|\mathcal{Y}-\mathcal{X} -\mathcal{E}\|_F^2+\Phi(\mathcal{Y}),
\end{equation}
where
\begin{equation}
\Phi(\mathcal{Y})=
     \begin{cases}
 0,& ~~\mathcal{P}_\Omega(\mathcal{Y}) =\mathcal{P}_\Omega(\mathcal{O}) \\
\infty,&~~  \text{otherwise}\\
    \end{cases},
\end{equation}
and $\beta$ is a penalty parameter.

\subsection{PAM-based algorithm for solving RNC-FCTN}
The objective function is not jointly convex for ($\mathcal{X}$, $\mathcal{E}$,$\{\mathcal{F}\}_{1:N}$, $\mathcal{Y}$), but is convex with respect to $\mathcal{X}$, $\mathcal{E}$, $\{\mathcal{F}\}_{1:N}$, and $\mathcal{Y}$, respectively. We employ the PAM framework to tackle the nonconvex optimization problem (\ref{eq14}). At each iteration, a single block of variables is optimized, while the remaining variables are fixed. Detailedly, the PAM-based algorithm is updated as the following iterative scheme:
\begin{equation}
\left \{
\begin{aligned}
\mathcal{F}_k^{t+1}=&\argmin_{\mathcal{F}_k} f(\{\mathcal{F}\}_{1:k-1}^{t+1},~\mathcal{F}_k, ~\{\mathcal{F}\}_{k+1:N}^t ,~\mathcal{X}^t, ~\mathcal{E}^t, ~\mathcal{Y}^t) +\dfrac \rho2\|\mathcal{F}_k-\mathcal{F}_k^t\|_F^2, ~~ k=1,~2,\cdots,~N,\\
\mathcal{X}^{t+1}=&\argmin_{\mathcal{X}} f(\{\mathcal{F}\}_{1:N}^{t+1},~\mathcal{X},~\mathcal{E}^t, ~\mathcal{Y}^t)+\dfrac \rho2\|\mathcal{X}-\mathcal{X}^t\|_F^2,\\
\mathcal{E}^{t+1}=&\argmin_{\mathcal{E}} f(\{\mathcal{F}\}_{1:N}^{t+1}, ~\mathcal{X}^{t+1},~\mathcal{E}, ~\mathcal{Y}^t) +\dfrac \rho2\|\mathcal{E}-\mathcal{E}^t\|_F^2,\\
\mathcal{Y}^{t+1}=&\argmin_{\mathcal{Y}} f(\{\mathcal{F}\}_{1:N}^{t+1}, ~\mathcal{X}^{t+1}, ~\mathcal{E}^{t+1}, ~\mathcal{Y}) +\dfrac \rho2\|\mathcal{Y}-\mathcal{Y}^t\|_F^2,
\end{aligned}
\right.
\end{equation}
where $f(\{\mathcal{F}\}_{1:N}, ~\mathcal{X}, ~\mathcal{E}, ~\mathcal{Y})$ is the objective function of (\ref{eq14}) and $\rho>0$ is a proximal parameter.

The corresponding details are as follows.

1) Update $\mathcal{F}_k$: the $\mathcal{F}_k$ $(k=1,2,\cdots,N)$ subproblems can be rewritten as
\begin{equation}\label{eq15}
\begin{aligned}
\mathcal{F}_k^{t+1}=&\argmin_{\mathcal{F}_k} \dfrac 12\|\mathcal{X}^t-\rm{FCTN}(\{\mathcal{F}\}_{1:\mathit{k}-1}^{\mathit{t}+1},~\mathcal{F}_\mathit{k},~\{\mathcal{F}\}_{\mathit{k}+1:\mathit{N}}^\mathit{t})\|_F^2+\dfrac \rho2\|\mathcal{F}_{\mathit{k}}-\mathcal{F}_\mathit{k}^\mathit{t}\|_F^2 \\
=&\argmin_{\mathcal{F}_k} \Big\{\dfrac 12\|\textbf{X}^t_{(k)}-(\textbf{F}_k)_{(k)} (\textbf{M}_k^t)_{[\textbf{m}_{1:N-1};\textbf{n}_{1:N-1}]}\|_F^2 +\dfrac \rho2 \|(\textbf{F}_k)_{(k)}- (\textbf{F}_k^t)_{(k)}\|_F^2 \Big\},
\end{aligned}
\end{equation}
where $\mathcal{M}_k^t = \rm{FCTN}(\{\mathcal{F}\}_{1:\mathit{k}-1}^ {\mathit{t}+1},~ \{\mathcal{F}\}_{\mathit{k}+1:\mathit{N}}^\mathit{t})$. The problem (\ref{eq15}) can be directly solved as
\begin{equation}\label{eq16}
(\textbf{F}_k^{t+1})_{(k)}= [\textbf{X}^t_{(k)}(\textbf{M}_k^t)_{[\textbf{n}_{1:N-1};\textbf{m}_{1:N-1}]}+\rho (\textbf{F}_k^t)_{(k)}][(\textbf{M}_k^t)_{[\textbf{m}_{1:N-1};\textbf{n}_{1:N-1}]}(\textbf{M}_k^t)
_{[\textbf{n}_{1:N-1};\textbf{m}_{1:N-1}]} +\rho \textbf{I}]^{-1}.
\end{equation}
To facilitate the calculation of complexity, we simply set $I_1=I_2=\cdots=I_N=I$ and the FCTN rank $R_{k_1,k_2}$ as the same value R. This step of calculation mainly includes tensor contraction, matrix multiplication, and matrix inversion. The computational complexity of updating $\textbf{F}_k$ is $\mathcal{O}(N\sum_{i=2}^{N} I^i R^{i(N-i)+i-1} +NI^{N-1}R^{2(N-1)}+NR^{3(N-1)})$.

2) Update $\mathcal{X}$: the $\mathcal{X}$ subproblem can be simplified as
\begin{equation}
\begin{aligned}
\mathcal{X}^{t+1}=&\argmin_{\mathcal{X}} \dfrac 12\|\mathcal{X}-\rm{FCTN}(\{\mathcal{F}\}_ {1:\mathit{N}}^{\mathit{t}+1})\|_F^2 + \dfrac \beta2 \|\mathcal{Y}^{\mathit{t}}-\mathcal{X} -\mathcal{E}^{\mathit{t}}\|_F^2+\dfrac \rho2\|\mathcal{X}-\mathcal{X}^{\mathit{t}}\|_F^2\\
=&\argmin_{\mathcal{X}} \dfrac {1+\beta +\rho}2 \|\mathcal{X}-\dfrac {\rm{FCTN}(\{\mathcal{F}\}_{\mathit{1:N}}^{\mathit{t}+1})+\beta(\mathcal{Y}^
\mathit{t}- \mathcal{E}^\mathit{t})+\rho \mathcal{X}^\mathit{t}}{1+\beta+\rho}\|_F^2,
\end{aligned}
\end{equation}
which is a least square problem and has the following closed-form solution:
\begin{equation}\label{eq17}
\mathcal{X}^{t+1}= \dfrac {\rm{FCTN}(\{\mathcal{F}\}_{\mathit{1:N }}^{\mathit{t}+1})+ \beta(\mathcal{Y}^
\mathit{t}- \mathcal{E}^\mathit{t})+\rho \mathcal{X}^\mathit{t}}{1+\beta+\rho}.
\end{equation}
The computational complexity of updating $\mathcal{X}$ is $\mathcal{O}(\sum_{i=2}^{N} I^i R^{i(N-i)+i-1})$.

3) Update $\mathcal{E}$: the $\mathcal{E}$ subproblem can be rewritten as
\begin{equation}
\begin{aligned}
\mathcal{E}^{t+1}=&\argmin_{\mathcal{E}} \lambda \|\mathcal{E}\|_1 +\dfrac \beta2 \|\mathcal{Y}^t-\mathcal{X}^{t+1} -\mathcal{E}\|_F^2 +\dfrac \rho2\|\mathcal{E}-\mathcal{E}^t\|_F^2\\
=&\argmin_{\mathcal{E}} \lambda \|\mathcal{E}\|_1 +\dfrac {\beta+\rho} 2 \|\mathcal{E}-\dfrac {\beta(\mathcal{Y}^t-\mathcal{X}^{t+1})+\rho \mathcal{E}^t}{\beta+ \rho}\|_F^2,
\end{aligned}
\end{equation}
which can be updated as follows
\begin{equation}\label{eq18}
\mathcal{E}^{t+1}=\rm{soft}(\dfrac {\beta(\mathcal{Y}^\mathit{t}-\mathcal{X}^{\mathit{t} +1})+\rho \mathcal{E}^\mathit{t}}{\beta+ \rho},\dfrac \lambda {\beta+ \rho}).
\end{equation}
The computational complexity of updating $\mathcal{E}$ is $\mathcal{O}(I^N)$.

4) Update $\mathcal{Y}$: the $\mathcal{Y}$ subproblem can be rewritten as
\begin{equation}
\begin{aligned}
\mathcal{Y}^{t+1}=&\argmin_{\mathcal{Y}} \Phi(\mathcal{Y})+\dfrac \beta2 \|\mathcal{Y}-\mathcal{X}^{t+1} -\mathcal{E}^{t+1}\|_F^2 +\dfrac \rho2\|\mathcal{Y}-\mathcal{Y}^t\|_F^2 \\
=&\argmin_{\mathcal{Y}} \Phi(\mathcal{Y}) + \dfrac{\beta+\rho}2\|\mathcal{Y}-\dfrac {\beta(\mathcal{X}^{t+1}+\mathcal{E}^{t+1})+\rho\mathcal{Y}^t}{\beta+ \rho}\|_F^2.
\end{aligned}
\end{equation}
The $\mathcal{Y}$ is updated via the following steps:
\begin{equation}\label{eq19}
\left \{
\begin{aligned}
\mathcal{Y}^{t+1/2}=&\dfrac {\beta(\mathcal{X}^{t+1}+\mathcal{E}^{t+1})+\rho\mathcal{Y}^t}{\beta+ \rho},\\
\mathcal{Y}^{t+1}=&(\mathcal{Y}^{t+1/2})_{\Omega^C}+\mathcal{Y}_\Omega.
\end{aligned}
\right.
\end{equation}
where $\Omega^C$ is the complementary set of $\Omega$. The computational complexity of updating $\mathcal{Y}$ is $\mathcal{O}(I^N)$.

\begin{algorithm}[h]
\caption {PAM-based algorithm for solving RNC-FCTN.}
\textbf{Input:} The observed tensor $\mathcal{Y}$, the maximal FCTN rank $R^{\rm{max}}$, and parameter $\lambda$.  \\
\textbf{Initialization:}  $t=0$, $\mathcal{X}^0$, $\mathcal{E}^0$, and $\mathcal{F }_k^0$, the original FCTN rank $R$, parameter $\beta$ and $\rho$;  \\
\textbf{1:} ~~~~ Update $\mathcal{F}_k^{t+1}$ via (\ref{eq16}); \\
\textbf{2:} ~~~~ Update $\mathcal{X}^{t+1}$ via (\ref{eq17}); \\
\textbf{3:} ~~~~ Update $\mathcal{E}^{t+1}$ via (\ref{eq18}); \\
\textbf{4:} ~~~~ Update $\mathcal{Y}^{t+1}$ via (\ref{eq19}); \\
\textbf{5:} ~~~~ If $\|\mathcal{X}^{t+1}-\mathcal{X}^t\|_F/\|\mathcal{X}^t\|_F \leq 10^{-2}$, set $R$=min$\{R+1, R^{\rm{max}}\}$ and expand $\mathcal{F}_k^{t+1}$; \\
\textbf{6:} ~~~~ Check convergence criteria: $ \|\mathcal{X}^{t+1}-\mathcal{X}^t\|_F/ \|\mathcal{X}^t\|_F \leq \epsilon$;  \\
\textbf{7:} ~~~~ If the convergence criteria is not meet, set $t:=t+1$ and go to Step 1. \\
\textbf{Output:} The low-rank part $\mathcal{X}$ and sparse part $\mathcal{E}$.
\label{FCTN2}
\end{algorithm}

\subsection{Convergence guarantee of PAM-based algorithm}
In this section, we establish the convergence guarantee of the PAM-based algorithm.

\begin {theorem}\label{The}
The sequence $\{\{\mathcal{F}\}_{1:N}^t, ~\mathcal{X}^t, ~\mathcal{E}^t, ~\mathcal{Y}^t\}_{t \in \mathbb{N}}$ obtained by Algorithm \ref{FCTN2} globally converges to a critical point of (\ref{eq13}).
\end {theorem}
{\bf{Proof:}}
To prove it, we mainly demonstrate that the proposed RNC-FCTN satisfies the following three conditions:

(1) $f(\{\mathcal{F}\}_{1:N},~\mathcal{X},~\mathcal{E},~\mathcal{Y})$ is a proper lower semi-continuous function.

(2) $f(\{\mathcal{F}\}_{1:N},~\mathcal{X},~\mathcal{E},~\mathcal{Y})$ satisfies the K-\L{} property \cite{KL} at each $\{\{\mathcal{F}\}^t_{1:N},~\mathcal{X}^t,~\mathcal{E}^t,~\mathcal{Y}^t \}$.

(3) The bounded sequence $\{\{\mathcal{F}\}_{1:N}^t,~\mathcal{X}^t,~\mathcal{E}^t,~\mathcal{Y}^t \}_{t \in \mathbb{N}}$ satisfies the sufficient decrease and relative error conditions.

For convenience, we rewrite the objective function as
\begin{equation}
f(\{\mathcal{F}\}_{1:N},~\mathcal{X},~\mathcal{E},~\mathcal{Y})= f_1 (\{\mathcal{F}\}_{1:N},~\mathcal{X}, ~\mathcal{E},~\mathcal{Y}) + f_2 (\mathcal{E}) + \Phi(\mathcal{Y}).
\end{equation}
where $f_1 (\{\mathcal{F}\}_{1:N},~\mathcal{X}, ~\mathcal{E},~\mathcal{Y}) = \dfrac 12\|\mathcal{X}-{\rm{FCTN}}(\{\mathcal{F}\}_{1:N})\|_F^2 +\dfrac \beta2 \|\mathcal{Y}-\mathcal{X} -\mathcal{E}\|_F^2$ and $f_2 (\mathcal{E})=\lambda \|\mathcal{E}\|_1$.

Thus, PAM-based algorithm is updated as the following iterative scheme:
\begin{equation}
\left \{
\begin{aligned}
\mathcal{F}_k^{t+1}=&\argmin_{\mathcal{F}_k} f(\{\mathcal{F}\}_{1:k-1}^{t+1},~\mathcal{F}_k, ~\{\mathcal{F}\}_{k+1:N}^t ,~\mathcal{X}^t, ~\mathcal{E}^t, ~\mathcal{Y}^t) +\dfrac \rho2\|\mathcal{F}_k-\mathcal{F}_k^t\|_F^2, ~~ k=1,~2,\cdots,~N,\\
\mathcal{X}^{t+1}=&\argmin_{\mathcal{X}} f(\{\mathcal{F}\}_{1:N}^{t+1},~\mathcal{X},~\mathcal{E}^t, ~\mathcal{Y}^t)+\dfrac \rho2\|\mathcal{X}-\mathcal{X}^t\|_F^2,\\
\mathcal{E}^{t+1}=&\argmin_{\mathcal{E}} f(\{\mathcal{F}\}_{1:N}^{t+1}, ~\mathcal{X}^{t+1},~\mathcal{E}, ~\mathcal{Y}^t) +\dfrac \rho2\|\mathcal{E}-\mathcal{E}^t\|_F^2,\\
\mathcal{Y}^{t+1}=&\argmin_{\mathcal{Y}} f(\{\mathcal{F}\}_{1:N}^{t+1}, ~\mathcal{X}^{t+1}, ~\mathcal{E}^{t+1}, ~\mathcal{Y}) +\dfrac \rho2\|\mathcal{Y}-\mathcal{Y}^t\|_F^2.
\end{aligned}
\right.
\end{equation}

Now, we prove that the three key conditions are holded respectively.

Firstly, it is easy to verify that $f_1$ is a $C^1$ function with locally Lipschitz continuous gradient, and  $f_2$ and $\Phi(\mathcal{Y})$ are proper and lower semi-continuous functions. Therefore, $f(\{\mathcal{F}\}_{1:N},~\mathcal{X},~\mathcal{E},~\mathcal{Y})$ is a proper lower semi-continuous function.

Secondly, since the semi-algebraic real-valued function satisfies the K-\L{} property \cite{KL}, we only need to illustrate that $f(\{\mathcal{F}\}_{1:N},~\mathcal{X},~\mathcal{E},~\mathcal{Y})$ is a semi-algebraic function. $f_1 (\{\mathcal{F}\}_{1:N},~\mathcal{X}, ~\mathcal{E},~\mathcal{Y})$, $f_2 (\mathcal{E})$, and $\Phi(\mathcal{Y})$ are the sum of  Frobenius norm, $\mathit{l}_1$-norm, and indicator function, respectively. It is easy to identity that they are semi-algebraic functions \cite{PAM}. As the sum of three  semi-algebraic functions, $f(\{\mathcal{F}\}_{1:N},~\mathcal{X},~\mathcal{E},~\mathcal{Y})$ is still a semi-algebraic function. Therefore, $f(\{\mathcal{F}\}_{1:N},~\mathcal{X},~\mathcal{E},~\mathcal{Y})$ satisfies the K-\L{} property at each $\{\{\mathcal{F}\}_{1:N}^t,~\mathcal{X}^t,~\mathcal{E}^t,~\mathcal{Y}^t \}$.

Thirdly, we prove that the bounded sequence $\{\{\mathcal{F}\}_{1:N}^t,~\mathcal{X}^t,~\mathcal{E}^t,~\mathcal{Y}^t \}_{t \in \mathbb{N}}$ satisfies the sufficient decrease and relative error conditions, respectively.

\begin{lemma}(\textbf{Sufficient decrease})\label{lemma1}
Suppose $\{\{\mathcal{F}\}_{1:N}^t,~\mathcal{X}^t,~\mathcal{E}^t,~\mathcal{Y}^t \}_{t \in \mathbb{N}}$ be the sequence obtained by Algorithm \ref{FCTN2}, then it satisfies
\begin{equation}\label{eq20}
\left \{
\begin{aligned}
&f(\{\mathcal{F}\}_{1:k}^{t+1},~\{\mathcal{F}\}_{k+1:N}^t,~\mathcal{X}^t,~\mathcal{E}^t, ~\mathcal{Y}^t) +\dfrac \rho2\|\mathcal{F}_k^{t+1}-\mathcal{F}_k^t\|_F^2 \leq f(\{\mathcal{F}\}_{1:k-1}^{t+1},~\{\mathcal{F}\}_{k:N}^t,~\mathcal{X}^t,~\mathcal{E}^t, ~\mathcal{Y}^t), \\
&f(\{\mathcal{F}\}_{1:N}^{t+1},~\mathcal{X}^{t+1},~\mathcal{E}^t, ~\mathcal{Y}^t)+\dfrac \rho2\|\mathcal{X}^{t+1}-\mathcal{X}^t\|_F^2 \leq f(\{\mathcal{F}\}_{1:N}^{t+1},~\mathcal{X}^t,~\mathcal{E}^t, ~\mathcal{Y}^t),\\
&f(\{\mathcal{F}\}_{1:N}^{t+1}, ~\mathcal{X}^{t+1},~\mathcal{E}^{t+1}, ~\mathcal{Y}^t) +\dfrac \rho2\|\mathcal{E}^{t+1}-\mathcal{E}^t\|_F^2 \leq f(\{\mathcal{F}\}_{1:N}^{t+1},~\mathcal{X}^{t+1},~\mathcal{E}^t, ~\mathcal{Y}^t),\\
&f(\{\mathcal{F}\}_{1:N}^{t+1}, ~\mathcal{X}^{t+1}, ~\mathcal{E}^{t+1}, ~\mathcal{Y}^{t+1}) +\dfrac \rho2\|\mathcal{Y}^{t+1}-\mathcal{Y}^t\|_F^2 \leq f(\{\mathcal{F}\}_{1:N}^{t+1}, ~\mathcal{X}^{t+1},~\mathcal{E}^{t+1}, ~\mathcal{Y}^t).
\end{aligned}
\right.
\end{equation}
\end{lemma}

We give the proof of Lemma \ref{lemma1}. Let $\{\mathcal{F}\}_{1:N}^{t+1}$, $\mathcal{X}^{t+1}$, $\mathcal{E}^{t+1}$, and $\mathcal{Y}^{t+1}$ are the optimal solutions of $\mathcal{F}_k$-subproblem, $\mathcal{X}$-subproblem, $\mathcal{E}$-subproblem, and $\mathcal{Y}$-subproblem, we have
\begin{equation}
\begin{aligned}
&f(\{\mathcal{F}\}_{1:k}^{t+1},~\{\mathcal{F}\}_{k+1:N}^t,~\mathcal{X}^t,~\mathcal{E}^t, ~\mathcal{Y}^t) +\dfrac \rho2\|\mathcal{F}_k^{t+1}-\mathcal{F}_k^t\|_F^2 \leq f(\{\mathcal{F}\}_{1:k-1}^{t+1},~\{\mathcal{F}\}_{k:N}^t,~\mathcal{X}^t,~\mathcal{E}^t, ~\mathcal{Y}^t),\\
&f(\{\mathcal{F}\}_{1:N}^{t+1},~\mathcal{X}^{t+1},~\mathcal{E}^t, ~\mathcal{Y}^t)+\dfrac \rho2\|\mathcal{X}^{t+1}-\mathcal{X}^t\|_F^2 \leq f(\{\mathcal{F}\}_{1:N}^{t+1},~\mathcal{X}^t,~\mathcal{E}^t, ~\mathcal{Y}^t),\\
&f(\{\mathcal{F}\}_{1:N}^{t+1}, ~\mathcal{X}^{t+1},~\mathcal{E}^{t+1}, ~\mathcal{Y}^t) +\dfrac \rho2\|\mathcal{E}^{t+1}-\mathcal{E}^t\|_F^2 \leq f(\{\mathcal{F}\}_{1:N}^{t+1},~\mathcal{X}^{t+1},~\mathcal{E}^t, ~\mathcal{Y}^t),\\
&f(\{\mathcal{F}\}_{1:N}^{t+1}, ~\mathcal{X}^{t+1}, ~\mathcal{E}^{t+1}, ~\mathcal{Y}^{t+1}) +\dfrac \rho2\|\mathcal{Y}^{t+1}-\mathcal{Y}^t\|_F^2 \leq f(\{\mathcal{F}\}_{1:N}^{t+1}, ~\mathcal{X}^{t+1},~\mathcal{E}^{t+1}, ~\mathcal{Y}^t).
\end{aligned}
\end{equation}

\begin{lemma}(\textbf{Relative error})\label{lemma2}
Suppose $\{\{\mathcal{F}\}_{1:N}^t,~\mathcal{X}^t,~\mathcal{E}^t,~\mathcal{Y}^t \}_{t \in \mathbb{N}}$ be the sequence obtained by Algorithm \ref{FCTN2}, then there exists $\mathcal{U}_k^{t+1} \in \textbf{0}$ $(k=1,~2,\cdots,~N)$, $\mathcal{V}_1^{t+1} \in \textbf{0}$, $\mathcal{V}_2^{t+1} \in \partial f_2(\mathcal{E}^{t+1})$, and $\mathcal{V}_3^{t+1} \in \partial \Phi(\mathcal{Y}^{t+1})$, such that
\begin{equation}
\left \{
\begin{aligned}
&\|\mathcal{U}_k^{t+1} + \nabla_{\mathcal{F}_k} f_1(\{\mathcal{F}\}_{1:k}^{t+1}, ~\{\mathcal{F}\}_{k+1:N}^t,~\mathcal{X}^t, ~\mathcal{E}^t,~\mathcal{Y}^t)\|_F \leq \rho \|\mathcal{F}_k^{t+1}-\mathcal{F}_k^t\|_F,~k=1,~2,\cdots,~N, \\
&\|\mathcal{V}_1^{t+1} +\nabla_{\mathcal{X}} f_1(\{\mathcal{F}\}_{1:N}^{t+1},~\mathcal{X}^{t+1}, ~\mathcal{E}^t,~\mathcal{Y}^t) \|_F \leq \rho \|\mathcal{X}^{t+1}-\mathcal{X}^t\|_F, \\
&\|\mathcal{V}_2^{t+1} +\nabla_{\mathcal{E}} f_1(\{\mathcal{F}\}_{1:N}^{t+1}, ~\mathcal{X}^{t+1}, ~\mathcal{E}^{t+1}, ~\mathcal{Y}^t)  \|_F \leq \rho \|\mathcal{E}^{t+1}-\mathcal{E}^t\|_F, \\
&\|\mathcal{V}_3^{t+1} +\nabla_{\mathcal{Y}} f(\{\mathcal{F}\}_{1:N}^{t+1}, ~\mathcal{X}^{t+1}, ~\mathcal{E}^{t+1}, ~\mathcal{Y}^{t+1}) \|_F \leq \rho \|\mathcal{Y}^{t+1}-\mathcal{Y}^t\|_F.
\end{aligned}
\right.
\end{equation}
\end{lemma}

To prove Lemma \ref{lemma2}, we first show that the sequence is bounded. Since the initial tensors $\{\mathcal{F}\}_{1:N}^0$, $\mathcal{X}^0$, $\mathcal{E}^0$, and $\mathcal{Y}^0$ are  apparently bounded, we prove that $\{\mathcal{F}\}_{1:N}^{t+1}$, $\mathcal{X}^{t+1}$, $\mathcal{E}^{t+1}$, and $\mathcal{Y}^{t+1}$ are bounded when $\{\mathcal{F}\}_{1:N}^t$, $\mathcal{X}^t$, $\mathcal{E}^t$, and $\mathcal{Y}^t$ are bounded.

(I) The sequence $\{\{\mathcal{F}\}_{1:N}^t\}_{t \in \mathbb{N}}$ are bounded: Supposing that $\|\mathcal{F}_k^t\|_F \leq a$ and $\|\mathcal{X}^t\|_F \leq b$, according to (\ref{eq16}), we have
\begin{equation}
\begin{aligned}
\|\mathcal{F}_1^{t+1}\|_F &\leq (\|\mathcal{X}^t\|_F \|\mathcal{M}_1^t\|_F +\rho \|\mathcal{F}_1^t\|_F )\|(\textbf{N}_1^t+\rho \textbf{I})^{-1}\|_F \\
&\leq (ba^{N-1} + \rho a)\sqrt{\sum_{i=1}^j (1/(\sigma_i+\rho))^2} \\
& \leq (ba^{N-1} + \rho a) \sqrt{j}/\rho,
\end{aligned}
\end{equation}
where $\textbf{N}_1^t= (\textbf{M}_k^t)_{[\textbf{m}_{1:N-1};\textbf{n}_{1:N-1}]}(\textbf{M}_k^t)
_{[\textbf{n}_{1:N-1};\textbf{m}_{1:N-1}]}$ and $\sigma_i$ is the eigenvalues of $\textbf{N}_1^t$. It is clearly to see that $\mathcal{F}_1^{t+1}$ is bounded. Similarly, we can obtain that $\mathcal{F}_2^{t+1},~\mathcal{F}_3^{t+1},\cdots,~\mathcal{F}_N^{t+1}$ are bounded.

(II) The sequence $\{\mathcal{X}^t\}_{t \in \mathbb{N}}$ is bounded: Supposing that $\|\mathcal{F}_k^{t+1}\|_F \leq c$, $\|\mathcal{Y}^t\|_F \leq d$, and $\|\mathcal{E}^t\|_F \leq e$, according to (\ref{eq17})
\begin{equation}
\begin{aligned}
\|\mathcal{X}^{t+1}\|_F &=\| {\rm{FCTN}(\mathcal{F}_{\mathit{1:N}}^{\mathit{t}+1}) +\beta(\mathcal{Y}^\mathit{t}- \mathcal{E}^\mathit{t}) +\rho \mathcal{X}^\mathit{t}}\|_F /(1+\beta+\rho) \\
& \leq (\|\rm{FCTN}(\mathcal{F}_{\mathit{1:N}}^{\mathit{t}+1})\|_F + \beta \|\mathcal{Y}^\mathit{t}\|_F + \beta \|\mathcal{E}^\mathit{t}\|_F +\rho \|\mathcal{X}^\mathit{t}\|_F )/(1+\beta+\rho) \\
& \leq (c^N+\beta d + \beta e +\rho b)/(1+\beta+\rho).
\end{aligned}
\end{equation}
Thus, $\mathcal{X}^{t+1}$ is bounded.

(III) The sequence $\{\mathcal{E}^t\}_{t \in \mathbb{N}}$ is bounded: Since $\mathcal{X}^{t+1}$, $\{\mathcal{F}\}_{1:\mathit{N}}^{t+1}$, $\mathcal{Y}^t$, and $\mathcal{E}^t$ are bounded, we suppose
 that $\|\mathcal{X}^{t+1}-\rm{FCTN}(\{\mathcal{F}\}_{1:\mathit{N}}^{t+1})\|_F \leq f$, $\|\mathcal{Y}^t-\mathcal{X}^{t+1} -\mathcal{E}^t\|_F \leq g$, and $\|\mathcal{E}^t\|_1 \leq h$, according to (\ref{eq20}), we have
\begin{equation}
\begin{aligned}
f_2(\mathcal{E}^{t+1})=\lambda \|\mathcal{E}^{t+1}\|_1 &\leq f(\{\mathcal{F}\}_{1:N}^{t+1},~\mathcal{X}^{t+1},~\mathcal{E}^t, ~\mathcal{Y}^t) \\
&=\dfrac 12\|\mathcal{X}^{t+1}-\rm{FCTN}(\{\mathcal{F}\}_{1:\mathit{N}}^{t+1})\|_F^2 +\dfrac \beta2 \|\mathcal{Y}^t-\mathcal{X}^{t+1} -\mathcal{E}^t\|_F^2+ \lambda \|\mathcal{E}^t\|_1+\Phi (\mathcal{Y}^t) \\
& \leq \dfrac 12 f^2 +\dfrac \beta2 g^2 +\lambda h.
\end{aligned}
\end{equation}
Therefore, $\mathcal{E}^{t+1}$ is bounded.

(IV) The sequence $\{\mathcal{Y}^t\}_{t \in \mathbb{N}}$ is bounded: Supposing that $\|\mathcal{X}^{t+1}\|_F \leq i $ and $\|\mathcal{E}^{t+1}\|_F \leq j$, according to (\ref{eq19}), we have
\begin{equation}
\begin{aligned}
\|\mathcal{Y}^{t+1}\|_F &\leq \|(\mathcal{Y}^{t+1/2})_{\Omega^C}\|_F + \|\mathcal{Y}_\Omega\|_F \\
& \leq \|\beta(\mathcal{X}^{t+1}+\mathcal{E}^{t+1})+\rho\mathcal{Y}^t\|_F /(\beta+ \rho)+\rho \|\mathcal{Y}^t\|_F \\
& \leq \beta(i+j)/(\beta+ \rho) + \rho d/(\beta+ \rho)+ \rho d.
\end{aligned}
\end{equation}
Thus, $\mathcal{Y}^{t+1}$ is bounded.

In summary, the sequence $\{\{\mathcal{F}\}_{1:N}^t,~\mathcal{X}^t,~\mathcal{E}^t,~\mathcal{Y}^t \}_{t \in \mathbb{N}}$ is bounded.

Let $\{\mathcal{F}\}_{1:N}^{t+1}$, $\mathcal{X}^{t+1}$, $\mathcal{E}^{t+1}$, and $\mathcal{Y}^{t+1}$ are the optimal solutions of $\mathcal{F}_k$-subproblem, $\mathcal{X}$-subproblem, $\mathcal{E}$-subproblem, and $\mathcal{Y}$-subproblem, we have
\begin{equation}
\left \{
\begin{aligned}
&\textbf{0} \in \nabla_{\mathcal{F}_k} f_1(\{\mathcal{F}\}_{1:k-1}^{t+1},~\mathcal{F}_k, ~\{\mathcal{F}\}_{k+1:N}^t ,~\mathcal{X}^t, ~\mathcal{E}^t, ~\mathcal{Y}^t) +\rho ( \mathcal{F}_k -\mathcal{F}_k^t), ~~ k=1,~2,\cdots,~N, \\
&\textbf{0} \in \nabla_{\mathcal{X}} f_1(\{\mathcal{F}\}_{1:N}^{t+1},~\mathcal{X}, ~\mathcal{E}^t, ~\mathcal{Y}^t) +\rho (\mathcal{X}-\mathcal{X}^t), \\
&\textbf{0} \in \nabla_{\mathcal{E}} f_1(\{\mathcal{F}\}_{1:N}^{t+1},~\mathcal{X}^{t+1}, ~\mathcal{E}, ~\mathcal{Y}^t)+\partial f_2(\mathcal{E}) +\rho (\mathcal{E}-\mathcal{E}^t),\\
&\textbf{0} \in \nabla_{\mathcal{Y}} f_1(\{\mathcal{F}\}_{1:N}^{t+1},~\mathcal{X}^{t+1}, ~\mathcal{E}^{t+1}, ~\mathcal{Y})+ \partial \Phi(\mathcal{Y}) +\rho (\mathcal{Y}-\mathcal{Y}^t).
\end{aligned}
\right.
\end{equation}
Then, we define $\mathcal{U}_k^{t+1}$, $\mathcal{V}_1^{t+1}$, $\mathcal{V}_2^{t+1}$, and $\mathcal{V}_3^{t+1}$ as
\begin{equation}
\left \{
\begin{aligned}
&\mathcal{U}_k^{t+1} = -\nabla_{\mathcal{F}_k} f_1(\{\mathcal{F}\}_{1:k}^{t+1}, ~\{\mathcal{F}\}_{k+1:N}^t ,~\mathcal{X}^t, ~\mathcal{E}^t, ~\mathcal{Y}^t)-\rho ( \mathcal{F}_k^{t+1} -\mathcal{F}_k^t) \in \textbf{0}, \\
&\mathcal{V}_1^{t+1} = -\nabla_{\mathcal{X}} f_1(\{\mathcal{F}\}_{1:N}^{t+1},~\mathcal{X}^{t+1}, ~\mathcal{E}^t, ~\mathcal{Y}^t) -\rho (\mathcal{X}^{t+1}-\mathcal{X}^t) \in \textbf{0}, \\
&\mathcal{V}_2^{t+1} = -\nabla_{\mathcal{E}} f_1(\{\mathcal{F}\}_{1:N}^{t+1},~\mathcal{X}^{t+1}, ~\mathcal{E}^{t+1}, ~\mathcal{Y}^t)-\rho (\mathcal{E}^{t+1}-\mathcal{E}^t)\in  \partial f_2(\mathcal{E}^{t+1}),\\
&\mathcal{V}_3^{t+1} = - \nabla_{\mathcal{Y}} f_1(\{\mathcal{F}\}_{1:N}^{t+1},~\mathcal{X}^{t+1}, ~\mathcal{E}^{t+1}, ~\mathcal{Y}^{t+1}) -\rho (\mathcal{Y}^{t+1}-\mathcal{Y}^t)\in \partial \Phi(\mathcal{Y}^{t+1}).
\end{aligned}
\right.
\end{equation}
Since the sequence $\{\{\mathcal{F}\}_{1:N}^t,~\mathcal{X}^t,~\mathcal{E}^t,~\mathcal{Y}^t \}_{t \in \mathbb{N}}$ is bounded, and $\nabla f_1$ is Lipschitz continuous on any bounded set. Then there exists $\mathcal{U}_k^{t+1} \in \textbf{0}$ $(k=1,~2,\cdots,~N)$, $\mathcal{V}_1^{t+1} \in \textbf{0}$, $\mathcal{V}_2^{t+1} \in \partial f_2(\mathcal{E}^{t+1})$, and $\mathcal{V}_3^{t+1} \in \partial \Phi(\mathcal{Y}^{t+1})$, such that
\begin{equation}
\begin{aligned}
&\|\mathcal{U}_k^{t+1} + \nabla_{\mathcal{F}_k} f_1(\{\mathcal{F}\}_{1:k}^{t+1}, ~\{\mathcal{F}\}_{k+1:N}^t,~\mathcal{X}^t, ~\mathcal{E}^t,~\mathcal{Y}^t)\|_F \leq \rho \|\mathcal{F}_k^{t+1}-\mathcal{F}_k^t\|_F,~k=1,~2,\cdots,~N, \\
&\|\mathcal{V}_1^{t+1} +\nabla_{\mathcal{X}} f_1(\{\mathcal{F}\}_{1:N}^{t+1},~\mathcal{X}^{t+1}, ~\mathcal{E}^t,~\mathcal{Y}^t) \|_F \leq \rho \|\mathcal{X}^{t+1}-\mathcal{X}^t\|_F, \\
&\|\mathcal{V}_2^{t+1} +\nabla_{\mathcal{E}} f_1(\{\mathcal{F}\}_{1:N}^{t+1}, ~\mathcal{X}^{t+1}, ~\mathcal{E}^{t+1}, ~\mathcal{Y}^t)  \|_F \leq \rho \|\mathcal{E}^{t+1}-\mathcal{E}^t\|_F, \\
&\|\mathcal{V}_3^{t+1} +\nabla_{\mathcal{Y}} f(\{\mathcal{F}\}_{1:N}^{t+1}, ~\mathcal{X}^{t+1}, ~\mathcal{E}^{t+1}, ~\mathcal{Y}^{t+1}) \|_F \leq \rho \|\mathcal{Y}^{t+1}-\mathcal{Y}^t\|_F.
\end{aligned}
\end{equation}

Combining these conditions, the proposed algorithm conforms to Theorem 6.2 in \cite{PAM1}, the bounded sequence $\{\{\mathcal{F}\}_{1:N}^t,~\mathcal{X}^t,~\mathcal{E}^t,~\mathcal{Y}^t \}_{t \in \mathbb{N}}$ converges to the critical point of $f(\{\mathcal{F}\}_{1:N},~\mathcal{X},~\mathcal{E},~\mathcal{Y})$. \hfill$\Box$

\section{Numerical experiments}\label{experiments}

 \renewcommand{\arraystretch}{1.55}
\begin{table}[h]
  \centering
  \small
  \caption{Exact recovery on random synthetic data in different cases.}
  \vspace{0.2cm}
 \setlength{\tabcolsep}{6.6pt}
\begin{tabular}{c|c|c|c|c|c|c|c|c|c}
\Xhline{1.2pt}
\multicolumn{1}{l|}{size $I$} & rank $r$ &$ \rho$ &$ s$ &$\|\mathcal{X}-\mathcal{X}^0\|_F/\|\mathcal{X}^0\|_F $& size $I$ & rank  $r$ & $ \rho$                 &$ s$& $\|\mathcal{X}-\mathcal{X}^0\|_F/\|\mathcal{X}^0\|_F $ \\ \Xhline{1.2pt}
\multirow{8}{*}{20}       & \multirow{4}{*}{0.1$I$} & \multirow{2}{*}{1}   & 0.05 &    $1.28 \times 10^{-4}$& \multirow{8}{*}{40} & \multirow{4}{*}{0.1$I$} & \multirow{2}{*}{1}   & 0.05 &$1.06 \times 10^{-4}$    \\ \cline{4-5} \cline{9-10}
  &                       &                      & 0.1  &$1.56 \times 10^{-4}$    &                   &                   &                      & 0.1  & $1.12 \times 10^{-4}$ \\ \cline{3-5} \cline{8-10}
  &                       & \multirow{2}{*}{0.9} & 0.05 &$1.93 \times 10^{-4}$    &                   &                   & \multirow{2}{*}{0.9} & 0.05 &$1.72 \times 10^{-4}$    \\ \cline{4-5} \cline{9-10}
  &                       &                      & 0.1  &$1.96 \times 10^{-4}$    &                   &                   &                      & 0.1  &$2.25 \times 10^{-4}$    \\ \cline{2-5} \cline{7-10}
  & \multirow{4}{*}{0.2$I$} & \multirow{2}{*}{1}   & 0.05 &$4.74 \times 10^{-4}$     &   & \multirow{4}{*}{0.2$I$} & \multirow{2}{*}{1}   & 0.05 &$1.77 \times 10^{-4}$    \\ \cline{4-5} \cline{9-10}
  &                       &                      & 0.1  &$7.83 \times 10^{-4}$&                   &                   &                      & 0.1  &$2.88 \times 10^{-4}$    \\ \cline{3-5} \cline{8-10}
  &                       & \multirow{2}{*}{0.9} & 0.05 &$6.55 \times 10^{-4}$&                   &                   & \multirow{2}{*}{0.9} & 0.05 &$2.66 \times 10^{-4}$    \\ \cline{4-5} \cline{9-10}
  &                       &                      & 0.1  &$8.35 \times 10^{-4}$&                   &                   &                      & 0.1  &$3.01 \times 10^{-4}$     \\
\Xhline{1.2pt}
\end{tabular}
\label{tab51}
\end{table}

In this section, we firstly conduct the RTC experiments on synthetic data in subsection \ref{rsy}, which further corroborates our theoretical results. In subsection \ref{rvideo}-\ref{rback}, we conduct numerical experiments on color videos and hyperspectral videos (HSV) to verify the effectiveness of the proposed RC-FCTN and RNC-FCTN. To adequately examine the recovery performance of RC-FCTN and RNC-FCTN, we compare the proposed methods with four representative RTC methods: Tucker rank based method \cite{2512321} (denoted as ``SNN''), tubal rank based method \cite{8606165} (denoted as ``TNN''), TT rank based method \cite{YANG2020124783} (denoted as ``TTNN''), and TR rank based method \cite{9136899} (denoted as ``RTRC'').

For all methods in all experiments of subsection \ref{rvideo}-\ref{rback}, we employ the following setup to make a fair comparison: (i) The data is normalized into $[0, ~1]$; (ii) The relative error is set to $10^{-4}$; (iii) We utilize a simple linear interpolation strategy \cite{502} to obtain the $\mathcal{Y}^0$. We employ the mean of peak signal-to-noise rate (MPSNR) and the mean of structural similarity (MSSIM) as the initial tensor quantitative metric \cite{8854307}. The parameters in compared methods are manually adjusted to the optimal performance, which refers to the discussion in their articles. Meanwhile, the best and the second-best results are highlighted by bold and underline, respectively.

\subsection{Synthetic tensor completion}\label{rsy}
 \renewcommand{\arraystretch}{1.45}
 \begin{table}[h]\small
  \centering
  \caption{The quantitative comparison of different methods on the color videos $\mathit{bunny}$ and $\mathit{elephants}$.}
  \begin{threeparttable}
  \vspace{0.2cm}
  \setlength{\tabcolsep}{4pt}
    \begin{tabular}{ccccccccccc}
       \Xhline{1.2pt}
     Color&Salt and & \multirow{2}[0]{*}{SR} & \multirow{2}[0]{*}{Indicators} & \multirow{2}[0]{*}{Noise} & \multirow{2}[0]{*}{SNN} & \multirow{2}[0]{*}{TNN} & \multirow{2}[0]{*}{TTNN} & \multirow{2}[0]{*}{RTRC}&\multirow{2}[0]{*}{RC-FCTN} &\multirow{2}[0]{*}{RNC-FCTN}\\
      video& pepper noise&  &  &  &  &   &  & & & \\
        \Xhline{1.2pt}
      & &\multirow{2}[0]{*}{0.6} &MPSNR&9.3569&26.796&34.613&32.860&34.051&                                       \underline{34.923}&\textbf {39.715} \\
      & &  &MSSIM&0.0948& 0.8323& 0.9554&0.9658&0.9624& \underline{0.9751}&\textbf{0.9834} \\
      \cline{3-11}
     \multirow{2}[0]{*}{$\mathit{bunny}$}& \multirow{2}[0]{*}{0.1}  &\multirow{2}[0]{*}{0.4} &MPSNR&8.0350&22.614& 30.594&29.795&30.873& \underline{31.856}&\textbf{37.105} \\
       & &  &MSSIM&0.0621& 0.6530& 0.9248&0.9371&0.9241& \underline{0.9455}&\textbf{0.9774} \\
         \cline{3-11}
      & &\multirow{2}[0]{*}{0.2} &MPSNR&7.0242&14.924&25.674&23.895&\underline{26.599}& 26.564&\textbf{32.847} \\
      &  &  &MSSIM&0.0328& 0.4154& 0.7738&0.7643&0.8349& \underline{0.8366}&\textbf{0.9411} \\
      \cline{3-11}
     &&& time (s)&&93.317&26.842&60.408&58.017&128.60&1586.1\\
     \Xhline{1.2pt}
     & &\multirow{2}[0]{*}{0.6} &MPSNR&7.1457&28.801&34.348&32.614&33.692& \underline{36.987}&\textbf {39.178} \\
      & &  &MSSIM&0.0494& 0.8634& 0.9576&0.9427&0.9605& \underline{0.9743}&\textbf{0.9748} \\
      \cline{3-11}
     \multirow{2}[0]{*}{$\mathit{elephants}$}& \multirow{2}[0]{*}{0.1}  &\multirow{2}[0]{*}{0.4} &MPSNR&5.6305&24.214& 31.045&29.745&30.517& \underline{33.246}&\textbf{36.203} \\
       & &  &MSSIM&0.0312& 0.7587& 0.9279&0.9329&0.9292& \underline{0.9555}&\textbf{0.9597} \\
         \cline{3-11}
      & &\multirow{2}[0]{*}{0.2} &MPSNR&4.5164&14.449&26.280&25.416&26.312&  \underline{27.936}&\textbf{29.913} \\
       &  &  &MSSIM&0.0169& 0.5156& 0.8110&0.8446&0.8471& \underline{0.8773}&\textbf{0.8882} \\
       \cline{3-11}
     &&&time (s)&&157.58&25.866&79.171&59.713&120.34&1731.5\\
     \Xhline{1.2pt}
    \end{tabular}%
    \end{threeparttable}
  \label{tab:addlabel1}%
\end{table}%

To verify the validity of exact recovery in Theorem \ref{TheoremM}, we firstly execute experiments on synthetic data. We simulate the tensor of size $I\times I \times I \times I$ by FCTN contraction with varying dimensions $I$=20 and 40. We generate the tensor $\mathcal{X}^0={\rm{FCTN}}(\mathcal{G}_1,\mathcal{G}_2,\mathcal{G}_3,\mathcal{G}_4)$ with FCTN rank $R_{i,j}~(1 \leq i <j \leq 4, ~{\text{and}} ~i,j \in \mathbb{N}^+)$ as the same value $r$. The core tensors are independently satisfied the uniform distribution $\mathcal{U}(0,1)$. Then, the whole entries are corrupted by salt and pepper (SaP) noise with density $s$. And we choose the observation with sampling ratio (SR) $\rho$. We test the recovery ability of the proposed RC-FCTN on eight cases and show the results in Table \ref{tab51}. We design $r=0.1I$ and $0.2I$, $\rho$ = 0.9 and 0.8, and $s$ = 0.05 and 0.1. As observed, our method obtains the negligible relative error $\|\mathcal{X}-\mathcal{X}^0\|_F/\|\mathcal{X}^0\|_F$. These numerical results of all cases validly corroborate the exact recovery in Theorem \ref{TheoremM} well.

\subsection{Color video completion}\label{rvideo}
To demonstrate the effectiveness of the proposed methods, we execute experiments on two color videos \footnote{The data is available at http://trace.eas.asu.edu/yuv/.} (height $\times$  width $\times$ color channel $\times$ frames) including $\mathit{bunny}$ and $\mathit{elephants}$. We consider RTC problem for the testing color videos with SaP=0.1 and different SRs $\{0.6, 0.4, 0.2\}$.

\begin{figure}[!t]
\footnotesize
\setlength{\tabcolsep}{0.97pt}
\begin{center}
\begin{tabular}{cccccccc}
Observed&SNN&TNN&TTNN&RTRC&RC-FCTN&RNC-FCTN& Ground truth \\
\includegraphics[width=0.124\textwidth]{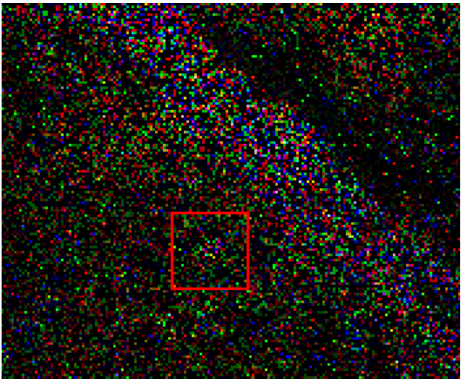}&
\includegraphics[width=0.124\textwidth]{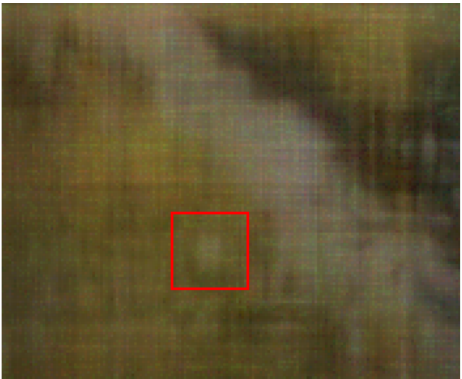}&
\includegraphics[width=0.124\textwidth]{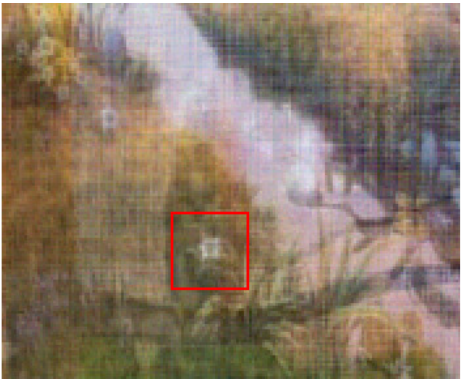}&
\includegraphics[width=0.124\textwidth]{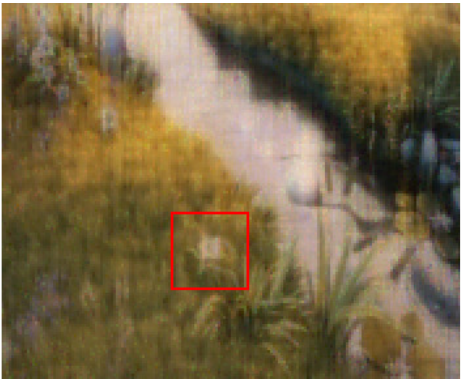}&
\includegraphics[width=0.124\textwidth]{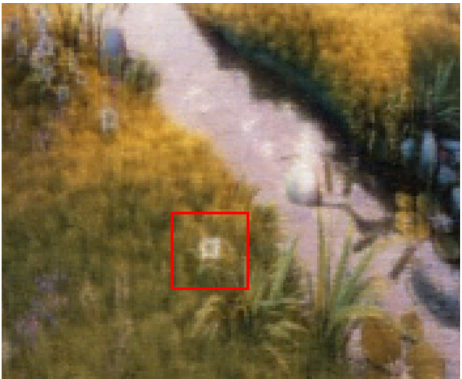}&
\includegraphics[width=0.124\textwidth]{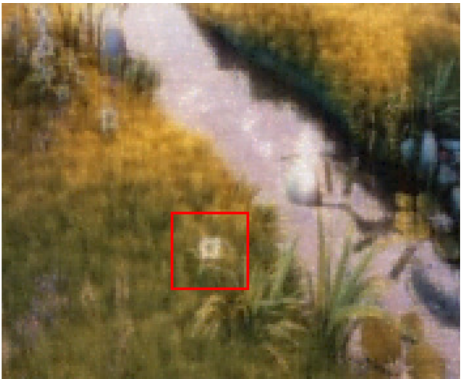}&
\includegraphics[width=0.124\textwidth]{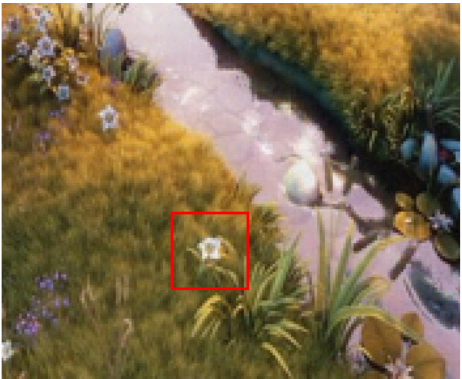}&
\includegraphics[width=0.124\textwidth]{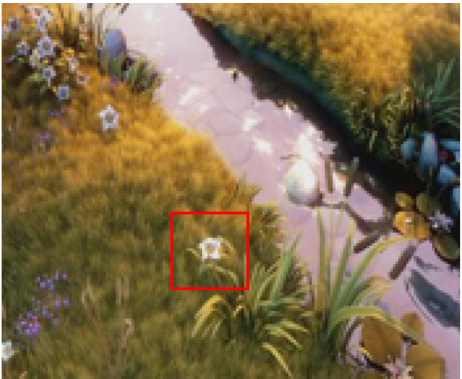}
\\
\includegraphics[width=0.124\textwidth]{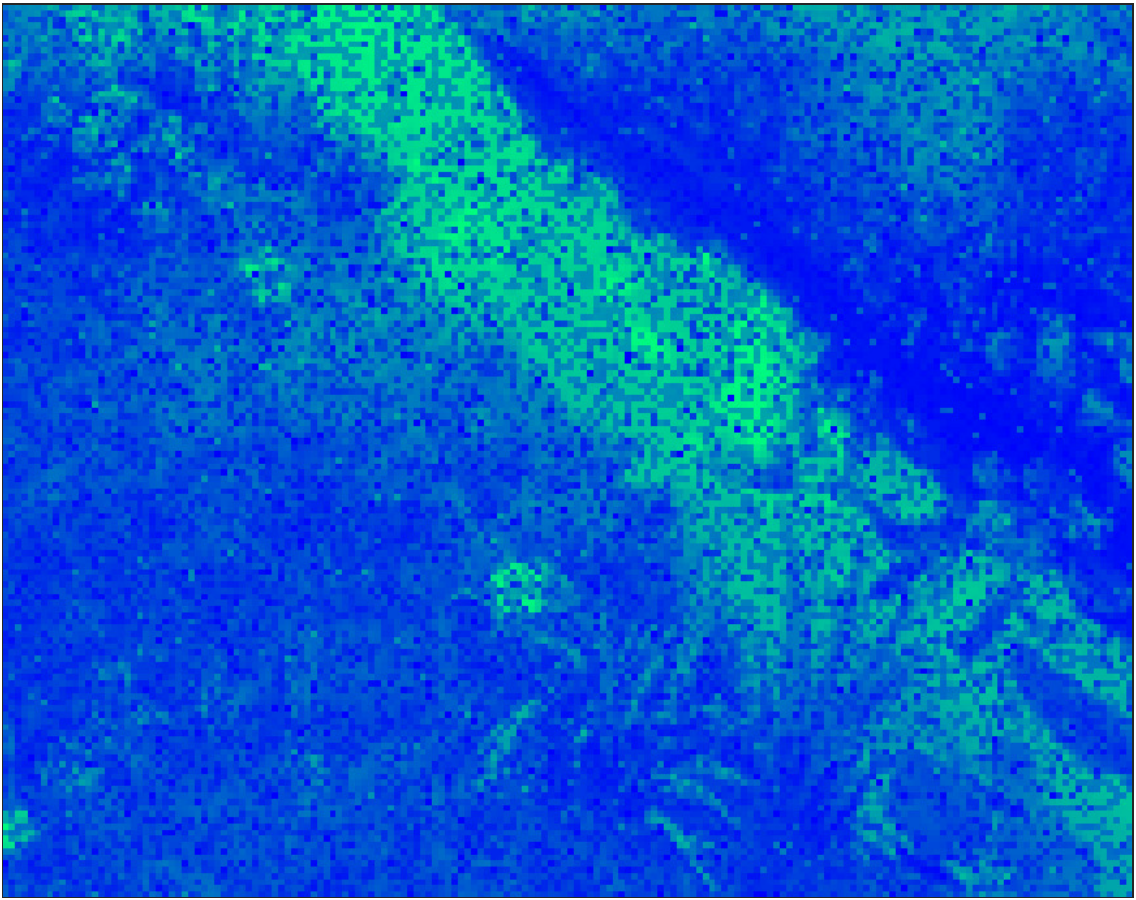}&
\includegraphics[width=0.124\textwidth]{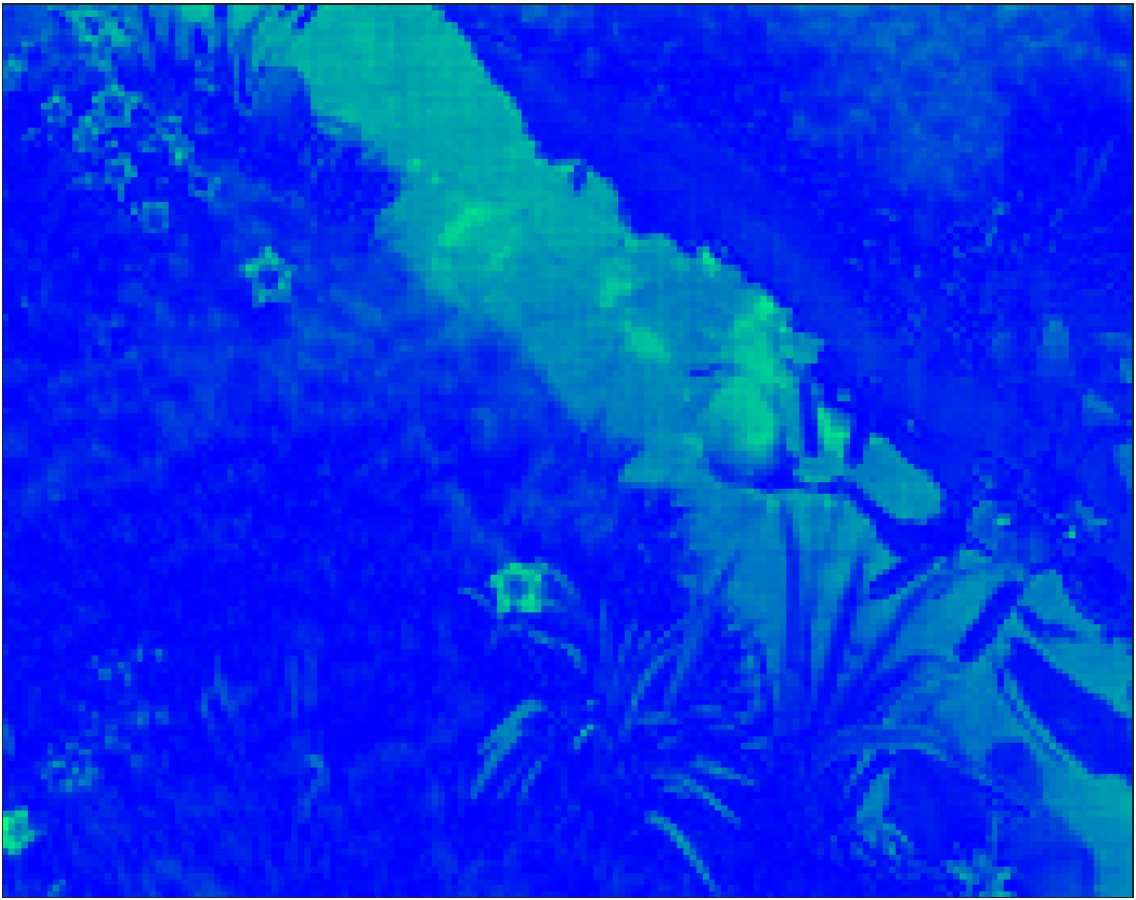}&
\includegraphics[width=0.124\textwidth]{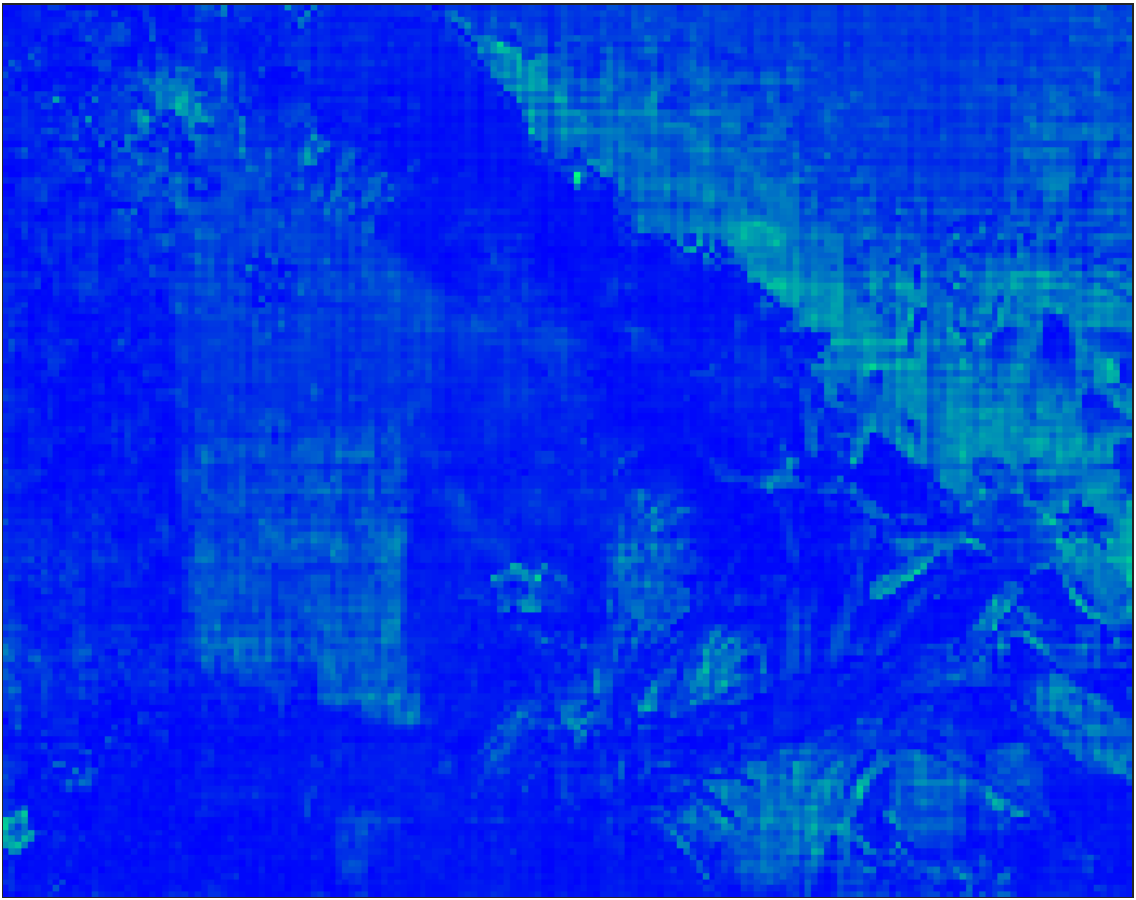}&
\includegraphics[width=0.124\textwidth]{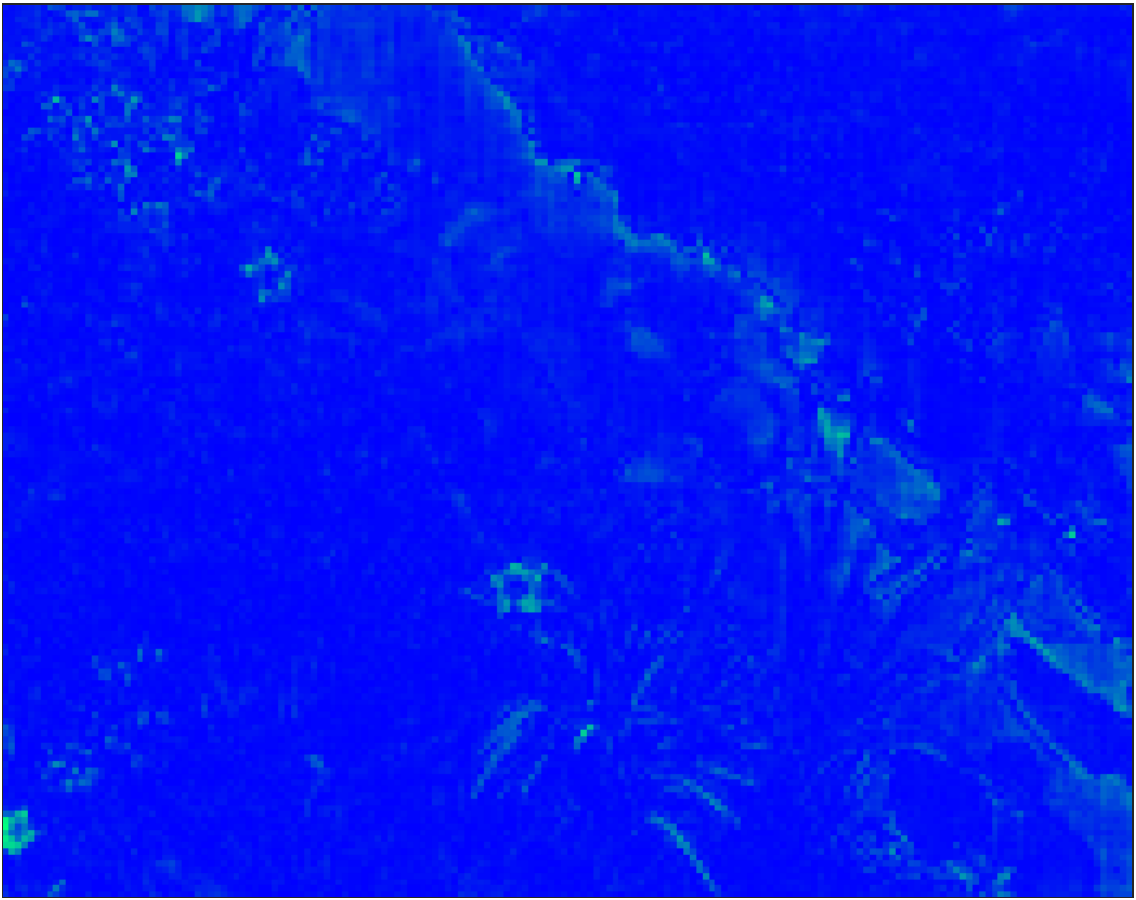}&
\includegraphics[width=0.124\textwidth]{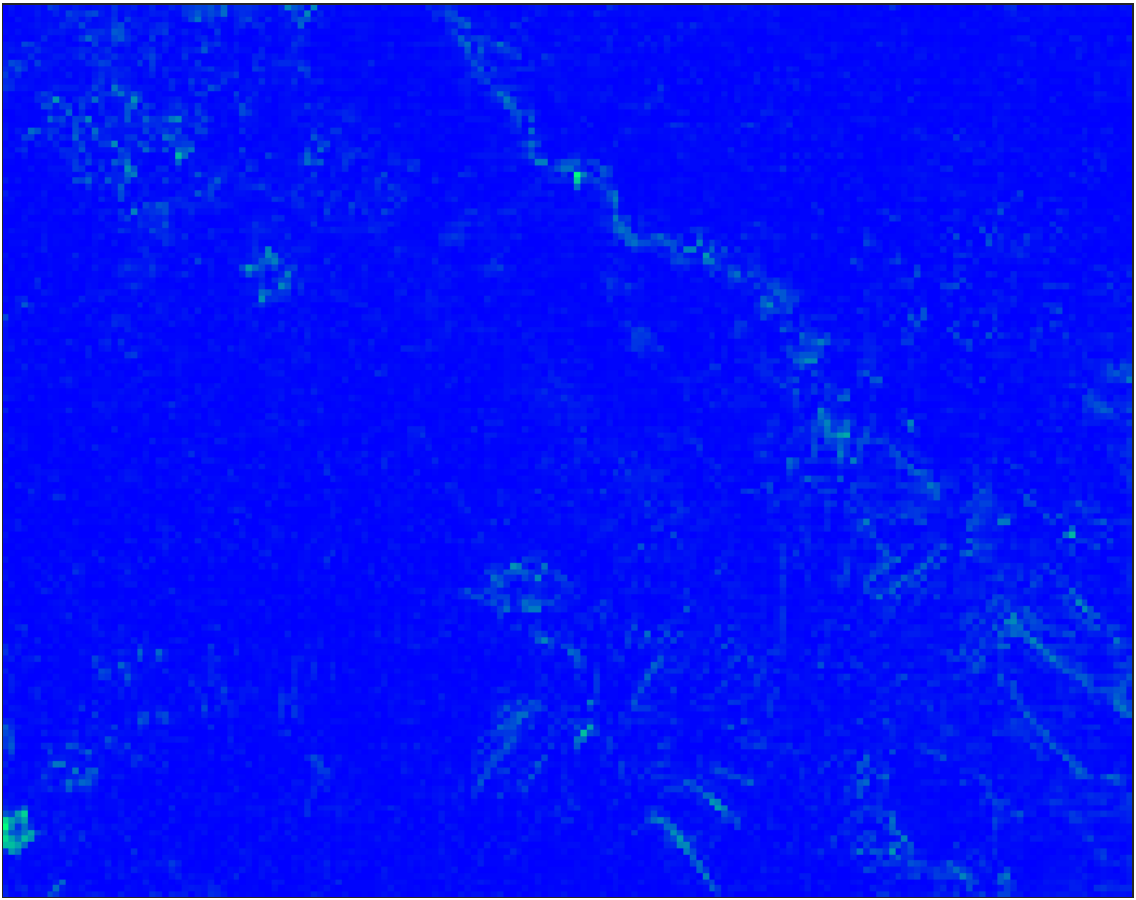}&
\includegraphics[width=0.124\textwidth]{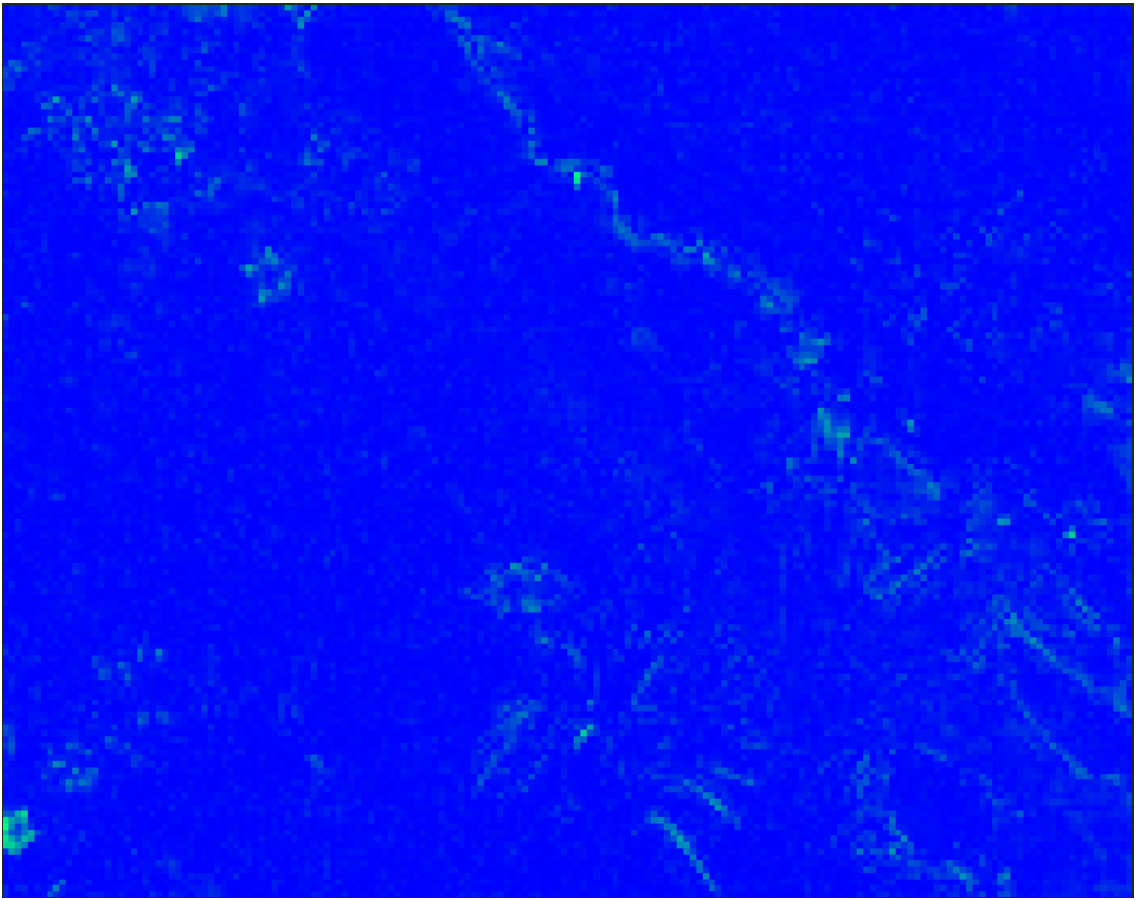}&
\includegraphics[width=0.124\textwidth]{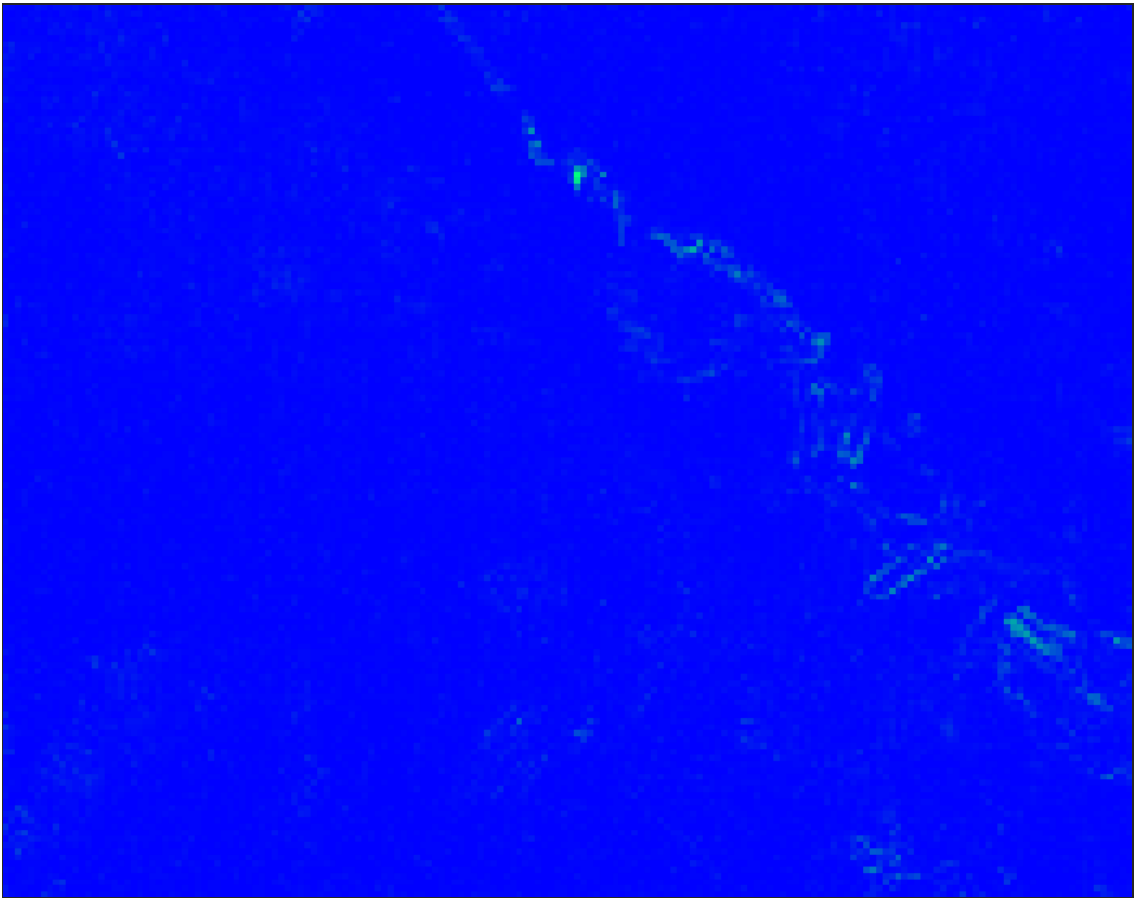}&
\includegraphics[width=0.124\textwidth]{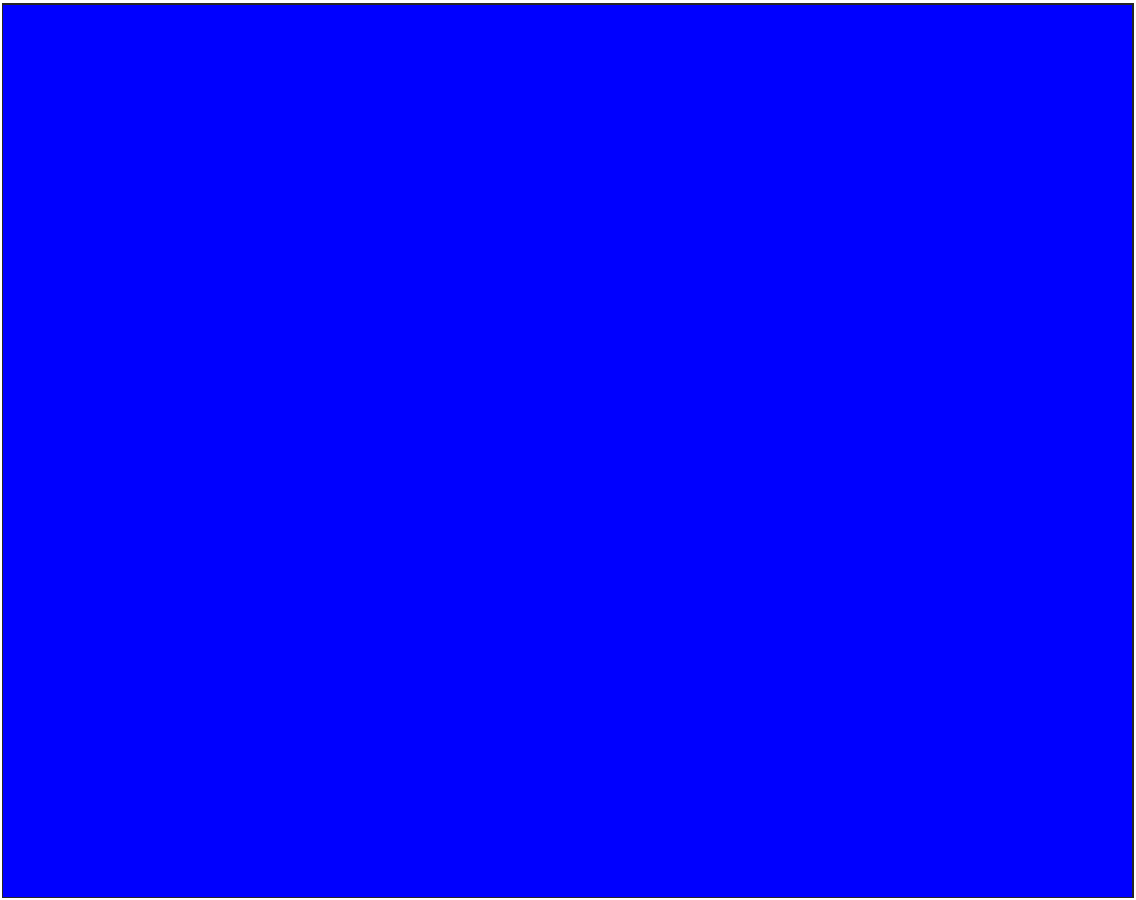}
\\
\includegraphics[width=0.124\textwidth]{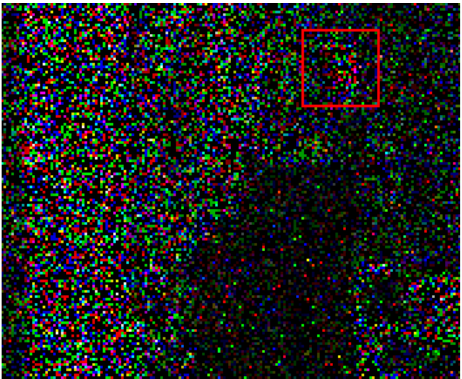}&
\includegraphics[width=0.124\textwidth]{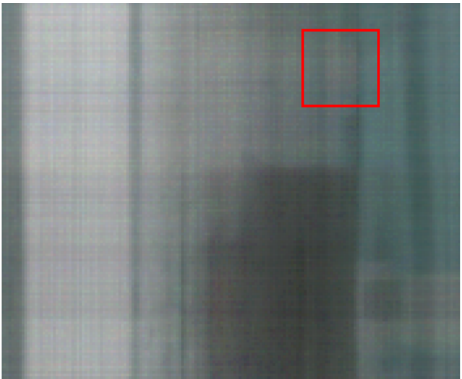}&
\includegraphics[width=0.124\textwidth]{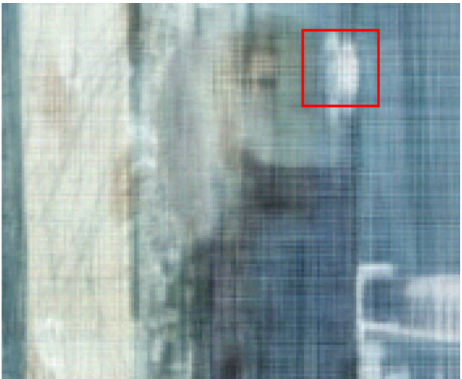}&
\includegraphics[width=0.124\textwidth]{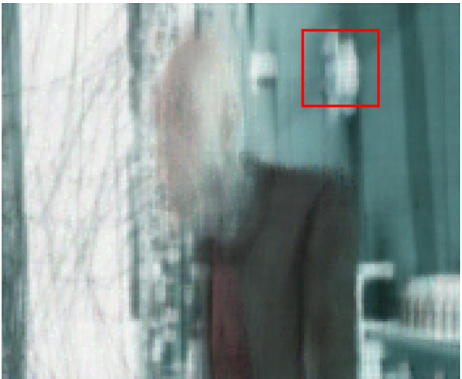}&
\includegraphics[width=0.124\textwidth]{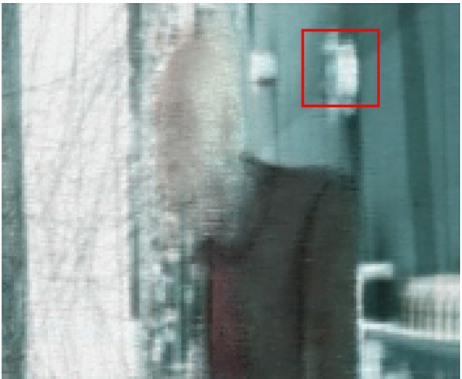}&
\includegraphics[width=0.124\textwidth]{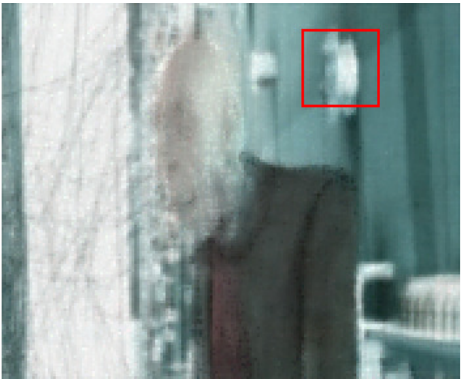}&
\includegraphics[width=0.124\textwidth]{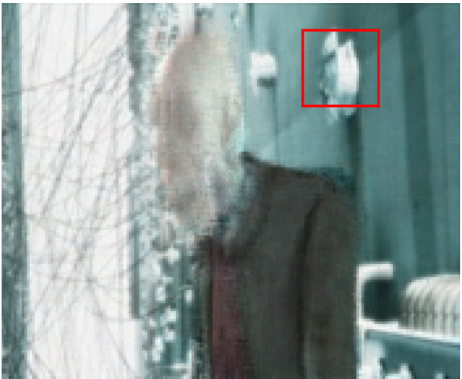}&
\includegraphics[width=0.124\textwidth]{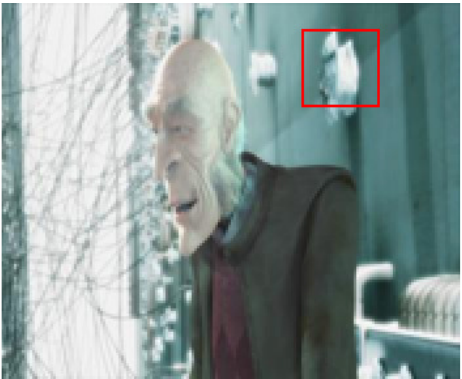}
\\
\includegraphics[width=0.124\textwidth]{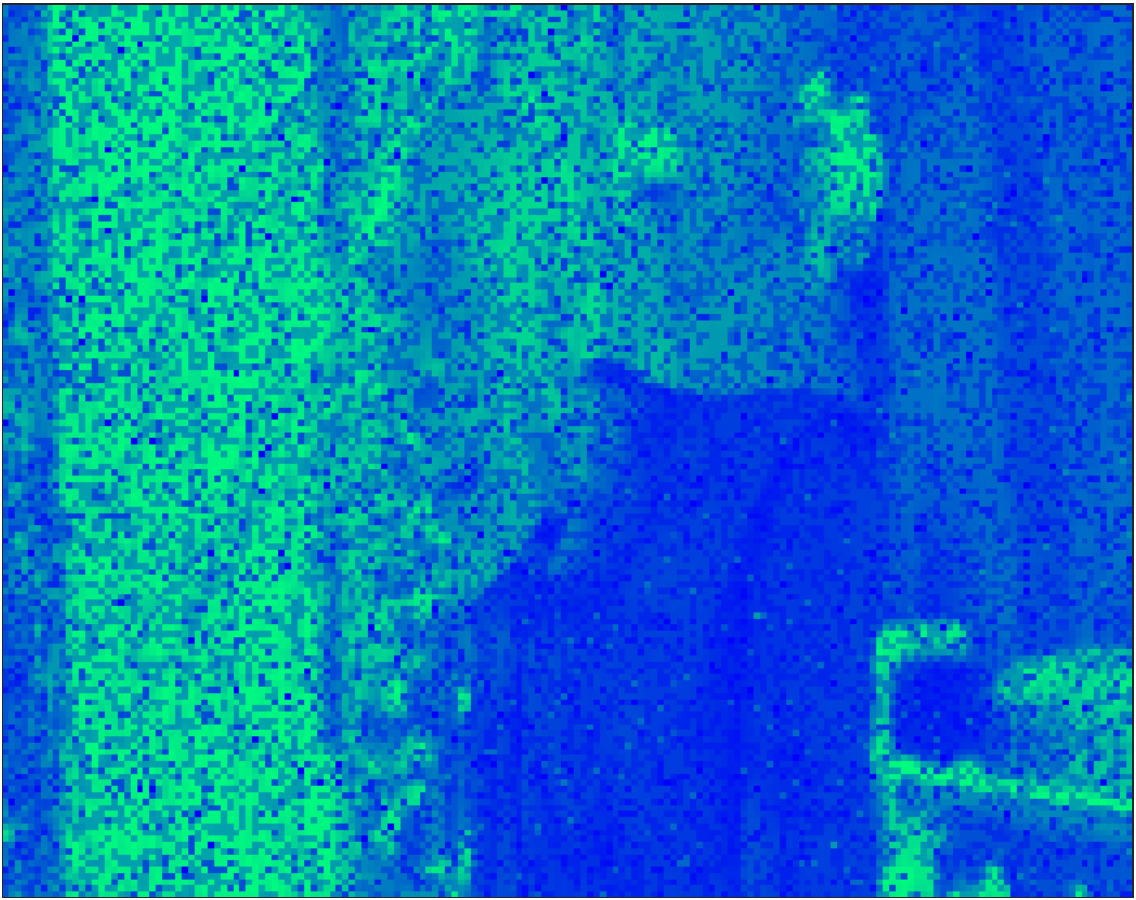}&
\includegraphics[width=0.124\textwidth]{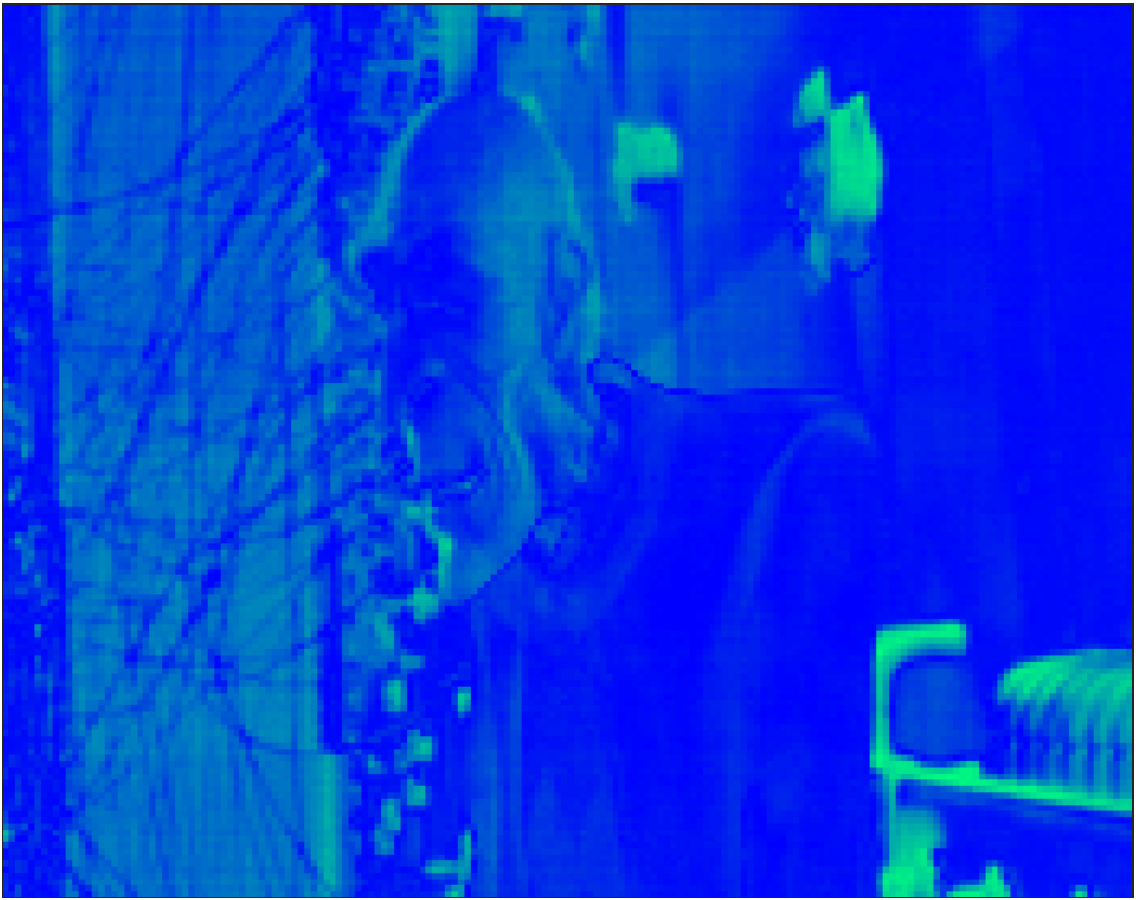}&
\includegraphics[width=0.124\textwidth]{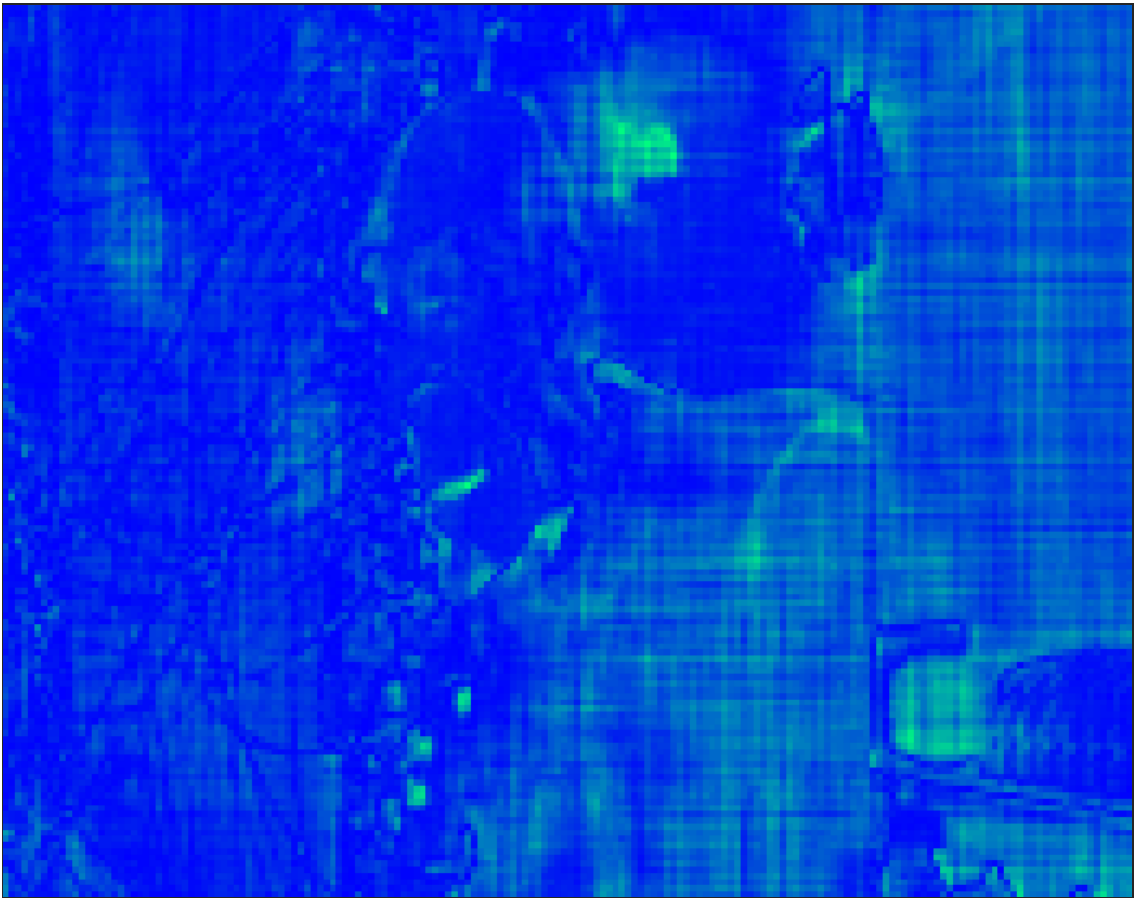}&
\includegraphics[width=0.124\textwidth]{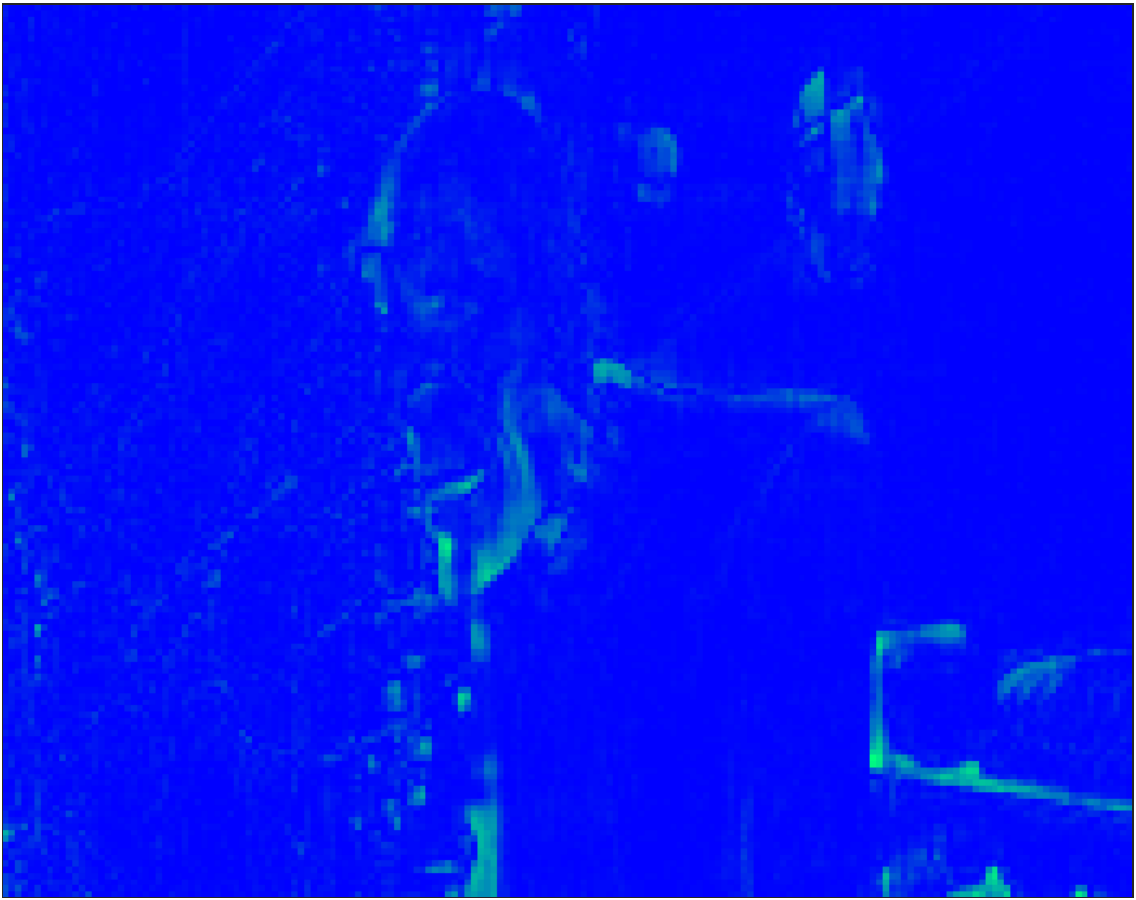}&
\includegraphics[width=0.124\textwidth]{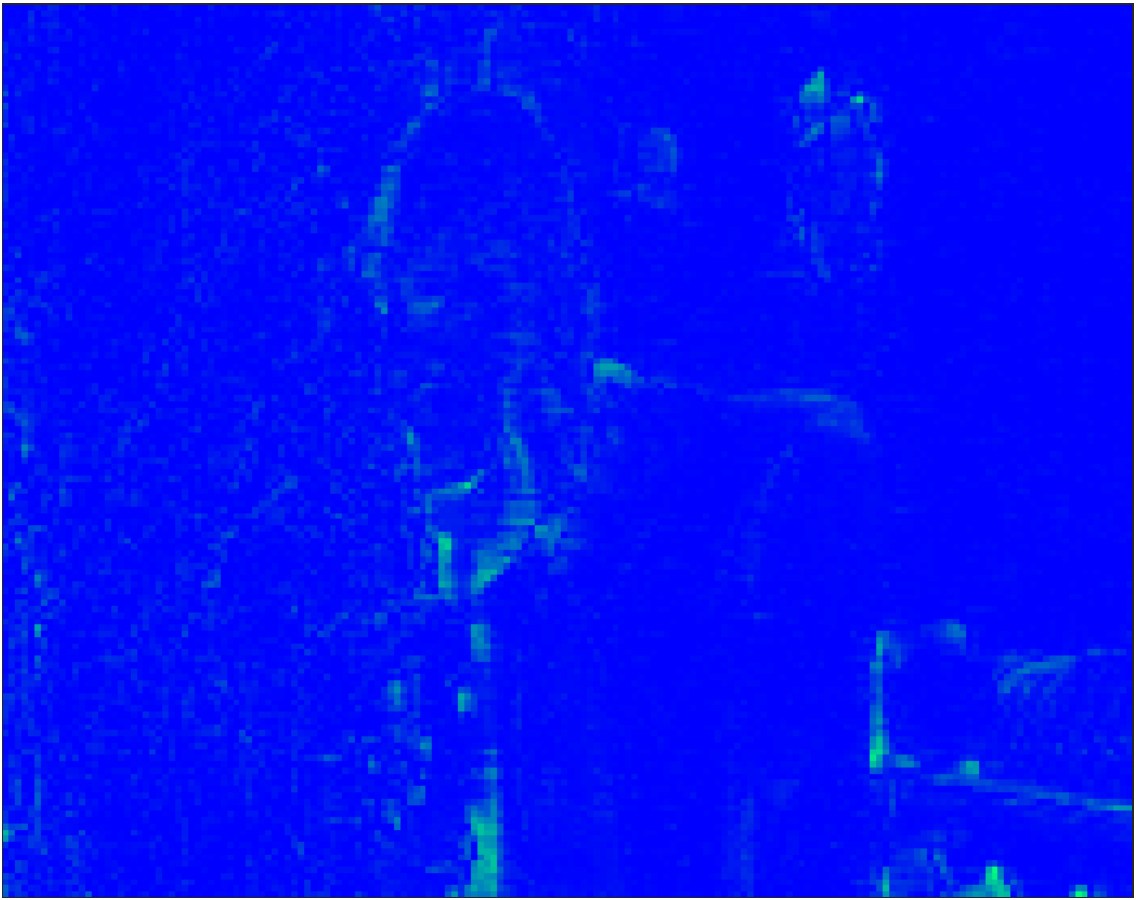}&
\includegraphics[width=0.124\textwidth]{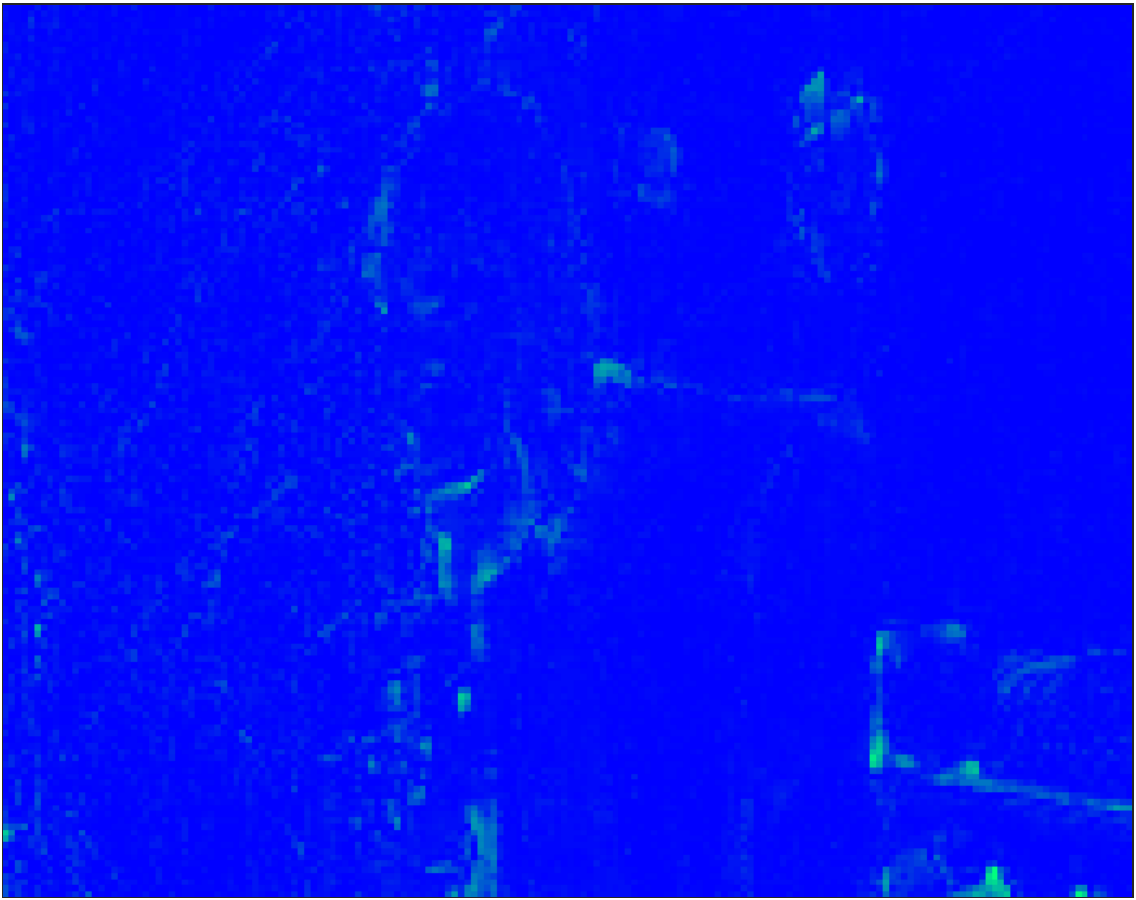}&
\includegraphics[width=0.124\textwidth]{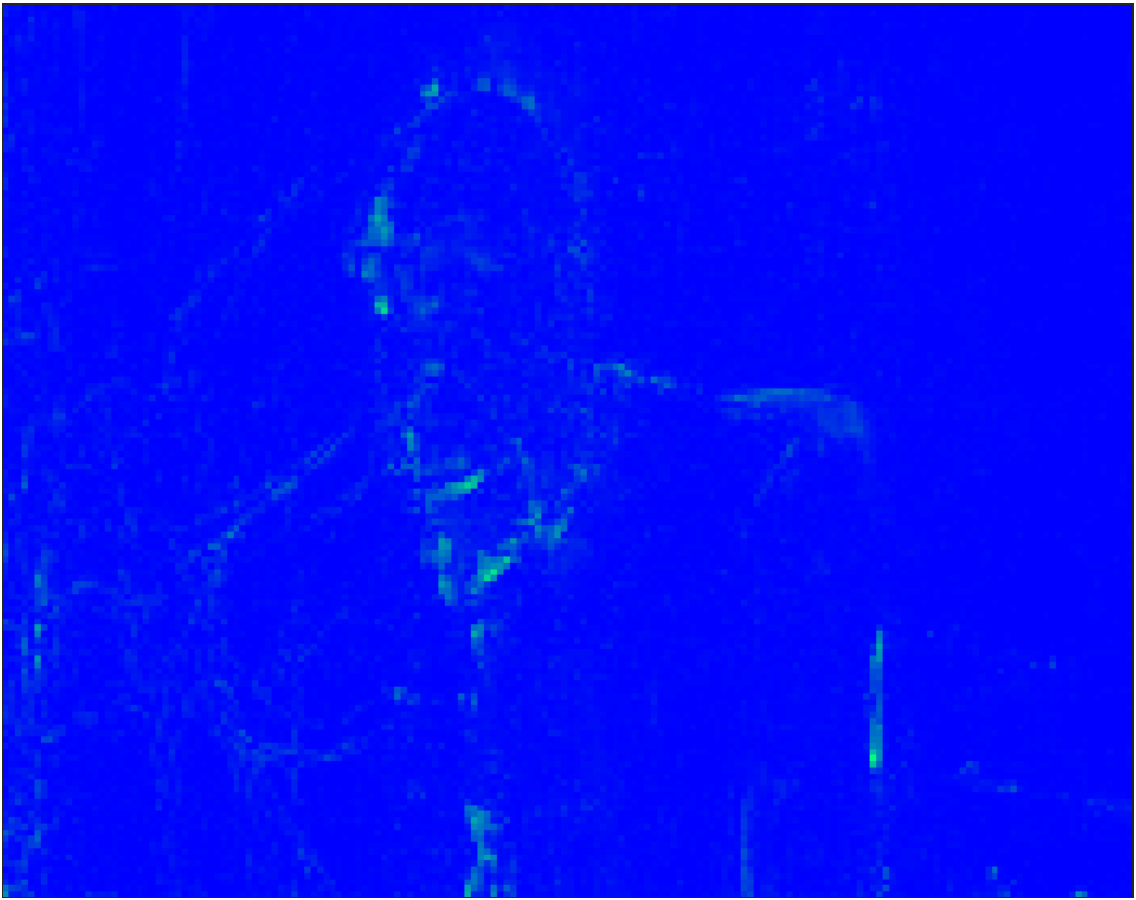}&
\includegraphics[width=0.124\textwidth]{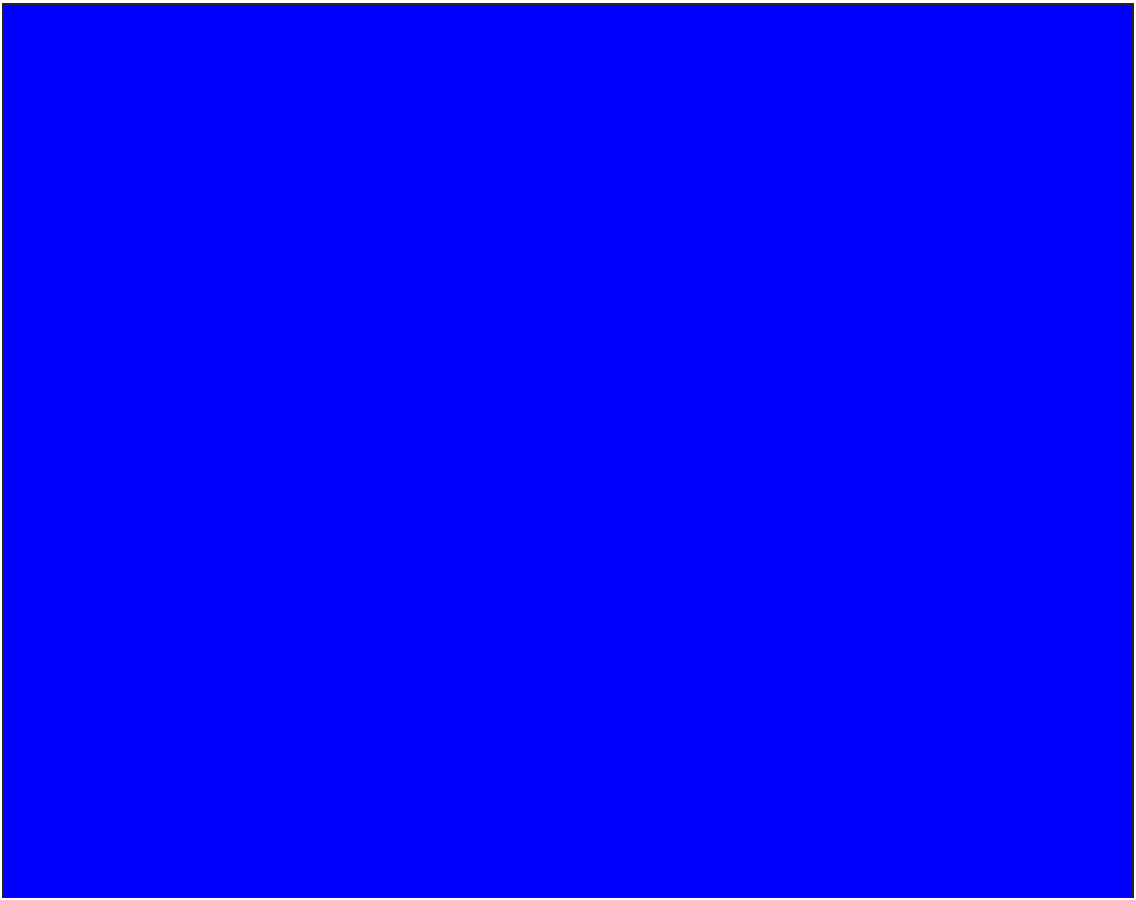}  \\
  \end{tabular}
  \vspace{-0.2cm}
  \includegraphics[width=0.95\textwidth]{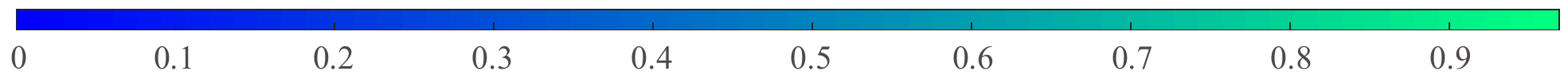}
  \caption{Recovered results on two color videos with SR=0.2 and SaP=0.1. The first row and third row are visual results at the 1st frame of $\mathit{bunny}$ and the 50th frame of $\mathit{elephants}$, respectively. The second row and fourth row are the corresponding residual images.} \vspace{-0.6cm}
  \label{videos}
  \end{center}
\end{figure}

We report MPSNR and MSSIM values obtained by all compared RTC methods on the videos $\mathit{bunny}$ and $\mathit{elephants}$ in Table \ref{tab:addlabel1}. As observed, the proposed methods obtain the overall optimality. In general, the MPSNR and MSSIM values obtained by RNC-FCTN are higher than those by RC-FCTN because the SaP can be better removed in the process of low-rank FCTN decomposition.

\renewcommand{\arraystretch}{1.45}
 \begin{table}[h]\small
  \centering
  \caption{The quantitative comparison of different methods on HSV.}
  \begin{threeparttable}
  \vspace{0.25cm}
  \setlength{\tabcolsep}{5.2pt}
    \begin{tabular}{cccccccccc}
       \Xhline{1.2pt}
     \multirow{2}[0]{*}{SR} & Salt and & \multirow{2}[0]{*}{Indicators} & \multirow{2}[0]{*}{Noise} & \multirow{2}[0]{*}{SNN} & \multirow{2}[0]{*}{TNN} & \multirow{2}[0]{*}{TTNN} & \multirow{2}[0]{*}{RTRC}&\multirow{2}[0]{*}{RC-FCTN} &\multirow{2}[0]{*}{RNC-FCTN}\\
     &pepper noise  &  &  &  &   &  & & & \\
        \Xhline{1.2pt}
      &\multirow{2}[0]{*}{0.1} &MPSNR&9.3671&26.891&43.225&45.713&45.732& \underline{46.639}&\textbf{46.912} \\
      \multirow{2}[0]{*}{0.3}& &MSSIM&0.0712& 0.8717& 0.9930&0.9954&0.9962& \underline{0.9965}&\textbf{0.9967} \\
      \cline{2-10}
      &\multirow{2}[0]{*}{0.2} &MPSNR&9.0154& 24.243& 39.754&43.426&42.619& \underline{43.792}&\textbf{44.727} \\
       &  &MSSIM&0.0547& 0.7589& 0.9873&0.9944&0.9943& \textbf{0.9955}&\underline{0.9952} \\
        \cline{1-10}
        \cline{1-10}
       &\multirow{2}[0]{*}{0.1} &MPSNR&8.9362& 24.124&37.765&43.178&42.704& \underline{43.978}&\textbf{44.383} \\
       \multirow{2}[0]{*}{0.2}&  &MSSIM&0.0477& 0.7899& 0.9828&0.9934&0.9944& \textbf{0.9949}&\underline{0.9946} \\
        \cline{2-10}
        &\multirow{2}[0]{*}{0.2} &MPSNR&8.7161&21.319&34.372&\underline{40.199}&39.088& 39.618&\textbf{41.928} \\
        &  &MSSIM&0.0381& 0.7217& 0.9687&\textbf{0.9925}&0.9906&0.9914&\underline{0.9924} \\
      \cline{1-10}
      \cline{1-10}
        &\multirow{2}[0]{*}{0.1} &MPSNR& 8.5452&19.232&27.719&36.681&36.558& \underline{37.883}&\textbf{39.195} \\
       \multirow{2}[0]{*}{0.1}&  &MSSIM&0.0256& 0.6276& 0.8930&0.9849&0.9831& \textbf{0.9879}&\underline{0.9855} \\
         \cline{2-10}
       &\multirow{2}[0]{*}{0.2} &MPSNR&8.4481&17.004& 25.141&34.379&33.654& \underline{34.492}&\textbf{36.021} \\
       &  &MSSIM&0.0216& 0.5265& 0.8206&0.9706&0.9715& \underline{0.9744}&\textbf{0.9745} \\
       \cline{1-10}
       \cline{1-10}
       &&average time (s)&&45.863&16.004&51.037&73.587&91.036&193.53\\
        \Xhline{1.2pt}
    \end{tabular}%
    \end{threeparttable}\label{footnote}
  \label{tab:addlabel2}%
\end{table}

\begin{figure}[!htp]
\footnotesize
\setlength{\tabcolsep}{0.97pt}
\begin{center}
\begin{tabular}{cccccccc}
Observed&SNN&TNN&TTNN&RTRC&RC-FCTN&RNC-FCTN& Ground truth \\
\includegraphics[width=0.124\textwidth]{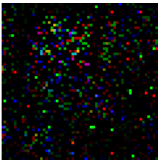}&
\includegraphics[width=0.124\textwidth]{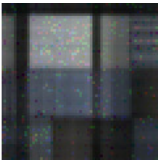}&
\includegraphics[width=0.124\textwidth]{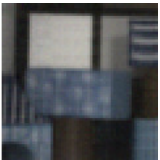}&
\includegraphics[width=0.124\textwidth]{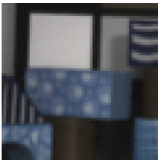}&
\includegraphics[width=0.124\textwidth]{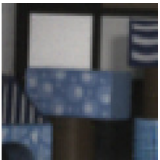}&
\includegraphics[width=0.124\textwidth]{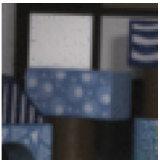}&
\includegraphics[width=0.124\textwidth]{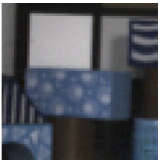}&
\includegraphics[width=0.124\textwidth]{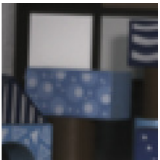}
\\
\includegraphics[width=0.124\textwidth]{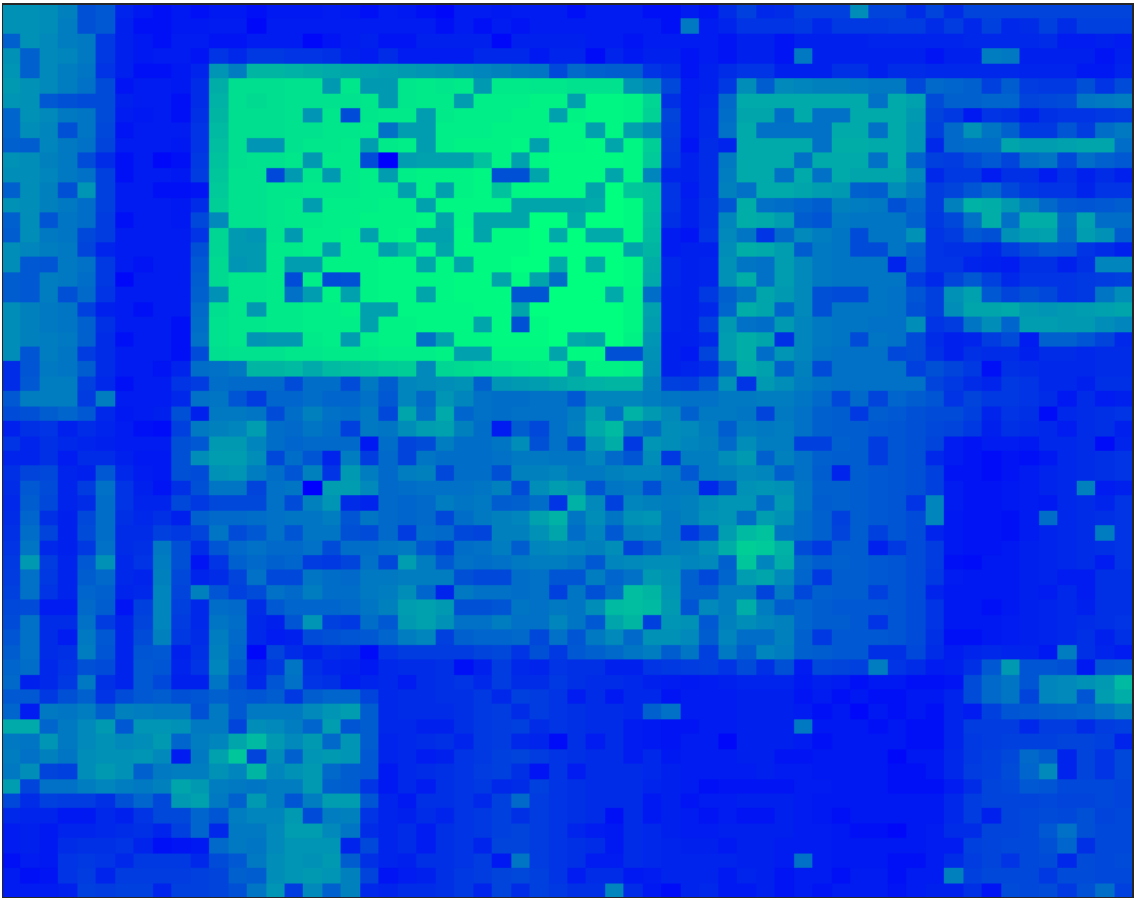}&
\includegraphics[width=0.124\textwidth]{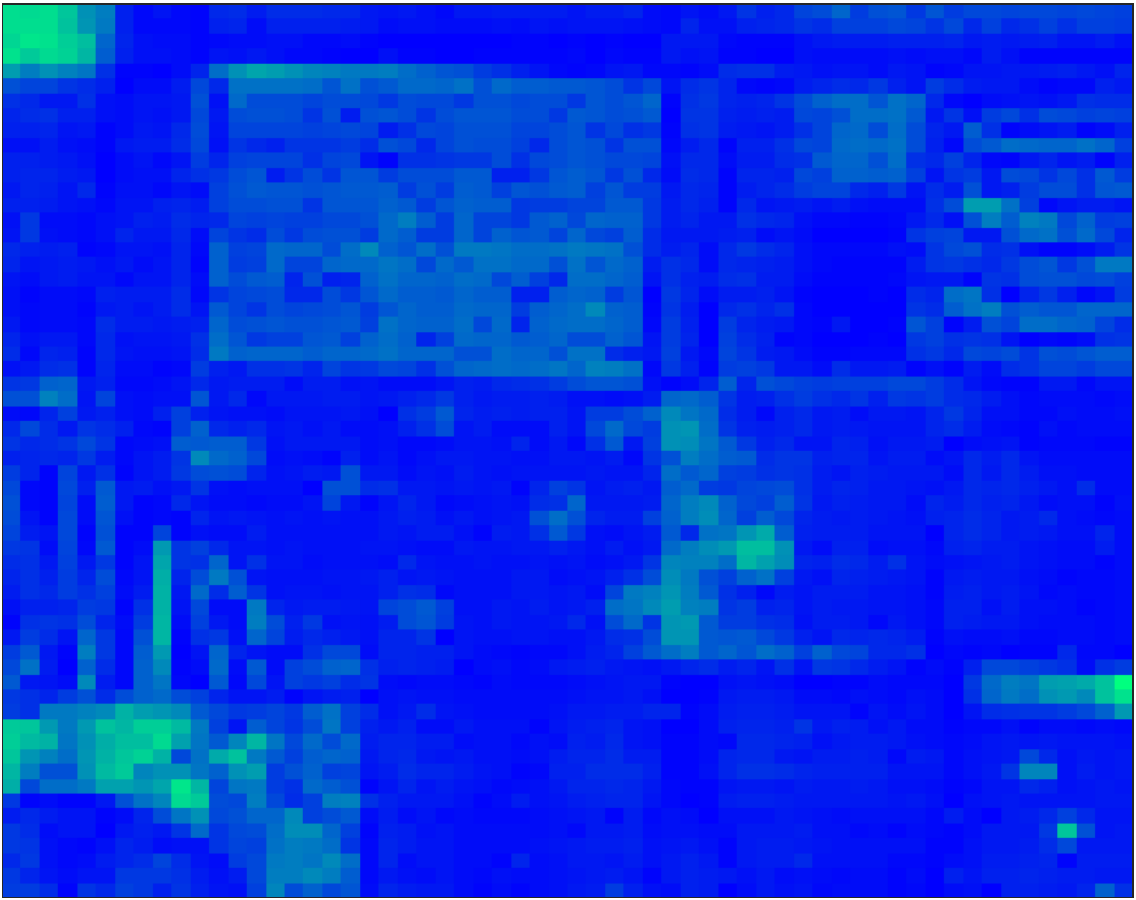}&
\includegraphics[width=0.124\textwidth]{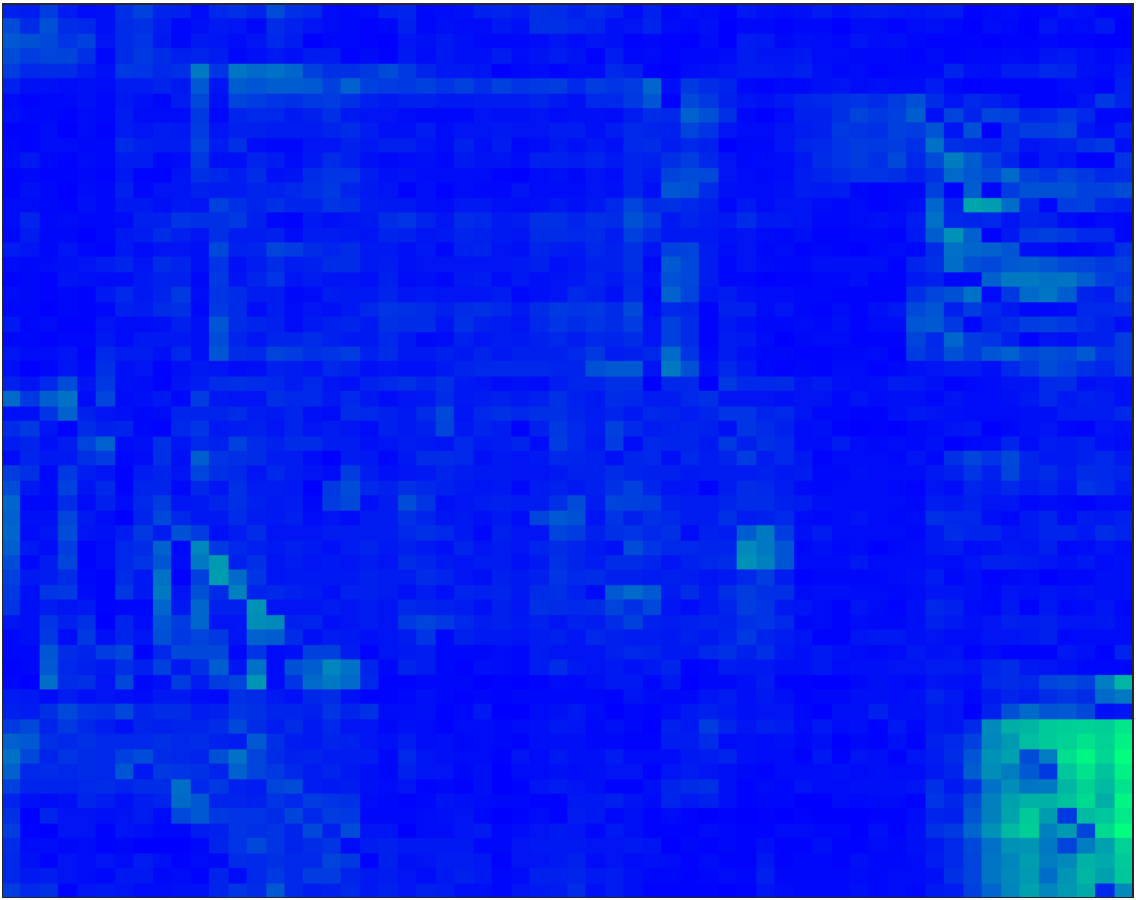}&
\includegraphics[width=0.124\textwidth]{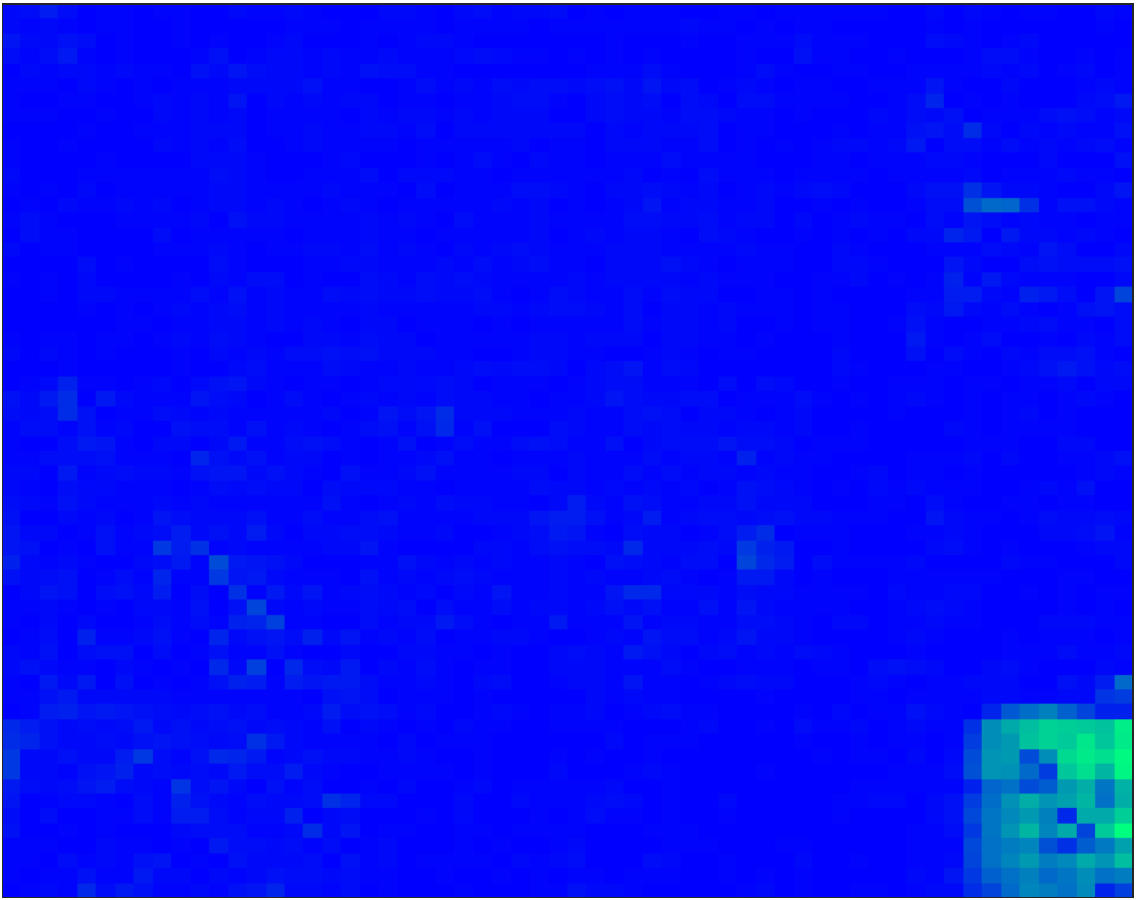}&
\includegraphics[width=0.124\textwidth]{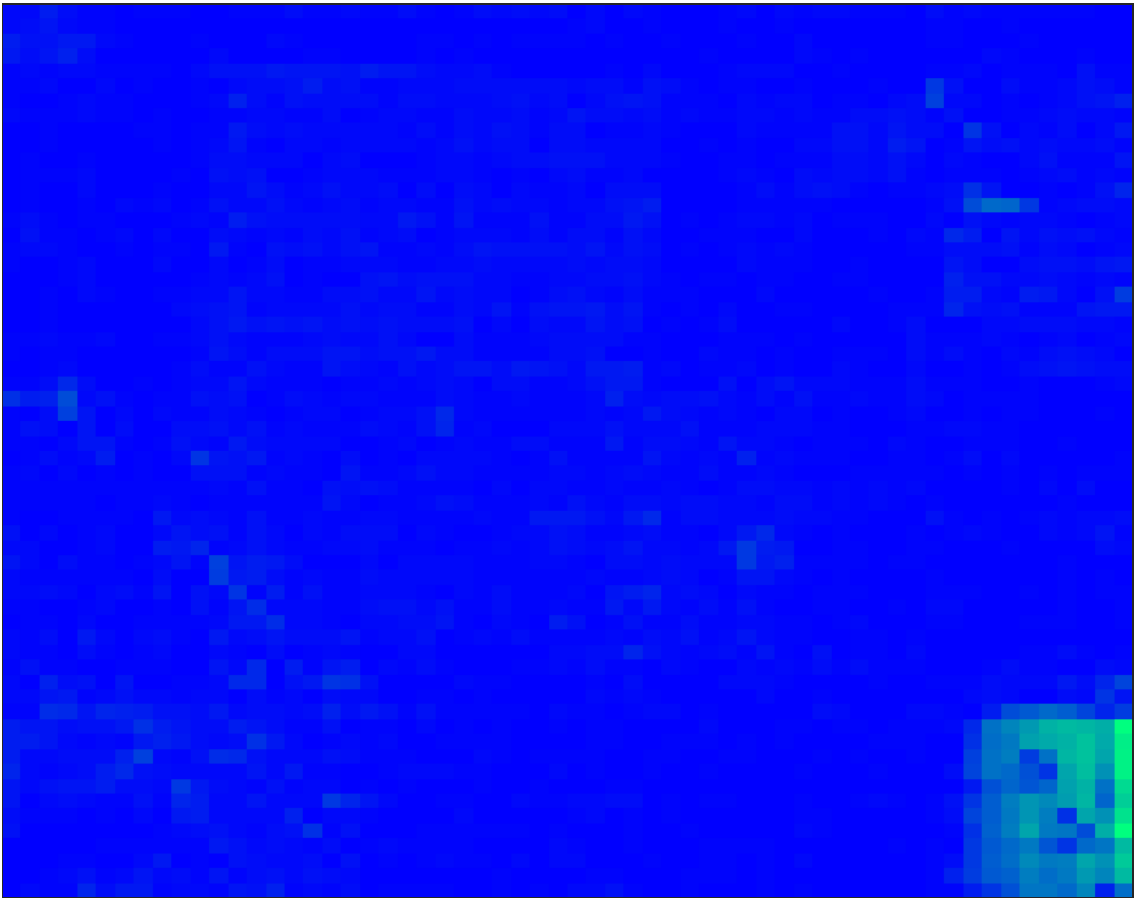}&
\includegraphics[width=0.124\textwidth]{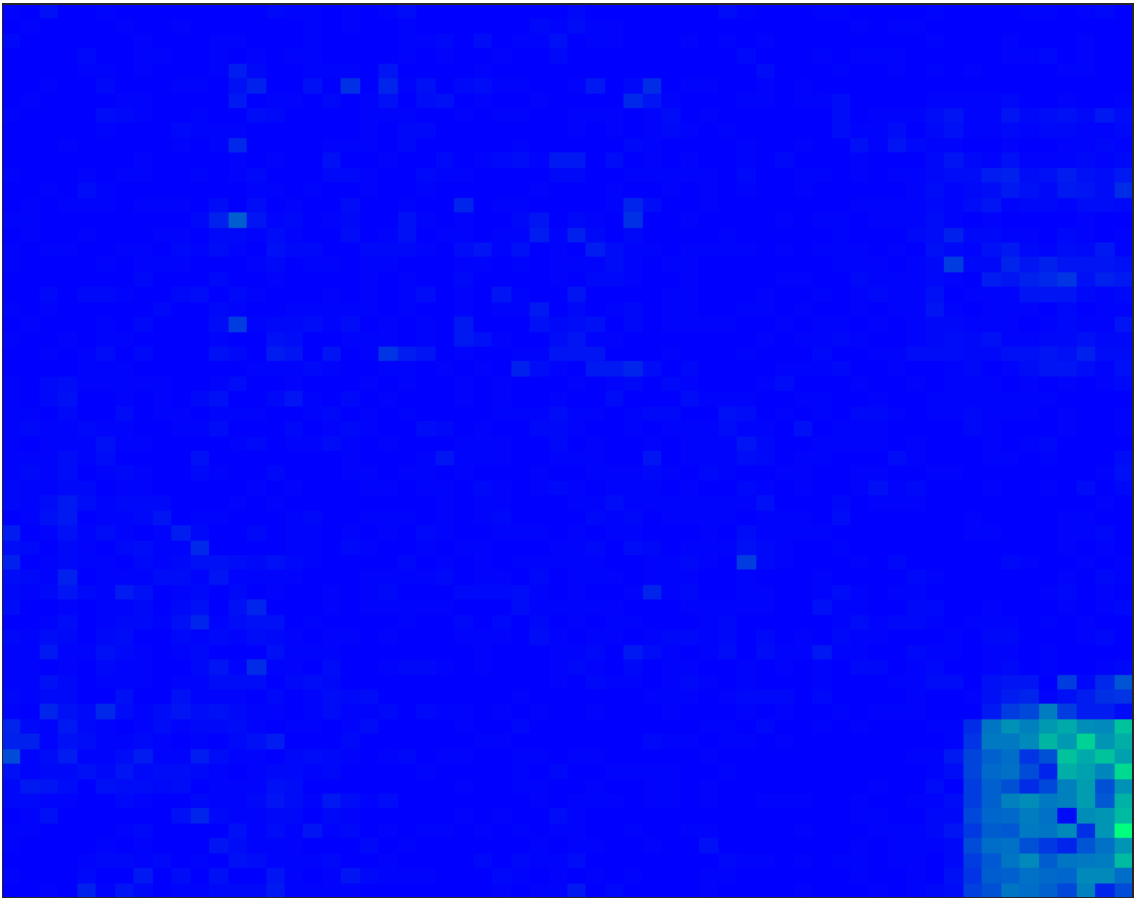}&
\includegraphics[width=0.124\textwidth]{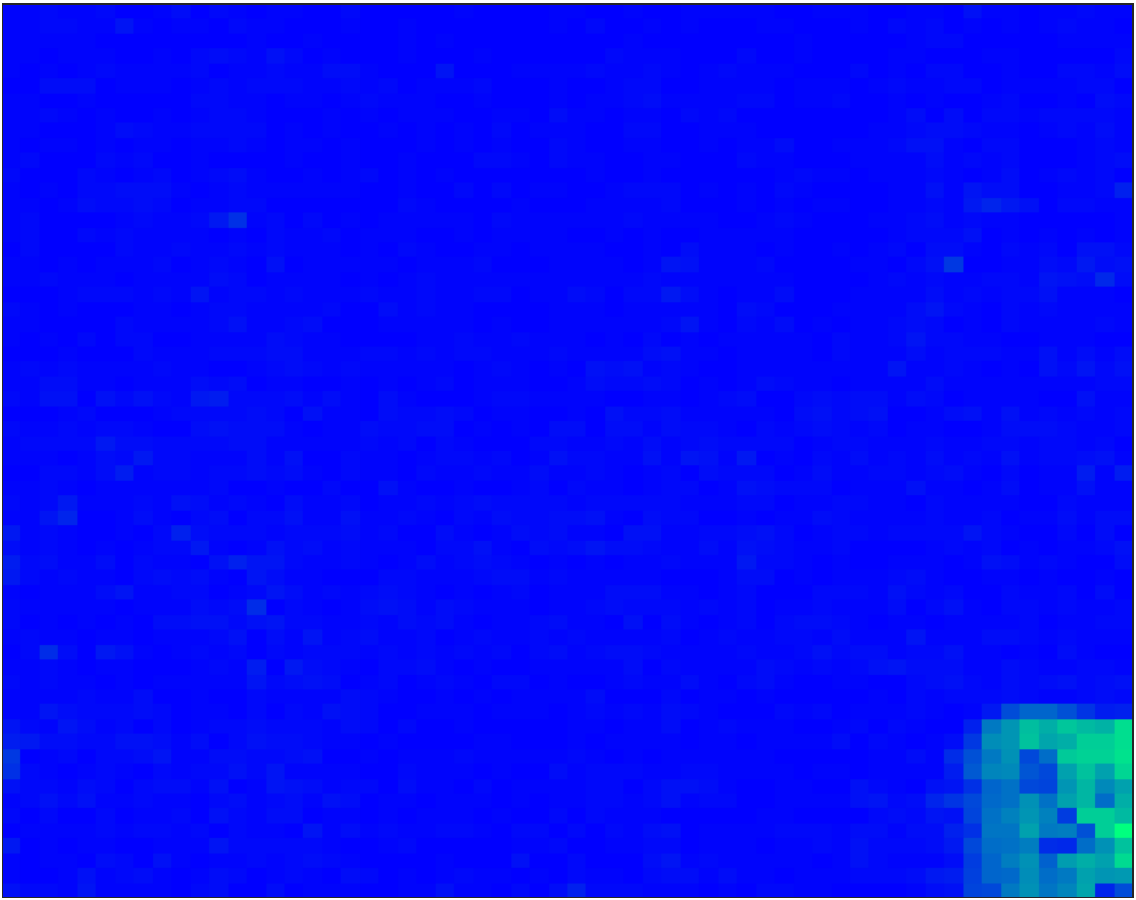}&
\includegraphics[width=0.124\textwidth]{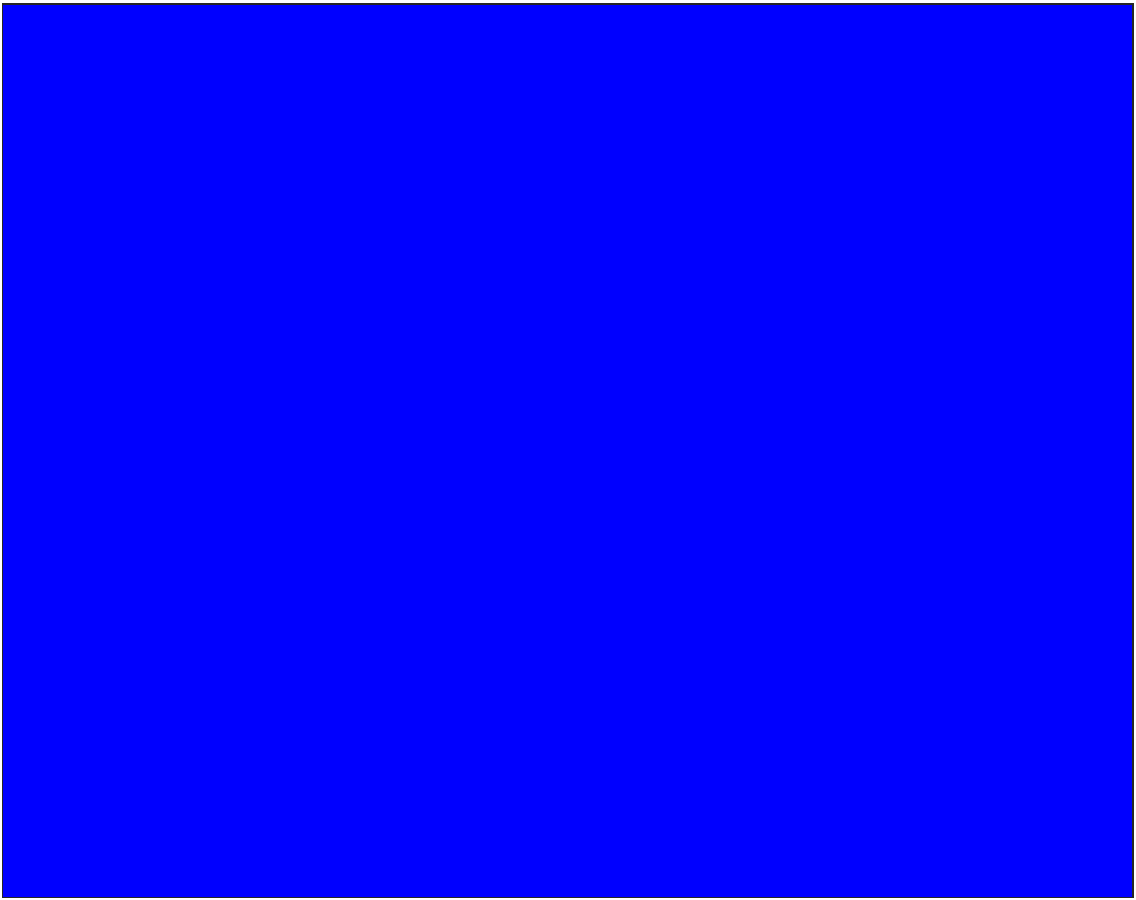}
\\
\includegraphics[width=0.124\textwidth]{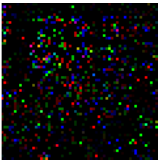}&
\includegraphics[width=0.124\textwidth]{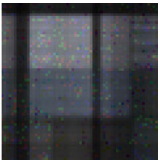}&
\includegraphics[width=0.124\textwidth]{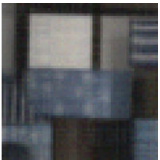}&
\includegraphics[width=0.124\textwidth]{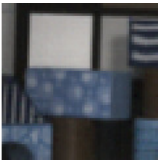}&
\includegraphics[width=0.124\textwidth]{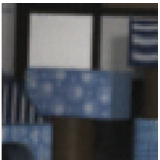}&
\includegraphics[width=0.124\textwidth]{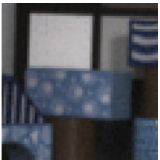}&
\includegraphics[width=0.124\textwidth]{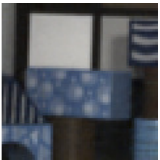}&
\includegraphics[width=0.124\textwidth]{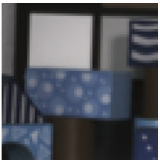}
\\
\includegraphics[width=0.124\textwidth]{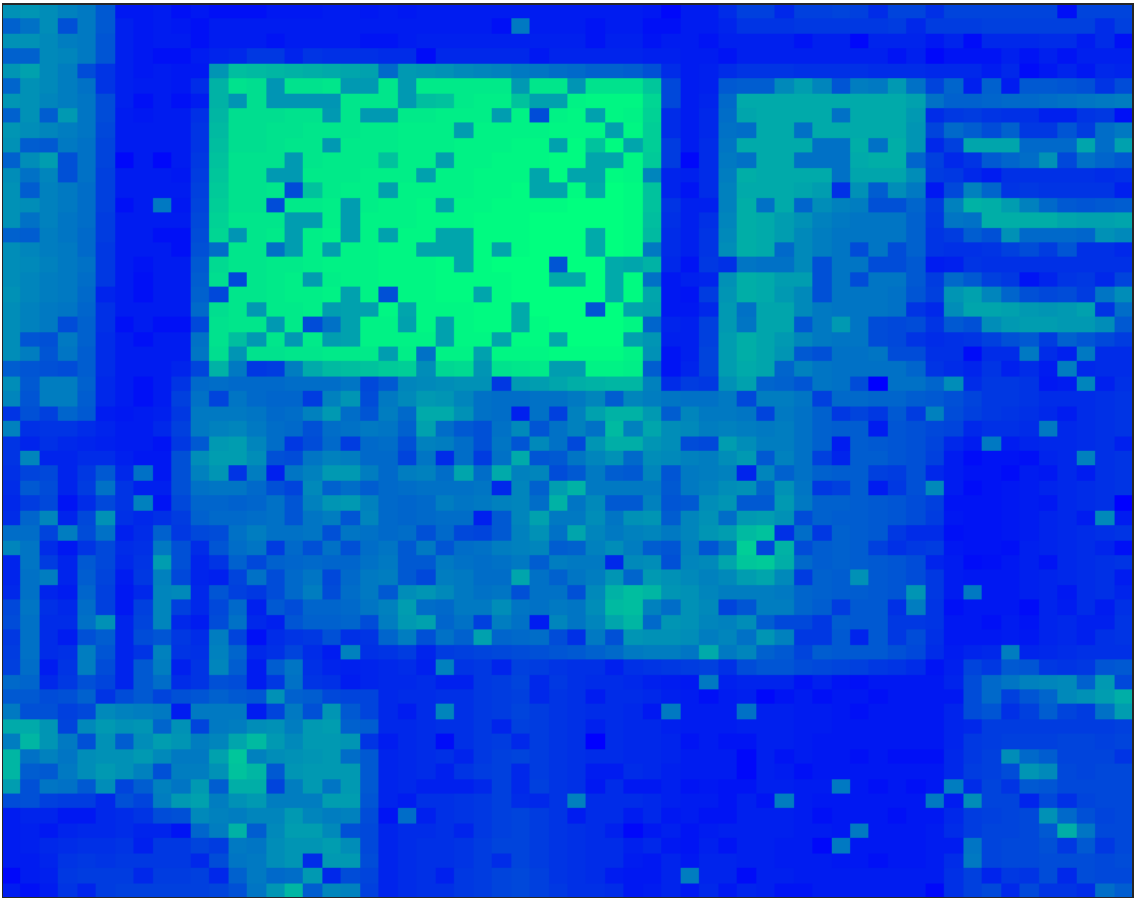}&
\includegraphics[width=0.124\textwidth]{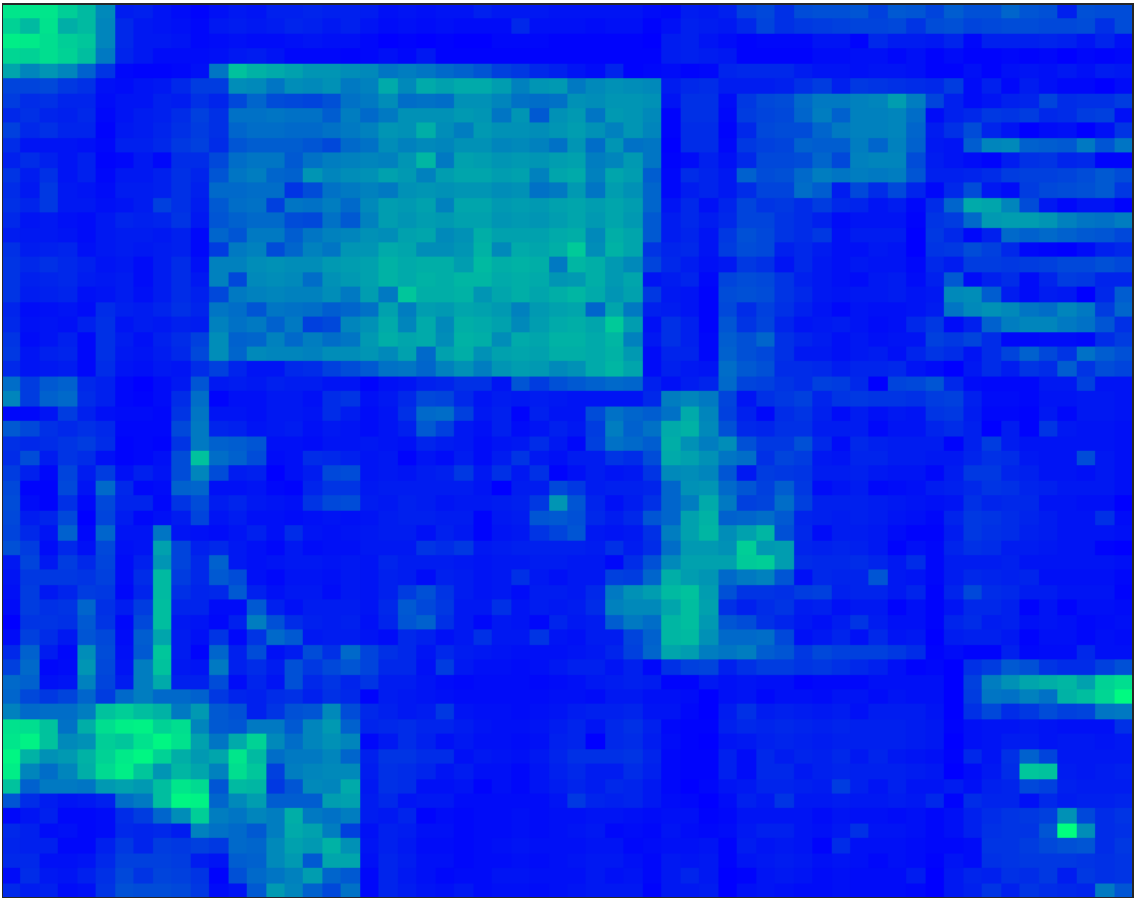}&
\includegraphics[width=0.124\textwidth]{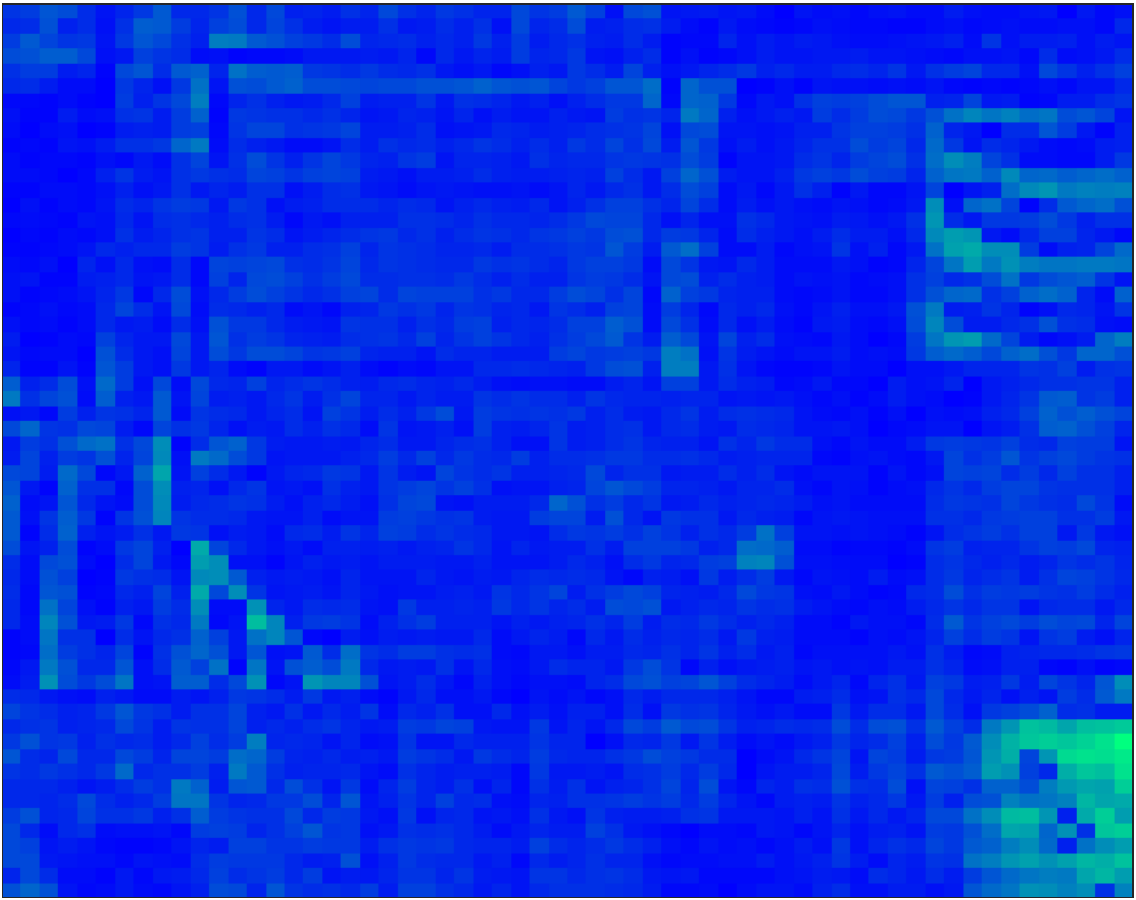}&
\includegraphics[width=0.124\textwidth]{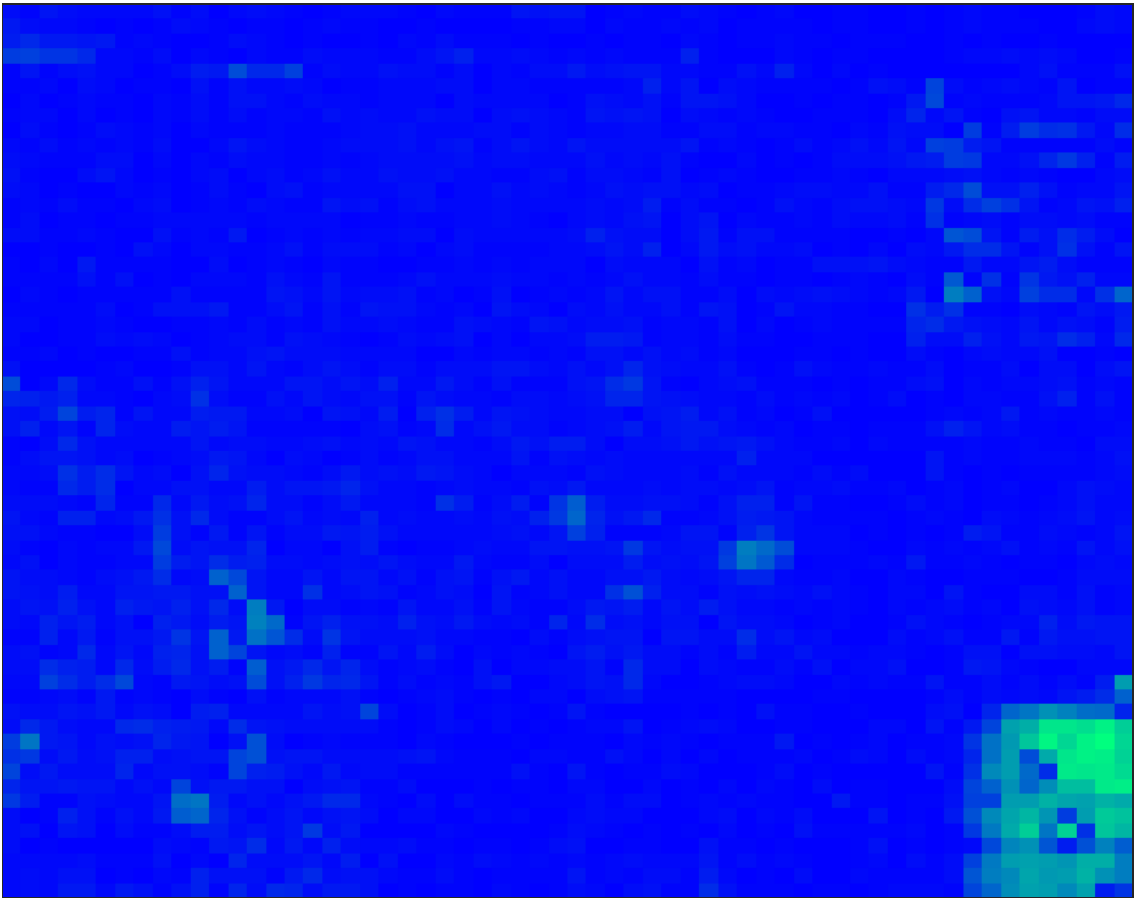}&
\includegraphics[width=0.124\textwidth]{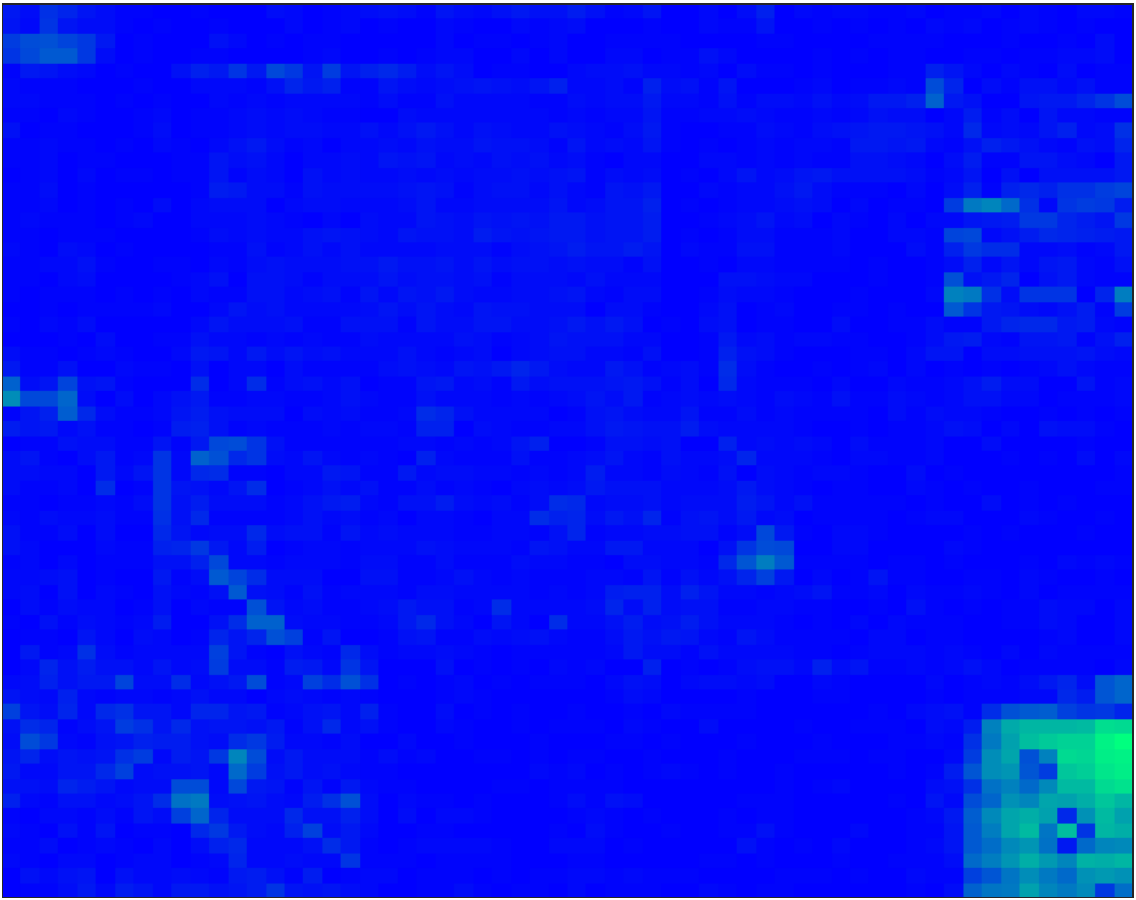}&
\includegraphics[width=0.124\textwidth]{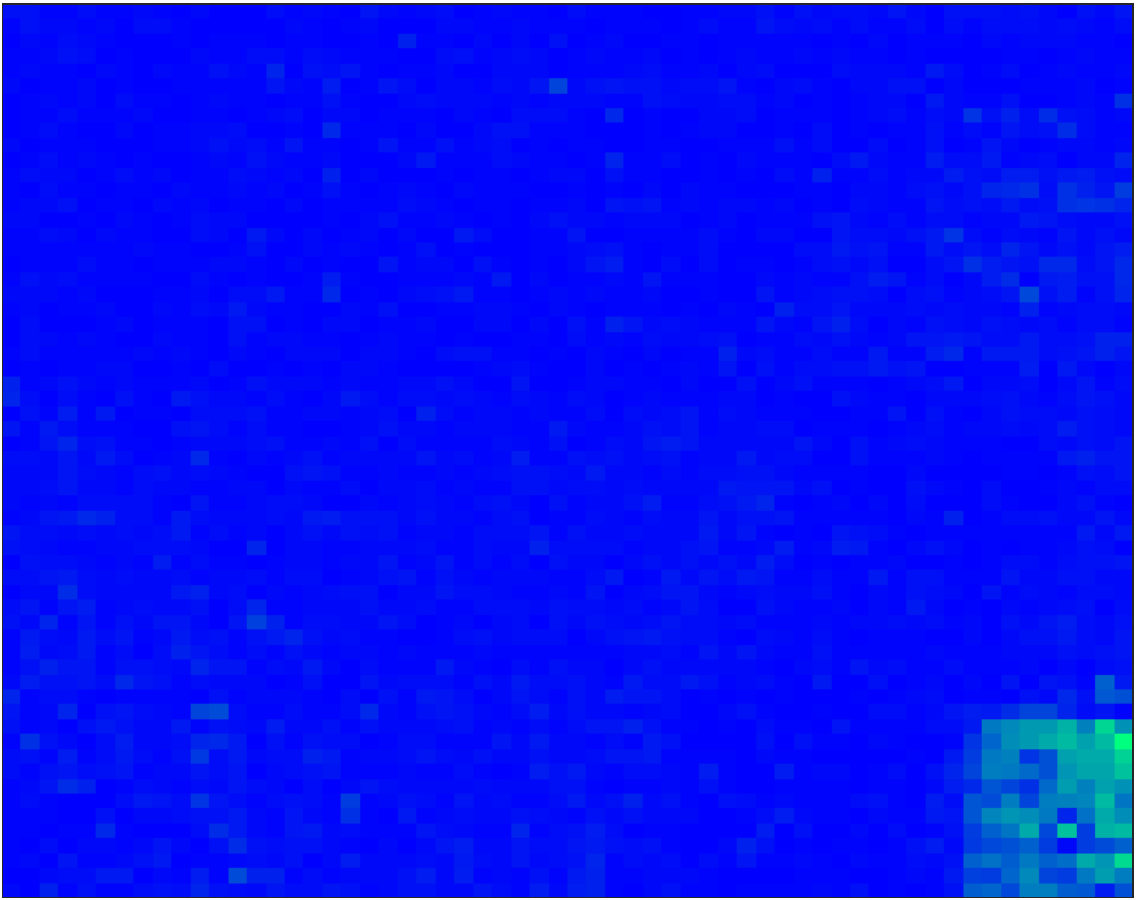}&
\includegraphics[width=0.124\textwidth]{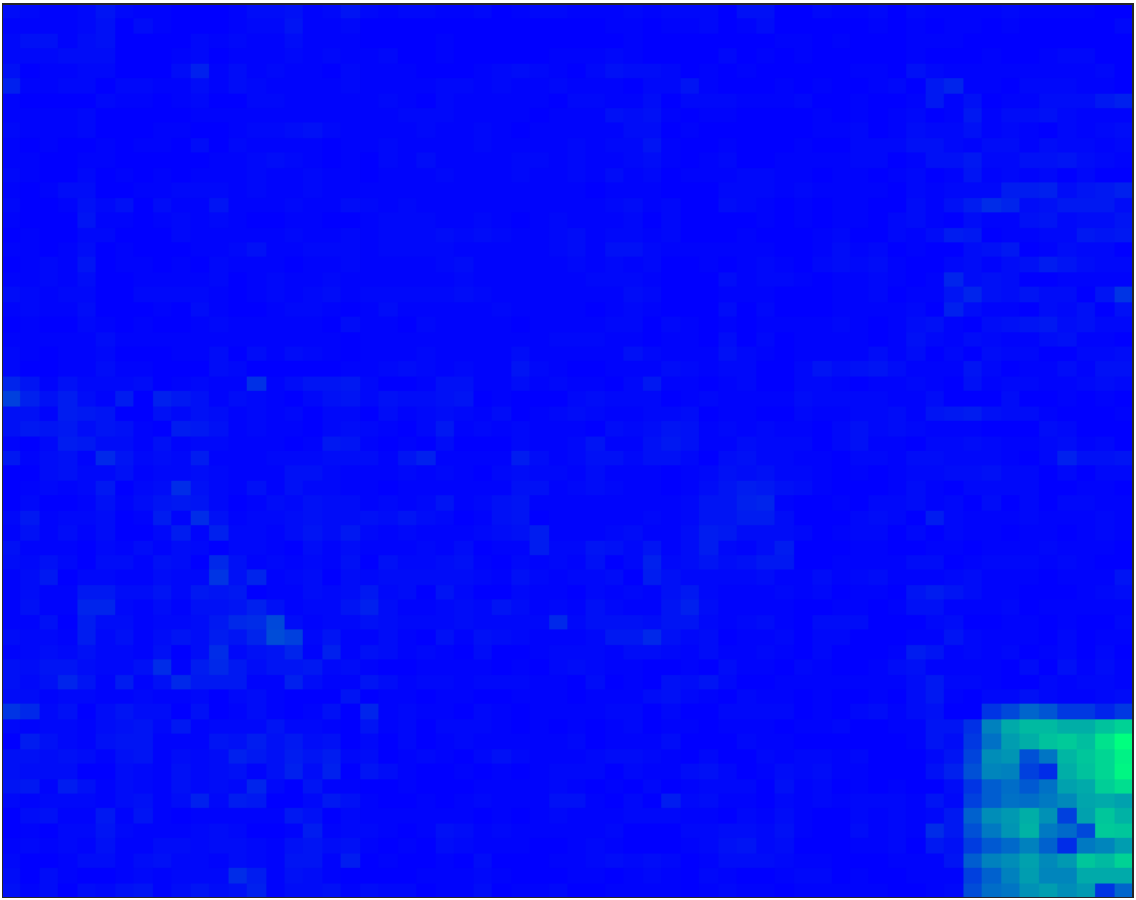}&
\includegraphics[width=0.124\textwidth]{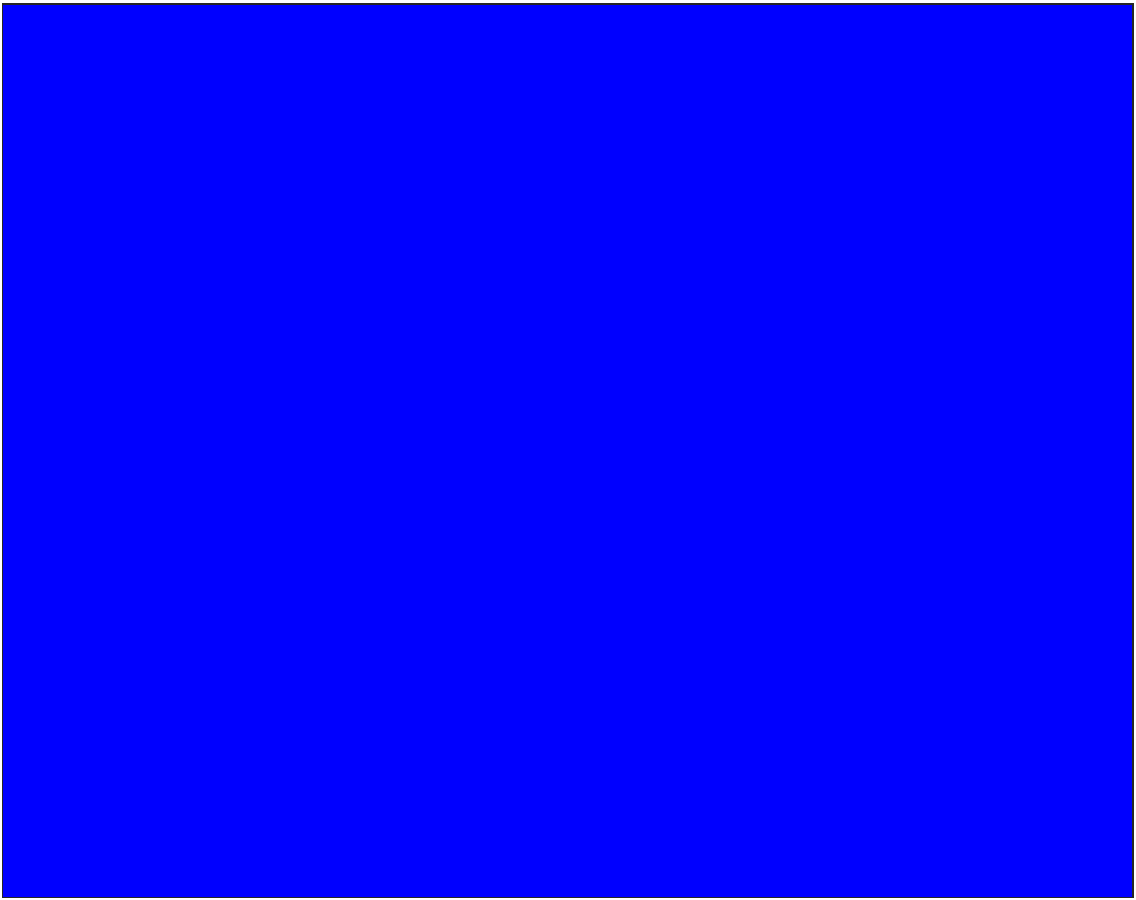}
\\
  \end{tabular}
    \vspace{-0.2cm}
  \includegraphics[width=0.95\textwidth]{eps2pdf/elephants/1.PDF}
  \caption{Recovered results on the HSV (band 8, 9, and 10 of first frame are picked as red, green, and blue channels). The first row and third row are visual results with SR=0.1, SaP=0.1 and 0.2, respectively. The second row and fourth row are the corresponding residual images.}\vspace{-0.6cm}
  \label{HSV}
  \end{center}
\end{figure}

\begin{figure}[t]
\footnotesize
\setlength{\tabcolsep}{0.97pt}
\begin{center}
\begin{tabular}{cccccccc}
\includegraphics[width=0.124\textwidth]{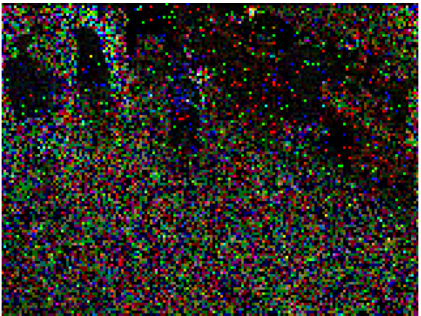}&
\includegraphics[width=0.124\textwidth]{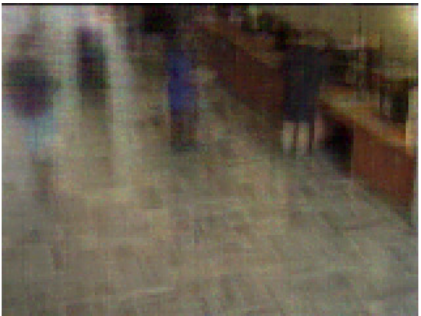}&
\includegraphics[width=0.124\textwidth]{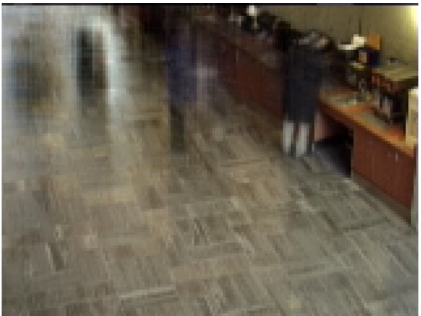}&
\includegraphics[width=0.124\textwidth]{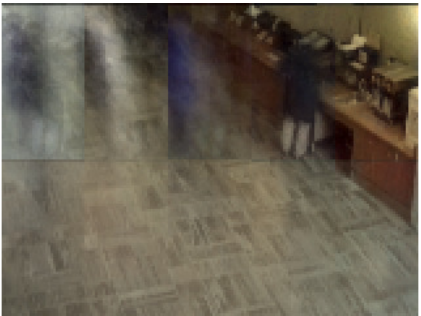}&
\includegraphics[width=0.124\textwidth]{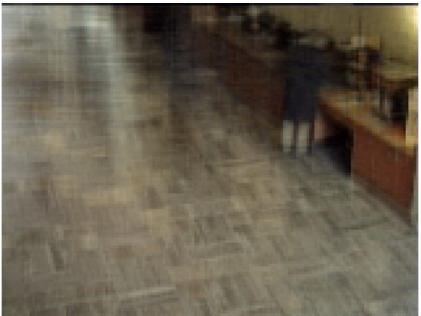}&
\includegraphics[width=0.124\textwidth]{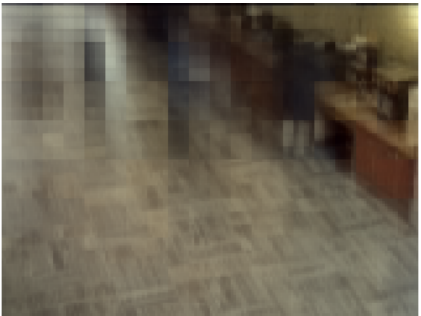}&
\includegraphics[width=0.124\textwidth]{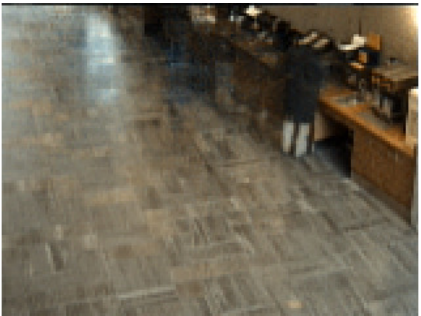}&
\includegraphics[width=0.124\textwidth]{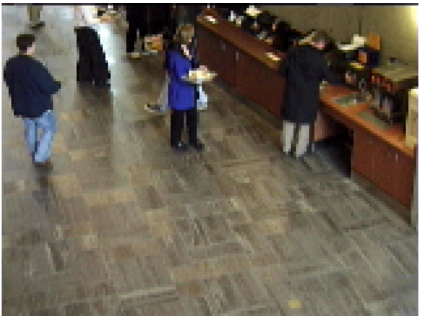} \\
\includegraphics[width=0.124\textwidth]{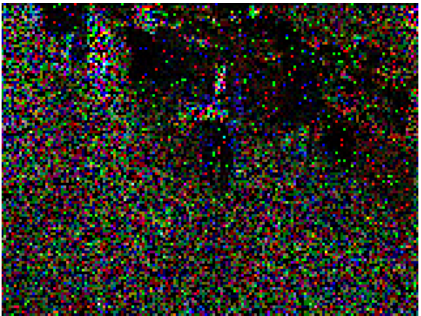}&
\includegraphics[width=0.124\textwidth]{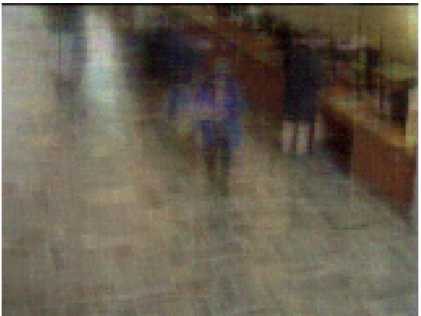}&
\includegraphics[width=0.124\textwidth]{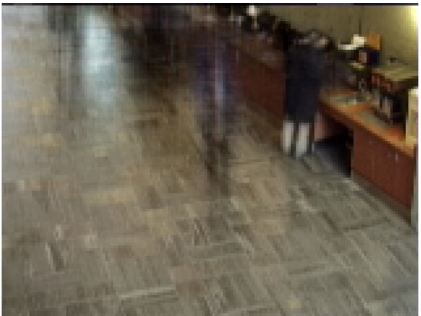}&
\includegraphics[width=0.124\textwidth]{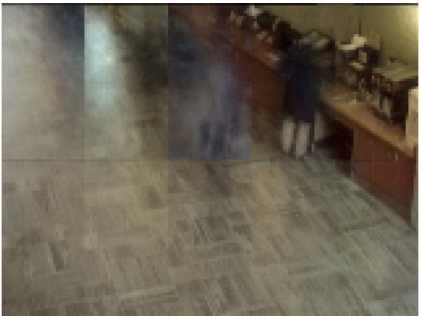}&
\includegraphics[width=0.124\textwidth]{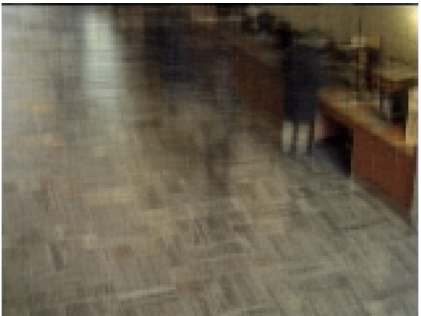}&
\includegraphics[width=0.124\textwidth]{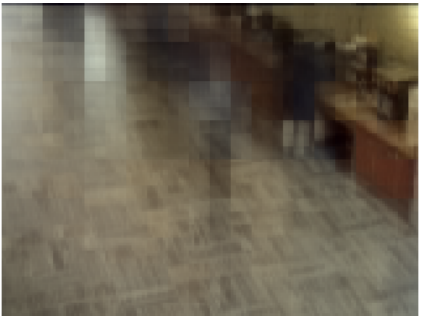}&
\includegraphics[width=0.124\textwidth]{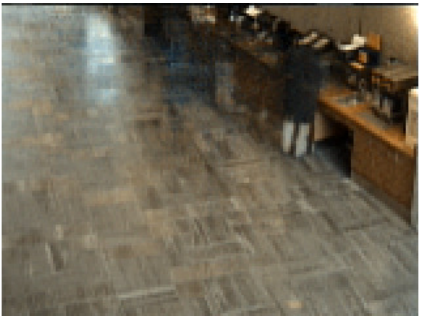}&
\includegraphics[width=0.124\textwidth]{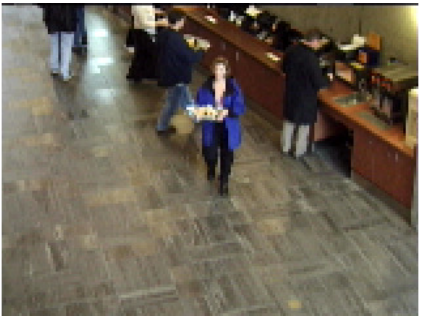} \\
\includegraphics[width=0.124\textwidth]{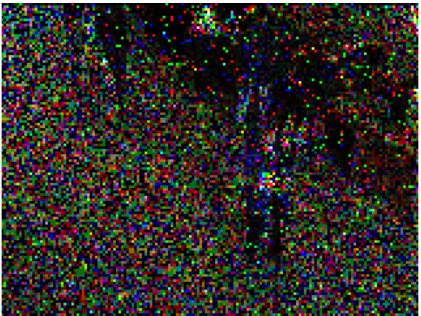}&
\includegraphics[width=0.124\textwidth]{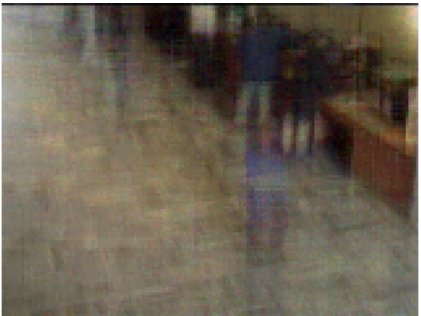}&
\includegraphics[width=0.124\textwidth]{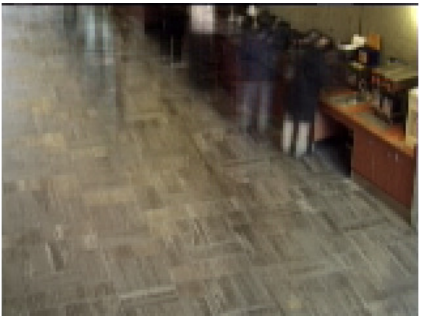}&
\includegraphics[width=0.124\textwidth]{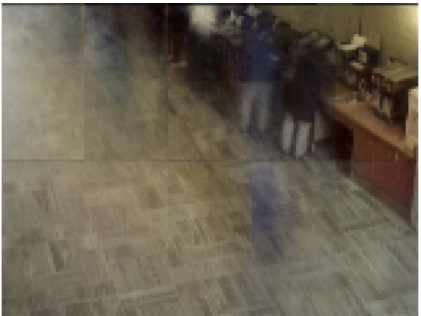}&
\includegraphics[width=0.124\textwidth]{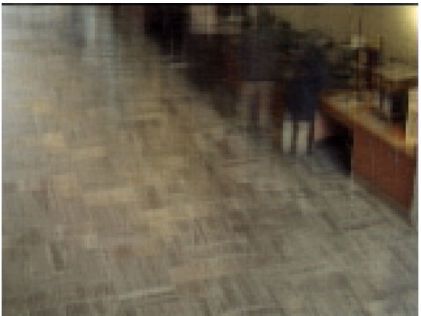}&
\includegraphics[width=0.124\textwidth]{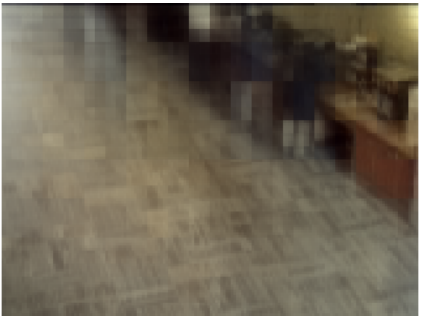}&
\includegraphics[width=0.124\textwidth]{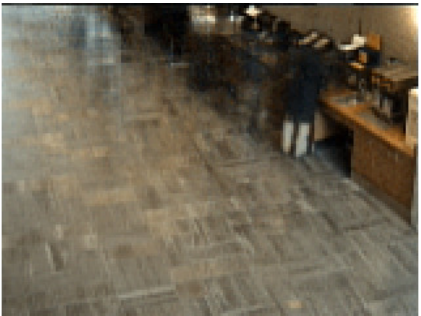}&
\includegraphics[width=0.124\textwidth]{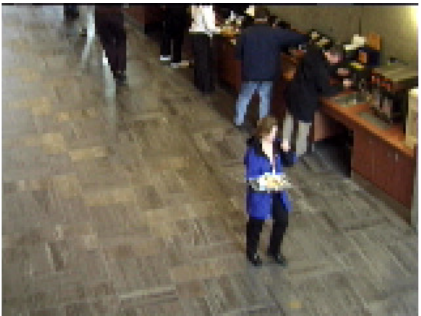} \\
\includegraphics[width=0.124\textwidth]{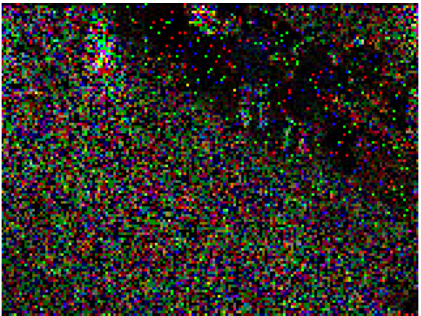}&
\includegraphics[width=0.124\textwidth]{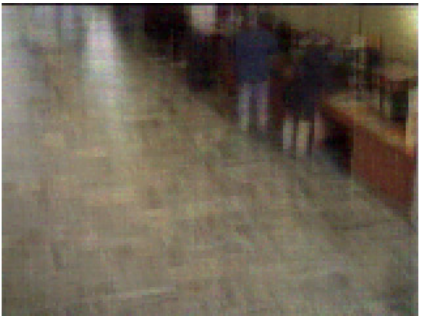}&
\includegraphics[width=0.124\textwidth]{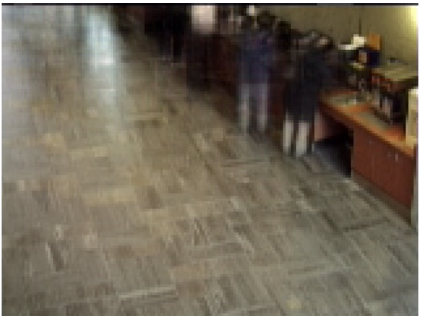}&
\includegraphics[width=0.124\textwidth]{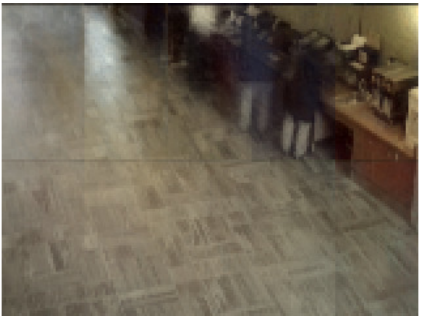}&
\includegraphics[width=0.124\textwidth]{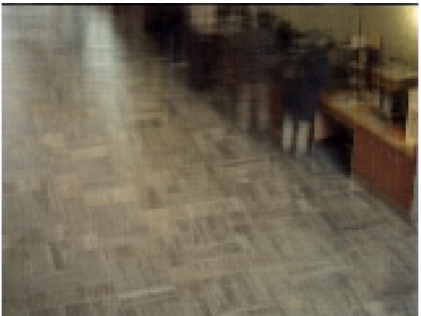}&
\includegraphics[width=0.124\textwidth]{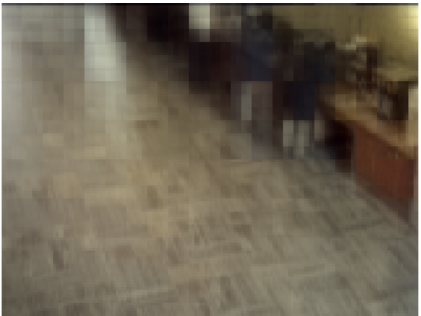}&
\includegraphics[width=0.124\textwidth]{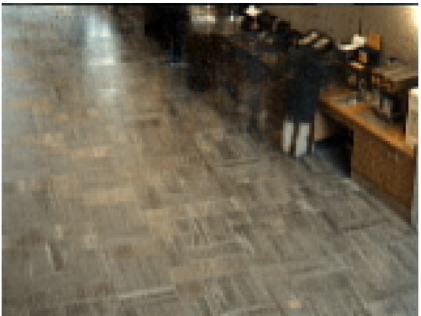}&
\includegraphics[width=0.124\textwidth]{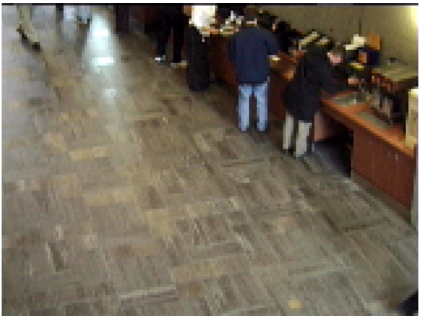} \\
Observed&SNN&TNN&TT&RTRC&RC-FCTN&RNC-FCTN&Ground truth  \\
  \end{tabular}
  \caption{The visual results of the 17th frame, the 27th frame, the 37th frame, and the 47th frame of six robust competition methods.} \vspace{-0.3cm}
  \label{back}
  \end{center}
\end{figure}

Furthermore, Fig. \ref{videos} shows the visual results and their corresponding residual images (the mean of the absolute difference between three color channels of the recovered images and the ground truth) of the two color videos with SR=0.2 and SaP=0.1. From Fig. \ref{videos}, it is easy to see that the proposed methods are markedly superior in removing noise and preserving details than the compared ones, such as grasses in the $\mathit{bunny}$ and construction in the $\mathit{elephants}$.

\subsection{Hyperspectral video completion}\label{rhsv}
To verify the effectiveness of the proposed methods, we conduct experiments on the hyperspectral video (HSV) \footnote{The data is available at http://openremotesensing.net/kb/data/.} (height $\times$  width $\times$ band $\times$ frames) of size $60 \times 60 \times 20 \times 20$. HSV contains a wealth of information, thus we consider more challenging situations. We conduct RTC problem for the HSV with different SRs $\{0.3, 0.2, 0.1\}$ and SaP $\{0.1,0.2\}$.

We report MPSNR and MSSIM values obtained by all compared restoration methods on the HSV in Table \ref{tab:addlabel2}. The data in Table \ref{tab:addlabel2} indicates that the proposed methods obtain an overall better performance than the compared methods. As the SaP increases, the advantage of the proposed RNC-FCTN over the compared methods is more prominent. Furthermore, Fig. \ref{HSV} shows the visual results and their corresponding residual images of the HSV with SR=0.1, SaP= 0.1 and 0.2, respectively. For visual effect, we have selected three bands to show pseudo-color images. From Fig. \ref{HSV}, the SaP in the reconstructed results of SNN and TNN is not removed well. The results reconstructed by the proposed methods are closer to the real image in terms of color than those of compared methods. The corresponding residual images can clearly confirm this phenomenon.

\subsection{Video background subtraction}\label{rback}
In this subsection, we apply the two proposed methods to color video background subtraction. We pick up consecutive 50 frames of $\mathit{bootstrap}$ \footnote{The data is available  https://www.microsoft.com/en-us/download/details.aspx?id=54651.} which forms a $120 \times 160 \times 3 \times 50$ tensor. This video consists of a static architectural background and moving foreground, such as moving people. Since the background components of all frames are highly correlated, it can be regarded as a low-rank tensor. The foreground components occupy a few locations of the entire video, and it can be regarded as a sparse tensor.

\begin{figure}[!t]
\centering
\includegraphics[width=0.85\linewidth]{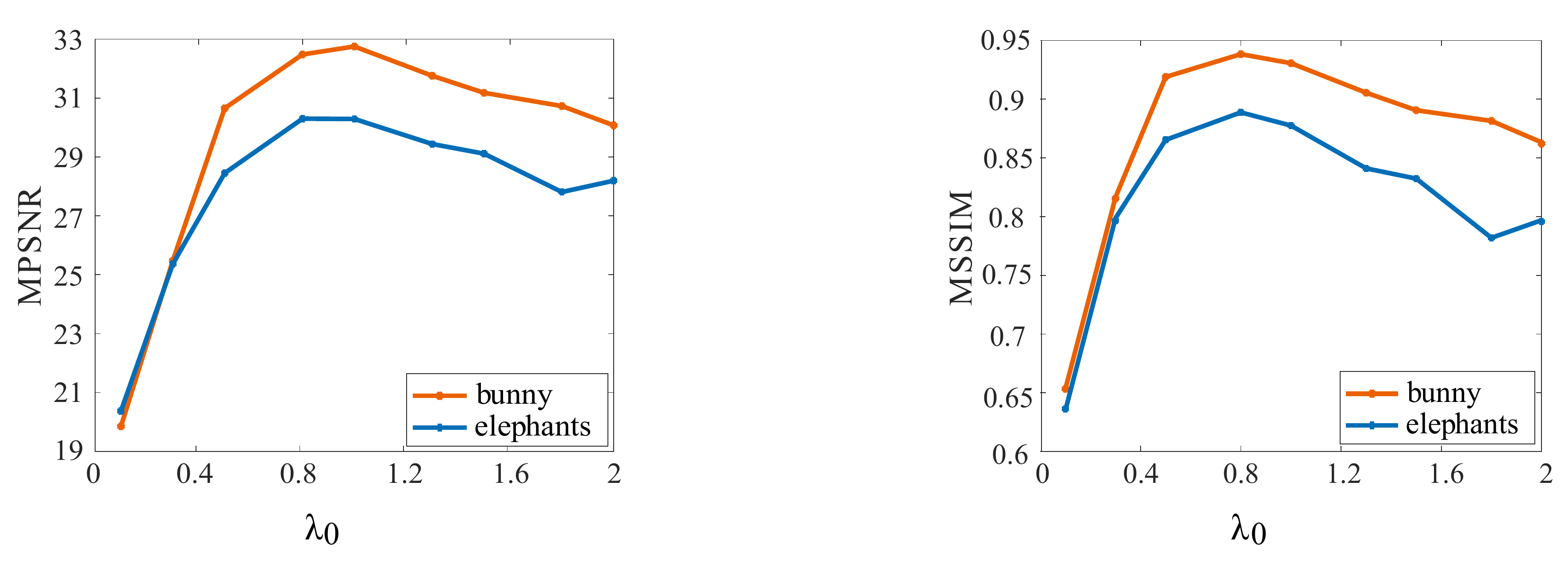}
\caption{The PSNR and SSIM values recovered by RNC-FCTN with respect to $\lambda$ for color videos.}\vspace{-0.2cm}
 \label{re1}
\end{figure}

\begin{figure}[htp]
\centering
\includegraphics[width=0.99\linewidth]{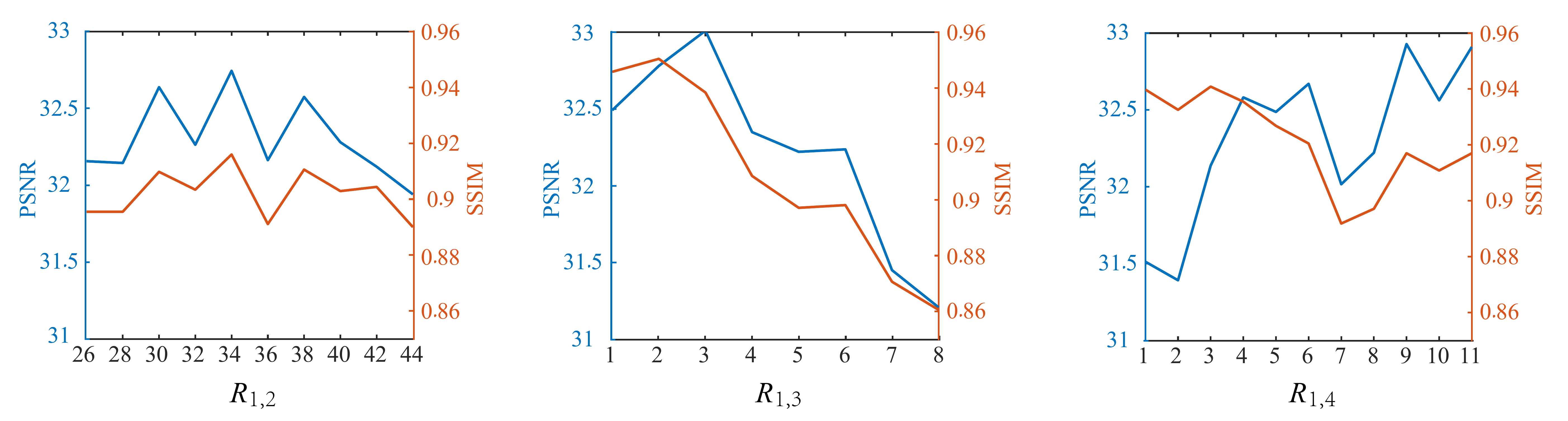}
  \caption{The influence of the FCTN rank.}
  \label{rank}
\end{figure}

\begin{figure}[htp]
\centering
\includegraphics[width=0.75\linewidth]{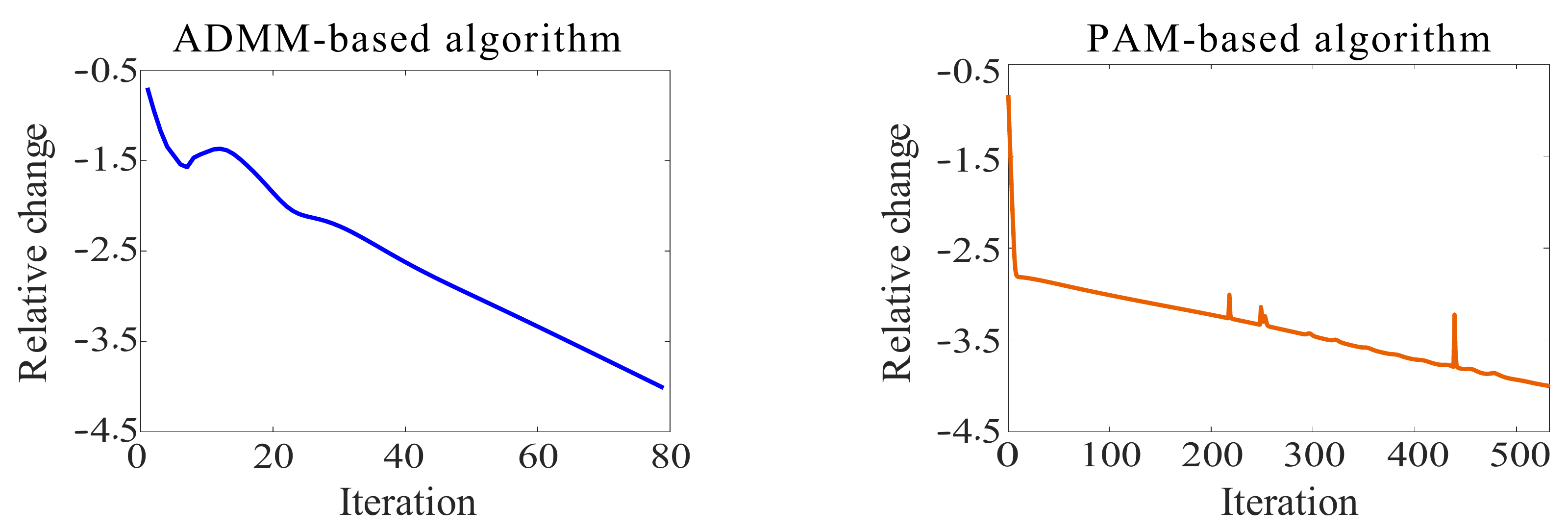}
  \caption{The history of relative change curves.}\vspace{-0.4cm}
  \label{error}
\end{figure}

We design a challenging task that executes a background subtraction from a corrupted video with SR=0.4 and SaP=0.1. Fig. \ref{back} presents the visual comparison of the 17th frame, the 27th frame, the 37th frame, and the 47th frame of six robust competition methods. As observed, the proposed methods, especially RNC-FCTN, simultaneously extract background and preserve the global structure in completing the missing entries. The reason is that the FCTN rank can be flexibly adjusted to separate the low-rank background and sparse foreground well.

\subsection{Discussion}
In this section, we experimentally analysis the influence of parameters in the proposed methods.

$\mathit{Influence}$ $\mathit{of}$ $\mathit{regularization}$ $\mathit{parameter}$ $\lambda$: The regularization parameter $\lambda$ actually balances the relationship between the low-FCTN-rank term and sparse term. Fig. \ref{re1} illustrates the influence of regularization parameter $\lambda$\footnote{The trend of the line graph obtained by RC-FCTN is the same as that obtained by RNC-FCTN, we only show the line graph obtained by RNC-FCTN here.}on two color videos with SR=0.2 and SaP=0.1, where $\lambda=\lambda_0/\sqrt{{\rm{max}}(I_1,I_2)I_3I_4}$. It achieves the optimal recovered performance with a suitable regularization parameter $\lambda$, balancing the low-FCTN-rank structure and sparse component well.

$\mathit{FCTN}$-$\mathit{rank}$: The FCTN-rank characterizes the correlation between any two modes. On color videos and hyperspectral videos, since $R_{1,4}^{\rm{max }}$ and $R_{2,4}^{\rm{max}}$ characterize the correlation between the spatial mode (height and width, respectively) and the temporal mode, we set them as the same value. And since $R_{1,3}^{\rm{max }}$, $R_{2,3}^{\rm{max }}$, and $R_{3,4}^{\rm{max}}$ characterize the correlation between the third mode (color channel and spectrum in color videos and multi-temporal remote sensing images, respectively) and other mode, we also set them as the same value. In brief, we just need to adjust $R_{1,2}^{\rm{max }}$, $R_{1,3}^{\rm{max }}$ and $R_{1,4}^{\rm{max}}$. On hyperspectral video, we simply set $R_{k_1,k_2}^{\rm{max}}$ as the same value.

With SR=0.2 and SaP=0.1 on color video $\mathit{bunny}$, we show the influence of FCTN-rank in Fig. \ref{rank}. With the growth of the estimated FCTN-rank, the MPSNR and MSSIM values first increase and then decrease. The reason is that when the estimated FCTN-rank is small, global correlation cannot be explored well. When the estimated FCTN-rank is large, noise cannot be well-removed.

$\mathit{Convergence}$ $\mathit{analysis}$: To corroborate the Theorem \ref{Thet1} and \ref{The}, we experimentally analysis the numerical convergence behavior. With SR=0.2 and SaP=0.1 on color video $\mathit{bunny}$, we show the relative change curves of ADMM-based algorithm and PAM-based algorithm in Fig. \ref{error}. The relative error curves finally converge to the relative error value $10^{-4}$ that we set in the Algorithm \ref{FCTN1} and \ref{FCTN2}, which confirms the convergence of our algorithm.
\section{Conclusion}\label{conclusion}
In this paper, we firstly suggest a FCTN nuclear norm as a convex surrogate of FCTN rank. Based on the FCTN nuclear norm, we propose a robust convex optimization model RC-FCTN for the RTC problem. Then, we theoretically establish the exact recovery conditions that one can recover a tensor of low-FCTN-rank exactly with overwhelming probability provided that its rank is sufficiently small and its corrupted entries are reasonably sparse. We develop an ADMM-based algorithm to solve the proposed RC-FCTN, which enjoys the global convergence guarantee. Moreover, we proposed a robust nonconvex optimization model RNC-FCTN for the RTC problem. Then, we theoretically derive the convergence guarantee of the PAM-based algorithm. Experimental results demonstrate the usefulness of proposed methods with compared ones.
\bibliography{reference}

\begin{thebibliography}{10}

\bibitem{multi}
D.~Qiu, M.~Bai, M.~K. Ng, and X.~Zhang.
\newblock Robust low transformed multi-rank tensor methods for image alignment.
\newblock {\em J. Sci. Comput.}, 87(24), 2021.

\bibitem{8606166}
C.~Lu, J.~Feng, Y.~Chen, W.~Liu, Z.~Lin, and S.~Yan.
\newblock Tensor robust principal component analysis with a new tensor nuclear
  norm.
\newblock {\em IEEE Trans. Pattern Anal. Mach. Intell.}, 42(4):925--938, 2020.

\bibitem{YANG2020124783}
J.~Yang, X.~Zhao, T.~Ji, T.~Ma, and T.~Huang.
\newblock Low-rank tensor train for tensor robust principal component analysis.
\newblock {\em Appl. Math. Comput.}, 367, 2020.

\bibitem{article1}
X.~Zhao, M.~Bai, and M.~K. Ng.
\newblock Nonconvex optimization for robust tensor completion from grossly
  sparse observations.
\newblock {\em J. Sci. Comput.}, 85(2):46, 2020.

\bibitem{CHEN2021100}
C.~Chen, Z.~Wu, Z.~Chen, Z.~Zheng, and X.~Zhang.
\newblock Auto-weighted robust low-rank tensor completion via tensor-train.
\newblock {\em Inform. Sci.}, 567:100--115, 2021.

\bibitem{xiong}
X.~Zhang, M.~K. Ng, and M.~Bai.
\newblock A fast algorithm for deconvolution and {Poisson} noise removal.
\newblock {\em J. Sci. Comput.}, 75(3):1535--1554, 2018.

\bibitem{zhuang}
L.~Zhuang, X.~Fu, M.~K. Ng, and J.~M. Bioucas-Dias.
\newblock Hyperspectral image denoising based on global and nonlocal low-rank
  factorizations.
\newblock {\em IEEE Trans. Geosci. Remote Sens}, pages 1--17, 2021.

\bibitem{jia}
Z.~Jia and M.~Wei.
\newblock A new {TV}-stokes model for image deblurring and denoising with fast
  algorithms.
\newblock {\em J. Sci. Comput.}, 72(2):522--541, 2017.

\bibitem{li}
J.~Li, W.~Li, S.~Vong, Q.~Luo, and M.~Xiao.
\newblock A riemannian optimization approach for solving the generalized
  eigenvalue problem for nonsquare matrix pencils.
\newblock {\em J. Sci. Comput.}, 82, 2020.

\bibitem{che}
M.~Che and Y.~Wei.
\newblock Multiplicative algorithms for symmetric nonnegative tensor
  factorizations and its applications.
\newblock {\em J. Sci. Comput.}, 83(53), 2020.

\bibitem{che2}
M.~Che, Y.~Wei, and H.~Yan.
\newblock An efficient randomized algorithm for computing the approximate
  {Tucker} decomposition.
\newblock {\em J. Sci. Comput.}, 88(32), 2021.

\bibitem{Tucker}
L.~R. Tucker.
\newblock Some mathematical notes on three-mode factor analysis.
\newblock {\em Psychometrika}, 31(3):279--311, 1966.

\bibitem{tubal}
M.~E. Kilmer, K.~Braman, N.~Hao, and R.~C. Hoover.
\newblock Third-order tensors as operators on matrices: A theoretical and
  computational framework with applications in imaging.
\newblock {\em SIAM J. Matrix Anal. Appl.}, 34(1):148--172, 2013.

\bibitem{TT}
I.~V. Oseledets.
\newblock Tensor-train decomposition.
\newblock {\em SIAM J. Sci. Comput.}, 33(5):2295--2317, 2011.

\bibitem{TR}
Q.~Zhao, G.~Zhou, S.~Xie, L.~Zhang, and A.~Cichocki.
\newblock Tensor ring decomposition.
\newblock {\em arXiv preprint arXiv:1606.05535}, 2016.

\bibitem{2512329}
Christopher~J. Hillar and Lek-Heng Lim.
\newblock Most tensor problems are {NP}-hard.
\newblock {\em J. ACM}, 60(6):1--39, 2013.

\bibitem{6138863}
J.~{Liu}, P.~{Musialski}, P.~{Wonka}, and J.~{Ye}.
\newblock Tensor completion for estimating missing values in visual data.
\newblock {\em IEEE Trans. Pattern Anal. Mach. Intell.}, 35(1):208--220, 2013.

\bibitem{2512321}
B.~Huang, C.~Mu, D.~Goldfarb, and J.~Wrigh.
\newblock Provable models for robust low-rank tensor completion.
\newblock {\em Pac. J. Optim.}, 11(2):339--364, 2015.

\bibitem{KILMER2011641}
M.~E. Kilmer and C.~D. Martin.
\newblock Factorization strategies for third-order tensors.
\newblock {\em Linear Algeb. Appl.}, 435(3):641--658, 2011.

\bibitem{6737273}
O.~Semerci, N.~Hao, M.~E. Kilmer, and E.~L. Miller.
\newblock Tensor-based formulation and nuclear norm regularization for
  multienergy computed tomography.
\newblock {\em IEEE Trans. Image Process.}, 23(4):1678--1693, 2014.

\bibitem{Jiang}
J.~Q. Jiang and M.~K. Ng.
\newblock Exact tensor completion from sparsely corrupted observations via
  convex optimization.
\newblock {\em arXiv: 1708.00601}, 2017.

\bibitem{8606165}
G.~Song, M.~K. Ng, and X.~Zhang.
\newblock Robust tensor completion using transformed tensor singular value
  decomposition.
\newblock {\em Numer. Linear Algeb. Appl.}, 27(3), 2020.

\bibitem{7859390}
J.~A. Bengua, H.~N. Phien, H.~D. Tuan, and M.~N. Do.
\newblock Efficient tensor completion for color image and video recovery:
  Low-rank tensor train.
\newblock {\em IEEE Trans. Image Process.}, 26(5):2466--2479, 2017.

\bibitem{yu2019tensor}
J.~{Yu}, C.~{Li}, Q.~{Zhao}, and G.~{Zhou}.
\newblock In {\em ICASSP 2019}, pages 3142--3146.

\bibitem{9136899}
H.~Huang, Y.~Liu, Z.~Long, and C.~Zhu.
\newblock Robust low-rank tensor ring completion.
\newblock {\em IEEE Trans. Comput. Imaging}, 6:1117--1126, 2020.

\bibitem{8606065}
Y.~Zheng, T.~Huang, X.~Zhao, Q.~Zhao, and T.~Jiang.
\newblock Fully-connected tensor network decomposition and its application to
  higher-order tensor completion.
\newblock {\em in Proceedings of the AAAI Conf. Artifi. Intell.}, 2021.

\bibitem{ye2019tensor}
K.~Ye and L.-H. Lim.
\newblock Tensor network ranks.
\newblock {\em arXiv: 1801.02662}, 2019.

\bibitem{1970395}
E.~J. Cand\`{e}s, X.~Li, Y.~Ma, and J.~Wright.
\newblock Robust principal component analysis?
\newblock {\em J. ACM}, 58(3):1--37, 2011.

\bibitem{ADMM1}
S.~Boyd, N.~Parikh, E.~Chu, B.~Peleato, and J.~Eckstein.
\newblock Distributed optimization and statistical learning via the alternating
  direction method of multipliers.
\newblock {\em Found. and Trends in Mach. Learn.}, 3(1):1--122, 2011.

\bibitem{ADMM2}
X.~Zhang.
\newblock A nonconvex relaxation approach to low-rank tensor completion.
\newblock {\em IEEE Trans. Neural Netws. Learn. Syst.}, 30(6):1659--1671, 2019.

\bibitem{KL}
J.~Bolte, A.~Daniilidis, A.~Lewis, and M.~Shiota.
\newblock Clarke subgradients of stratifiable functions.
\newblock {\em SIAM J. Optim.}, 18(2):556--572, 2007.

\bibitem{PAM}
J.~Bolte, S.~Sabach, and M.~Teboulle.
\newblock Proximal alternating linearized minimization for nonconvex and
  nonsmooth problems.
\newblock {\em Math. Prog.}, 146:459--494, 2014.

\bibitem{PAM1}
H.~Attouch, J.~Bolte, and B.~Svaiter.
\newblock Convergence of descent methods for semi-algebraic and tame problems:
  {Proximal} algorithms, forward-backward splitting, and regularized
  {Gauss-Seidel} methods.
\newblock {\em Math. Prog.}, 137:91--129, 2013.

\bibitem{502}
N.~Yair and T.~Michaeli.
\newblock Multi-scale weighted nuclear norm image restoration.
\newblock {\em in Proceedings of the CVPR}, pages 3165--3174, 2021.

\bibitem{8854307}
Y.~Zheng, T.~Huang, X.~Zhao, T.~Jiang, T.~Ma, and T.~Ji.
\newblock Mixed noise removal in hyperspectral image via low-fibered-rank
  regularization.
\newblock {\em IEEE Trans. Geosci. Remote Sens}, 58(1):734--749, 2020.

\end{thebibliography}
\bibliographystyle{unsrt}

\end{document}